\documentclass{article}
\usepackage{arxiv}

\usepackage{microtype}
\usepackage{graphicx}
\usepackage{subfigure}
\usepackage{booktabs} 
\usepackage{xcolor}
\usepackage{hyperref}
\usepackage{wrapfig}

\usepackage{amsmath}
\usepackage{amssymb}
\usepackage{mathtools}
\usepackage{amsthm}
\usepackage{booktabs, multirow, multicol}
\usepackage{listings}

\usepackage[capitalize,noabbrev]{cleveref}

\theoremstyle{plain}

\theoremstyle{definition}

\theoremstyle{remark}

\usepackage{amsmath}
\usepackage{bm, amsmath, physics, amssymb}

\usepackage[utf8]{inputenc} 
\usepackage[T1]{fontenc}    
\usepackage{hyperref}       
\usepackage{url}            
\usepackage{booktabs}       
\usepackage{amsfonts}       
\usepackage{nicefrac}       
\usepackage{microtype}      
\usepackage{lipsum}		
\usepackage{graphicx}
\usepackage{natbib}
\usepackage{doi}
\usepackage{caption}
\usepackage{algorithmic}
\usepackage[ruled,vlined,linesnumbered]{algorithm2e}
\usepackage{subcaption}
\usepackage{multicol}
\usepackage{multirow}

\title{Towards Foundation Time Series Model: To Synthesize Or Not To Synthesize?}

\author{%
Kseniia Kuvshinova$^{1,3}$\thanks{These authors contributed equally to this work.} \ \thanks{Corresponding author.} \quad Olga Tsymboi$^{1,2}$\footnotemark[1] \quad Alina Kostromina$^{1,4}$ \quad Dmitry Simakov$^{1}$ \quad Elizaveta Kovtun$^{1,3}$
\\
$^1$Sber AI Lab, Moscow, Russia \\ $^2$Moscow Institute of Physics and Technology, Moscow, Russia
\\ $^3$Skolkovo Institute of Science and Technology, Moscow, Russia\\
$^4$HSE University, Moscow, Russia\\
\texttt{Kseniia.Kuvshinova@skoltech.ru}, \texttt{tsimboy.oa@phystech.edu}}

\date{}

\hypersetup{
pdftitle={A template for the arxiv style},
pdfsubject={q-bio.NC, q-bio.QM},
pdfauthor={David S.~Hippocampus, Elias D.~Striatum},
pdfkeywords={First keyword, Second keyword, More},
}

\begin{document}
\maketitle
\begin{abstract}
The industry is rich in cases when we are required to make forecasting for large amounts of time series at once. However, we might be in a situation where we can not afford to train a separate model for each of them. Such issue in time series modeling remains without due attention. The remedy for this setting is the establishment of a foundation model. Such a model is expected to work in zero-shot and few-shot regimes. However, what should we take as a training dataset for such kind of model?

Witnessing the benefits from the enrichment of NLP datasets with artificially-generated data, we might want to adopt their experience for time series. In contrast to natural language, the process of generation of synthetic time series data is even more favorable because it provides full control of series patterns, time horizons, and number of samples. In this work, we consider the essential question if it is advantageous to train a foundation model on synthetic data or it is better to utilize only a limited number of real-life examples. Our experiments are conducted only for regular time series and speak in favor of leveraging solely the real time series. Moreover, the choice of the proper source dataset strongly influences the performance during inference. When provided access even to a limited quantity of short time series data, employing it within a supervised framework yields more favorable results than training on a larger volume of synthetic data. The code for our experiments is publicly available on Github \url{https://github.com/sb-ai-lab/synthesize_or_not}.
\end{abstract}   
\section{Introduction}
\label{sec:intro}

Time series data, characterized by sequential observations indexed in chronological order, serve as a foundational element across diverse domains such as finance, weather forecasting, and healthcare. The inherent temporal dependencies within these data necessitate sophisticated models capable of capturing intricate patterns and making accurate predictions. Traditional methods like autoregressive models, moving averages, and recurrent neural networks had been employed for a long time until Transformer-based models \cite{zhou2021informer,wu2021autoformer,zhou2022fedformer} sparked a paradigm shift in time series modeling.

\begin{figure}
    \centering
    \includegraphics[width=0.5\textwidth]{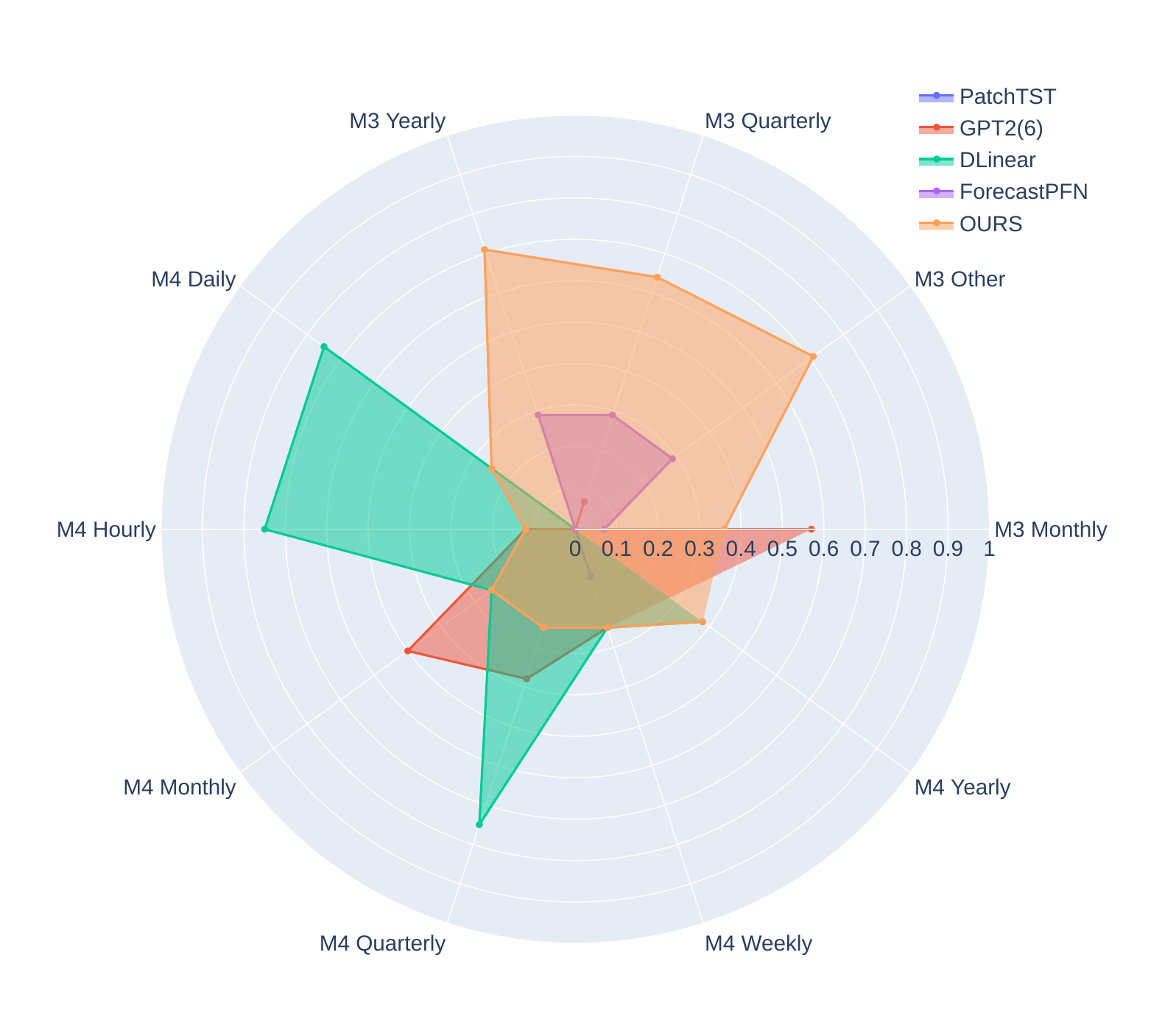}
    \caption{Zero-shot performance of time-series forecasters, trained on various source datasets, excluding ForecastPFN and OURS which were trained on synthetic data. The model's success rate for target datasets (i.e., all benchmark datasets except for the M datasets) is assessed based on Mean Absolute Error (MAE) for five different forecasting models. The findings indicate that the zero-shot performance of these models is significantly influenced by the source dataset, thereby prompting inquiries into the efficiency of transferability.
    }
    \label{fig:teaser}
\end{figure}

Most industrial time series data, such as demand and sales data, tends to be relatively short \cite{smyl2016data}. This limitation is especially noticeable in dynamic online environments, like marketplaces, where frequent updates occur. For instance, in the M3 time series competition \cite{makridakis2000m3}, there were 645 series with annual seasonality, ranging from 14 to 40 data points, with a maximum prediction horizon of 6. This constraint is also evident in datasets related to traffic\footnote{https://www.kaggle.com/datasets/leonardo00/istanbul-traffic-index}, stock\footnote{https://www.kaggle.com/datasets/yeemeitsang/tsmc-stock-exchange-2022}, and power\footnote{https://www.kaggle.com/datasets/dharanikra/electrical-power-demand-in-turkey}. The scarcity of data points poses a challenge for effectively training large neural networks.

Another common challenge in the industry is a large amount of incoming series, which requires forecast construction. Meanwhile, the traditional ARIMA-type \cite{box1968some} methods train on individual series. This issue is also present in the open-source solution from Meta - Prophet \cite{taylor2018forecasting}, where a model is devoted only to one time series. Such logic entails the training of a large number of time series models to use them for inference. This may turn into heightened computational costs \cite{shi2020block} and demand in out-of-reach amount of resources.

Several approaches have been proposed to address the mentioned challenges. They include global modeling \cite{gasthaus2019probabilistic, oreshkin2019n}, accelerating ARIMA \cite{shi2020block}, online forecasting \cite{pham2022learning, george2023online, michael2023ofter, zhang2023onenet}, utilizing synthetic data \cite{dooley2023forecastpfn}, and methods involving learning from one dataset and applying the absorbed knowledge to deal with another \cite{oreshkin2021meta, zhou2023one}. The latter approach is often called zero-shot in time series, but we also use the term transferability for it. However, in the existing approaches for zero-shot the problem of consistency within characteristics of source and target datasets should be addressed. For instance, in the Frozen Pretrained Transformer (FPT)  \cite{zhou2023one} based on GPT2 backbone \cite{radford2019language}, transferability involves aligning the granularity of data, namely source and target data seasonality. However, in the case of datasets containing series of different natures, the most suitable solution might be the production of synthetic data with predefined characteristics rather than handpicking a proper source dataset for the zero-shot setup. To the best of our knowledge, the only paper achieving true zero-shot by pretraining on synthetic data, ForecastPFN \cite{dooley2023forecastpfn}, has limited options for granularity, in particular, only five types of seasonality. Hence, there is a need for more versatile capabilities of zero-shot models capable in terms of coping with diverse time series patterns. Furthermore, recent efforts have focused on pretraining time series models \cite{yeh2023toward, jin2023time, garza2023timegpt} for multitasking such as detrending, denoising, and anomaly detection without the need for finetuning. The serious restriction is the possible absence of access to the true trends, anomalies, or other forms of markup. Additionally, the fixed time horizon in real train datasets can impose limits on forecasting length. Again, these problems can be easily addressed with the agile synthesis of synthetic series.

In general, we observe clear endeavors from the community to overcome different parts of one central problem, namely the drawbacks of real-world data available for training foundation time series model. At first glance, the synthetically-generated data with rich properties and big volumes has the potential to become a perfect alternative to scarce and inconvenient real data. Nevertheless, there is no conducted comprehensive comparison between cases of availing artificial and natural time series. To analyze this, we generate diversified series within the proposed framework to train foundation models in a zero-shot setting. What is more, there is no detailed analysis of the influence of the source and target dataset choice in the transferability setup.
Summing up, our main contribution is as follows: 
\begin{itemize}
    \item   We reassess the utilization of synthetic data in pretraining for time series models and propose an alternative method for its generation. We also investigate whether synthetic data aids in achieving higher-quality zero-shot forecasts or we should rest content with available real-life examples. As a result of thorough experiments with regular time series, we find out that the use of synthetically-generated data does not lead to performance gain in a zero-shot setting compared to the utilization of even a limited number of real-life data.
    \item We demonstrate that the choice of the source dataset greatly affects metrics on the target dataset in transferability setup. For instance, pretraining using the M3 dataset as a source results in inferior performance across various models compared to pretraining with the M4 dataset for most models.
    \item We show that in case of access to at least a few real examples, one should opt for the supervised setting instead of zero-shot for greater predictive power.
\end{itemize}

\section{Related Work}
\label{seq:related}

\textbf{Time Series Forecasting} The need for time series forecasting without having a narrow domain expertise arises in many applications. A prominent representative of classical statistical tool is ARIMA \cite{box1968some, box1970distribution}. However, the necessity of manual parameter selection and challenges with long-term forecasting for statistical models make the adoption of deep learning models more appealing. To deal with the sequential data, recurrent neural networks (RNN) were designed. Their weakness with gradients vanishing and explosion provokes the creation of LSTM \cite{hochreiter1997long} and GRU \cite{chung2014empirical}, which control the information flow with the gated elements. The significant drawbacks of recurrent architecture consist in the inability to effectively capture the long-term dependencies and poor performance with prediction on long horizons \cite{zhou2021informer}. Despite of conventional use of Convolutional Neural Networks (CNNs) in the Computer Vision domain, there are effective methods that employ convolutional filters in time series forecasting tasks \cite{bai2018empirical, wang2022micn}.

Transformer \cite{vaswani2017attention}, which is associated with the undisputed triumph in natural language processing (NLP), contributes to solving such the urgent problem from times series area as managing long context. The basic implementation implies the quadratic computation complexity and memory requirements. There was a burst of activity in the transformation of the original Transformer to the more efficient and time series-oriented models. For instance, Informer \cite{zhou2021informer} introduces the ProbSparse and distilling self-attention, reducing space and time complexity. Autoformer \cite{wu2021autoformer} employs season-trend decomposition and Auto-Correlation calculations instead of self-attention, processing better time series parts. Inspired by decomposition boost from earlier solutions and acknowledging the sparse representation of time series data within the Fourier basis, researchers introduced frequency-enhanced FEDformer \cite{zhou2022fedformer}. PatchTST \cite{Yuqietal-2023-PatchTST} operates with a series parts called patches and stands out among the discussed transformer-based models for time series with its performance. A notable and efficacious approach involving the application of attention and feedforward blocks to inverted dimensions was introduced (Liu et al., 2023). Notably, the modification of some soft spots from a time series perspective in Transformer, e.g. inefficient treatment with pronounced non-stationarity, can substantially enhance the performance. An example is the work \cite{liu2022non} that proposes Non-stationary Transformer with the modified attention block. 

Despite the vigorous development of transformed-based models and their strong performance, there are highly efficient models with attention-free architectures. For example, NHITS \cite{challu2023nhits} is composed of several MLPs with intervening nonlinearities and upgrades N-BEATS \cite{oreshkin2019n} architecture with multi-rate input data sampling and multi-scale interpolation technique. HNITS demonstrates performance superiority over such models as Informer \cite{zhou2021informer},  Autoformer \cite{wu2021autoformer}, and FEDformer \cite{zhou2022fedformer} in long-horizon forecasting setup. A recent contribution involves the introduction of a solution utilizing a parameter-efficient inception block (Wu et al., 2022), which entails transforming 1D time series data into a 2D space, thereby addressing the challenge of multi-periodicity. On the whole, the provocative question of the viability to exploit the sophisticated Transformers for time series forecasting is raised in \cite{zeng2023transformers}. The authors showcase that models (NLinear, DLinear) comprised of linear layers and transformation of input series manage to exceed time series Transformers in the prediction quality. TSMixer \cite{chen2023tsmixer} with a design of stacked multi-layer perceptrons that play the role of time and feature information mixer and foster predictive power. One more evidence of the linear model strength with a combination of reversible normalization (RevIN) called RLinear is presented in \cite{li2023revisiting}. Hence, there is an open opposition between Transformer-based and simpler linear-based models for time series prediction. One should choose the particular architecture relying on the specific problem limitations and peculiarities.

\textbf{Time Series Pre-Training.} The pre-training for time series models shows a great potential to become a reason for the performance improvement on the target tasks under various settings \cite{ma2023survey}. The work \cite{zhang2023self} provides an overview of self-supervised learning methods for time series data from three perspectives: generative-, contrastive-, and adversarial-based. The most common consequent target tasks are time series forecasting, classification, anomaly detection, and clustering which often benefit substantially from the presence of pre-training stage. TS2Vec \cite{yue2022ts2vec} is an example of a well-established framework for learning representations of time series in a contrastive way. The authors of \cite{wang2022learning} offer pre-training with autoencoders individually for trend and seasonal components.

\paragraph{Zero-shot and Transfer Learning for Time Series.} Despite the success of analogous approaches in other areas, zero-shot time series prediction remains largely unexplored territory with few available solutions. N-Beats \cite{oreshkin2019n, oreshkin2021meta} was among the first approaches to tackle this challenge for univariate time series. There are experiments with a prediction of time series with the models pre-trained on texts \cite{zhou2023one, gruver2023large, rasul2023lag, jin2023time}. For example, TimeGPT \cite{garza2023timegpt} based on vanilla Transformer is the first pre-trained foundation model for time series forecasting. In turn, Time-LLM \cite{jin2023time} has an intact backbone language model but expresses the input time series through text prototypes.

Overall, very little attention is paid to the consideration of the pre-training phase as a step towards the creation of a foundation model with the ability to work under zero-shot and few-shot settings. Even fewer works try to get value fully or partially from artificially-created samples for training zero-shot models \cite{dooley2023forecastpfn, das2023decoder}. In our work, we aim to close these gaps and propose a flexible approach to generate diverse and representative series, in contrast with more complicated Generative Adversarial Networks (GANs) \cite{yoon2019time, pei2021towards}, and conduct an in-depth study of the rationality behind exploiting synthetic training data. 
\section{Rethinking synthetic prior}
\label{sec:framework}
Frequently, historical time series data lacks the volume required to effectively train extensive neural networks. Consequently, there is a growing demand for models capable of learning from few or even zero-shot instances (for task definition see Appendix~\ref{app:task_definition}).
When examining time series patterns, it becomes evident that they comprise trend and seasonal elements. These components lend themselves well to modeling through Fourier series, augmented by diverse additive or multiplicative noises.

Drawing inspiration from Prior Fitted Networks \cite{muller2021transformers} and TabPFN \cite{hollmann2022tabpfn} approach, we hypothesize the existence of a priori distribution that serves as the foundation for all real series data. We establish seasonality by sampling Fourier coefficients and trend through the inclusion of diverse analytical functions combined with the overall seasonal pattern; full description of datasets prior can be found in the Appendices \ref{app:generation_details}.

Unlike ForecastPFN, our model uses priors independent of natural time periodicity (weekly, monthly, and yearly seasonalities), causing a wider functional family underlying synthetic data. The usage of Fourier coefficients as parametrization for the seasonality allows us to model low-frequency functions, which are controlled via periodicity.

Inspired by TabPFN \cite{hollmann2022tabpfn}, as well as various methods and pretrainings for foundation models \cite{devlin2018bert}, we use an encoder-based transformer model which consists of 5 transformer encoder layers with 4 heads, hidden size 128 and feedforward network. We use a min-max scaler to put inputs in the interval [0, 1].
The scaled raw input series without time stamps is projected into hidden size via a simple linear layer. We use learnable positional encoding with sinus-based initialization for better convergence. Predictions are obtained from CLS-token embeddings after LeakyReLU activation. The last linear layer is 720, the same as the maximum horizon length in the long-term forecasting benchmark.
Driven by the fact that the maximum period length within our synthetic data is constrained to 200 and guided by the intuition that the model necessitates a minimum of two periods of data featuring a trend to learn this sequence effectively, we established a maximum model history window size of 500. This adjustment maintains adherence to the definition of PFN, as each series sample is treated as a distinct dataset within univariate time series models. Consequently, every synthetic sample is derived from the hypothesis set, as elaborated in Appendix~\ref{app:generation_details}.

We performed the computation of our model on one  GPU’s NVIDIA A100 of 80GB. Full model training took 40 GPU hours on 500 000 synthetic train samples with batch size $512$. We trained our transformer for 400 epochs with Adam optimizer with $0.003$ learning rate and cosine scheduler with half of an epoch warmup.

\section{Experiments}
\label{seq:experiments}
In this section, we present our empirical study on questions raised and concerns above.

\subsection{Experiments setup}
\label{sseq:setup}

\paragraph{Datasets.}

To evaluate the performance of our and other models, we use 8 real-world datasets generally accepted in the field of time series forecasting beginning with \cite{wu2021autoformer}: Weather, Electricity, Traffic, ILI and 4 ETT datasets. Also similar to \cite{zhou2023one}, we use M3 and M4 datasets from \cite{godahewa2021monash} with different granularities for zero-shot Forecasting. Additional description of datasets can be found in \ref{sseq:real_data_details}.

The datasets under consideration offer a variety of different time series behaviors. They vary not only in the number of data points and time series but also in the unique structures of the time series, including seasonal patterns, local and global trends, as well as shifts in the mean. That's why they are suitable candidates for the systematic comparison of model outcomes.

\paragraph{Models.}
In our empirical study we select representative baselines including MLP-based models: DLinear \cite{zeng2023transformers}; CNN-based models: TimesNet \cite{wu2022timesnet}; transformer-based models: PatchTST \cite{dosovitskiy2020image}, Autoformer \cite{wu2021autoformer}, FEDformer \cite{zhou2022fedformer}, GPT2(6)-adapter, GPT2(6)-frozen and GPT2(0) FPT \cite{zhou2023one}, Inverted Transformer (ITransformer) \cite{liu2023itransformer}, and, finally, ForecastPFN \cite{dooley2023forecastpfn}~--- the first to our knowledge zero-shot model trained on synthetic data.



\paragraph{Evaluation metrics.}

For comparison, we use common metrics to evaluate the quality of time series prediction models: MSE and MAE for long-term and few-shot multivariate forecasting; for zero-shot forecasting, we additionally report SMAPE.

\paragraph{Tasks definition.}
\label{app:task_definition}

Following the methodology outlined in GPT2 for time series \cite{zhou2023one}, we conducted few-shot and zero-shot experiments. However, we significantly modify the approach by introducing a step-wise methodology for our few-shot benchmark. Unlike \cite{zhou2023one} for time series, which defined few-shot as utilizing only 5\% and 10\% of the training data, our methodology estimates the seasonality period. Subsequently, it establishes 8 different history lengths based on this estimation. This approach, characterized by a more significant number of steps, mitigates the impact of substantial variations in training data volume across different datasets and facilitates a more in-depth exploration of the relationship between training size and model performance (See details in Appendix~\ref{app:few_shot_details}).

\subsection{Results}
\label{sseq:main_results}

\paragraph{Long-Term Multivariate forecasting.} 

New methods of zero-shot forecasting usually use specific methodology, new or rare datasets, and metrics. This situation makes it difficult to objectively compare different methods with each other and determine the position of new models in real-world applications.
Our goal was a fair comparison of zero-shot in-context learning approaches with SOTA supervised method on the commonly known, widely-used benchmark with established methodology. 

We fully reproduced methodology of long-term multivariate forecasting task from \cite{Yuqietal-2023-PatchTST,zhou2023one,liu2023itransformer} including data splits and specific prediction horizons: $\{24, 36, 48, 60\}$ for ILI and $T \in \{96, 192, 336, 720\}$ for the rest of datasets.

For zero-shot methods, we used a fixed look-back window from the original work of $L = 36$ for ForecastPFN and $L = 250$ for our PFN approach. Though, we could not conduct results for TimeGPT-1\cite{garza2023timegpt} due to the close-source model`s origin and API restrictions.  

Unfortunately, the technical realization of ForecastPFN produces a minor amount of NaN values in predictions
, so we were forced to drop such points during evaluation, and thus, PFN metrics might be slightly biased.

\begin{table*}[!htbt]
\resizebox{\textwidth}{!}{%
\begin{tabular}{c|rr|rr||rr|rr|rr|rr|rr|rr|rr|rr|rr}
\toprule
\multicolumn{1}{c|}{\textbf{Models}} & \multicolumn{2}{c|}{\textbf{Ours}} & \multicolumn{2}{c||}{\textbf{ForecastPFN}} & \multicolumn{2}{c|}{\textbf{GPT2(6)-A}} & \multicolumn{2}{c|}{\textbf{GPT2(6)-F}} & \multicolumn{2}{c|}{\textbf{GPT2(0)}} & \multicolumn{2}{c|}{\textbf{DLinear}} & \multicolumn{2}{c|}{\textbf{PatchTST}} & \multicolumn{2}{c|}{\textbf{TimesNet}} & \multicolumn{2}{c|}{\textbf{FEDformer}} & \multicolumn{2}{c|}{\textbf{Autoformer}} & \multicolumn{2}{c}{\textbf{ITransformer}} \\ 
\multicolumn{1}{c|}{\textbf{Dataset}}  & \multicolumn{1}{c}{\textbf{MSE}} & \multicolumn{1}{c|}{\textbf{MAE}} & \multicolumn{1}{c}{\textbf{MSE}} & \multicolumn{1}{c||}{\textbf{MAE}} & \multicolumn{1}{c}{\textbf{MSE}} & \multicolumn{1}{c|}{\textbf{MAE}} & \multicolumn{1}{c}{\textbf{MSE}} & \multicolumn{1}{c|}{\textbf{MAE}} & \multicolumn{1}{c}{\textbf{MSE}} & \multicolumn{1}{c|}{\textbf{MAE}} & \multicolumn{1}{c}{\textbf{MSE}} & \multicolumn{1}{c|}{\textbf{MAE}} & \multicolumn{1}{c}{\textbf{MSE}} & \multicolumn{1}{c|}{\textbf{MAE}} & \multicolumn{1}{c}{\textbf{MSE}} & \multicolumn{1}{c|}{\textbf{MAE}} & \multicolumn{1}{c}{\textbf{MSE}} & \multicolumn{1}{c|}{\textbf{MAE}} & \multicolumn{1}{c}{\textbf{MSE}} & \multicolumn{1}{c|}{\textbf{MAE}} & \multicolumn{1}{c}{\textbf{MSE}} & \multicolumn{1}{c}{\textbf{MAE}} \\ \midrule

\multicolumn{1}{c|}{\textbf{Weather}}   & 2.311	& 0.666	& \textbf{0.327}	& \textbf{0.331}	& \textbf{0.220}	& \textbf{0.250}	& 0.237	& 0.271	& 0.253	& 0.286	& 0.249	& 0.300	& 0.226	& 0.264	& 0.259	& 0.287	& 0.309	& 0.360	& 0.338	& 0.382	& 0.258	& 0.278 \\ 


\multicolumn{1}{c|}{\textbf{ETTh1}}   & 2.351	& 0.985	& \textbf{1.901}	& \textbf{0.944}	& \textbf{0.406}	& 0.427	& 0.428	& \textbf{0.426}	& 0.465	& 0.455	& 0.423	& 0.437	& 0.413	& 0.431	& 0.458	& 0.450	& 0.440	& 0.460	& 0.496	& 0.487	& 0.454	& 0.448 \\ 

\multicolumn{1}{c|}{\textbf{ETTh2}}   & 4.388	& 1.013	& \textbf{0.406}	& \textbf{0.441}	& 0.339	& 0.385	& 0.355	& 0.395	& 0.382	& 0.412	& 0.431	& 0.447	& \textbf{0.330}	& \textbf{0.379}	& 0.414	& 0.427	& 0.437	& 0.449	& 0.450	& 0.459	& 0.383	& 0.407 \\


\multicolumn{1}{c|}{\textbf{ETTm1}}   & 3.105	& 0.972	& \textbf{1.339}	& \textbf{0.729}	& \textbf{0.349}	& \textbf{0.372}	& 0.352	& 0.383	& 0.388	& 0.404	& 0.357	& 0.379	& 0.351	& 0.381	& 0.400	& 0.406	& 0.448	& 0.452	& 0.588	& 0.517	& 0.407	& 0.410 \\ 


\multicolumn{1}{c|}{\textbf{ETTm2}}   & 3.634	& 0.763	& \textbf{0.237}	& \textbf{0.321}& \textbf{0.248}	& \textbf{0.307}	& 0.267	& 0.326	& 0.285	& 0.340	& 0.267	& 0.334	& 0.255	& 0.315	& 0.291	& 0.333	& 0.305	& 0.349	& 0.327	& 0.371	& 0.288	& 0.332 \\ 


\multicolumn{1}{c|}{\textbf{ECL}}   & 1.769	& \textbf{0.757}	& \textbf{1.623}	& 0.977	& \textbf{0.159}	& \textbf{0.252}	& 0.167	& 0.263	& 0.166	& 0.260	& 0.166	& 0.264	& 0.162	& 0.253	& 0.193	& 0.295	& 0.214	& 0.327	& 0.227	& 0.338	& 0.178	& 0.270 \\ 


\multicolumn{1}{c|}{\textbf{ILI}}  & \textbf{2.423}	& \textbf{1.015}	& 4.653	& 1.599	& \multicolumn{1}{c}{-}	& \multicolumn{1}{c|}{-}	& 1.925	& 0.904	& 2.233	& 1.018	& 2.169	& 1.041	& \textbf{1.480}	& \textbf{0.806}	& 2.139	& 0.931	& 2.597	& 1.070	& 2.819	& 1.120	& \multicolumn{1}{c}{-} & \multicolumn{1}{c}{-} \\

\multicolumn{1}{c|}{\textbf{Traffic}}   & \textbf{2.009}	& \textbf{0.877}	& 2.996	& 1.218	& 0.397	& \textbf{0.257}	& 0.414	& 0.295	& 0.413	& 0.282	& 0.434	& 0.295	& \textbf{0.391}	& 0.264	& 0.620	& 0.336	& 0.610	& 0.376	& 0.628	& 0.379	& 0.428	& 0.282 \\ 

\midrule
\multicolumn{1}{c|}{\textbf{Average}}   & 2.749	& 0.881	& \textbf{1.685}	& \textbf{0.820}	& 0.265	& 0.281	& 0.518	& 0.408	& 0.573	& 0.432	& 0.562	& 0.437	& \textbf{0.451}	& \textbf{0.386}	& 0.597	& 0.433	& 0.670	& 0.480	& 0.734	& 0.507	& 0.300	& 0.303 \\ 

\bottomrule
\end{tabular}%
}
\caption{Supervised long-term multivariate forecasting. We present the average of all horizons predictions; detailed results are shown in the Appendix, Table~\ref{table:sup}. Results for GPT2(6)-adapter (GPT(6)-A), GPT2(6)-frozen (GLT(6)-F), GPT2(0), DLinear, PatchTST, TimesNet, FEDformer and Autoformer were taken from \cite{zhou2023one}, for ILI dataset, the results are taken from \cite{Yuqietal-2023-PatchTST} and \cite{wu2022timesnet}; ITransformer metrics are cited form its paper \cite{liu2023itransformer}. The best results are highlighted in \textbf{bold} separately for zero-shot and supervised approaches. Missing values, denoted by "--", correspond to the cases where the original paper does not provide metrics for the corresponding datasets. A double line separates the columns with zero-shot approaches based on synthetics (Ours and ForecastPFN) and supervised models.}
\label{table:sup_short}
\end{table*}

The results are presented in Table~\ref{table:sup_short}. The main expected takeaway is the poor performance of zero-shot methods for large datasets and large horizons compared to supervised approaches. The score for the SOTA supervised model is more than twice better than the closest true zero-shot approach trained on synthetic data: 0.386 on average MAE for PatchTST and near 0.820 for PFNs. Nevertheless, promising results are shown on a small ILI dataset a with relativity large horizon where our approach is on par with the main supervised methods and come close to the SOTA result. In comparison with ForecastPFN, our approach shows better performance on 5 datasets out of 8 for small horizon length (mostly without 720 due to a specific approach to sample synthetic datasets with the trend during training).

\begin{table*}[!htbt]
\resizebox{\textwidth}{!}{%
\begin{tabular}{@{}lcc|ll|ll|ll|ll|ll|lc|ll@{}}
\toprule
\multicolumn{3}{c|}{} & \multicolumn{2}{c|}{\textbf{Ours}} & \multicolumn{2}{c|}{\textbf{ForecastPFN}} & \multicolumn{2}{c|}{\textbf{GPT2(6)}} & \multicolumn{2}{c|}{\textbf{GPT2(0)}} & \multicolumn{2}{c|}{\textbf{DLinear}} & \multicolumn{2}{c|}{\textbf{PatchTST}} & \multicolumn{2}{c}{\textbf{ITransformer}} \\ 
\multicolumn{1}{l}{\textbf{Dataset}} & \multicolumn{1}{c}{\textbf{Horizon}} & \textbf{Train Size} & \textbf{MSE} & \textbf{MAE} & \textbf{MSE} & \textbf{MAE} & \textbf{MSE} & \textbf{MAE} & \textbf{MSE} & \textbf{MAE} & \textbf{MSE} & \textbf{MAE} & \textbf{MSE} & \textbf{MAE} & \textbf{MSE} & \textbf{MAE} \\ \midrule
\multicolumn{1}{l|}{\multirow{8}{*}{ETTh1}} & \multicolumn{1}{c|}{\multirow{8}{*}{96}} & 24 & 5.136 & 1.433 & 1.880* & 0.905* & \multicolumn{1}{c}{-} & \multicolumn{1}{c|}{-} & \multicolumn{1}{c}{-} & \multicolumn{1}{c|}{-} & \multicolumn{1}{c}{-} & \multicolumn{1}{c|}{-} & \multicolumn{1}{c}{-} & - & \multicolumn{1}{c}{-} & \multicolumn{1}{c}{-} \\
\multicolumn{1}{l|}{} & \multicolumn{1}{l|}{} & 48 & 2.443 & 0.999 & 1.923* & 0.930* & \multicolumn{1}{c}{-} & \multicolumn{1}{c|}{-} & \multicolumn{1}{c}{-} & \multicolumn{1}{c|}{-} & \multicolumn{1}{c}{-} & \multicolumn{1}{c|}{-} & \multicolumn{1}{c}{-} & - & \multicolumn{1}{c}{-} & \multicolumn{1}{c}{-} \\
\multicolumn{1}{l|}{} & \multicolumn{1}{l|}{} & 96 & 1.645 & 0.851 & 1.952* & 0.952* & \multicolumn{1}{c}{-} & \multicolumn{1}{c|}{-} & \multicolumn{1}{c}{-} & \multicolumn{1}{c|}{-} & \multicolumn{1}{c}{-} & \multicolumn{1}{c|}{-} & \multicolumn{1}{c}{-} & - & \multicolumn{1}{c}{-} & \multicolumn{1}{c}{-} \\
\multicolumn{1}{l|}{} & \multicolumn{1}{l|}{} & 192 & 0.899 & 0.672 & 1.944* & 0.951* & \multicolumn{1}{c}{-} & \multicolumn{1}{c|}{-} & \multicolumn{1}{c}{-} & \multicolumn{1}{c|}{-} & \multicolumn{1}{c}{-} & \multicolumn{1}{c|}{-} & \multicolumn{1}{c}{-} & - & 0.905 & 0.635 \\
\multicolumn{1}{l|}{} & \multicolumn{1}{l|}{} & 384 & 0.778 & 0.627 &  \multicolumn{1}{l}{\textcolor{gray}{1.944*}} &  \multicolumn{1}{l|}{\textcolor{gray}{0.951*}} & \multicolumn{1}{c}{-} & \multicolumn{1}{c|}{-} & \multicolumn{1}{c}{-} & \multicolumn{1}{c|}{-} & \multicolumn{1}{c}{-} & \multicolumn{1}{c|}{-} & \multicolumn{1}{c}{-} & - & 0.609 & 0.526 \\
\multicolumn{1}{l|}{} & \multicolumn{1}{l|}{} & 768 & 0.735 & 0.615 & \multicolumn{1}{l}{\textcolor{gray}{1.944*}} & \multicolumn{1}{l|}{\textcolor{gray}{0.951*}} & 0.410 & 0.429 & 0.562 & 0.516 & 0.451 & 0.445 & 0.467 & \multicolumn{1}{l|}{0.468} & 0.470 & 0.459 \\
\multicolumn{1}{l|}{} & \multicolumn{1}{l|}{} & 1536 & \multicolumn{1}{l}{\textcolor{gray}{0.735}} & \multicolumn{1}{l|}{\textcolor{gray}{0.615}} & \multicolumn{1}{l}{\textcolor{gray}{1.944*}} & \multicolumn{1}{l|}{\textcolor{gray}{0.951*}} & 0.406 & 0.422 & 0.507 & 0.483 & 0.422 & 0.437 & 0.387 & \multicolumn{1}{l|}{0.417} & 0.421 & 0.428 \\
\multicolumn{1}{l|}{} & \multicolumn{1}{l|}{} & 3072 & \multicolumn{1}{l}{\textcolor{gray}{0.735}} & \multicolumn{1}{l|}{\textcolor{gray}{0.615}} & \multicolumn{1}{l}{\textcolor{gray}{1.944*}} & \multicolumn{1}{l|}{\textcolor{gray}{0.951*}} & 0.393 & 0.410 & 0.552 & 0.507 & 0.397 & 0.415 & 0.378 & \multicolumn{1}{l|}{0.406} & 0.397 & 0.411 \\ \midrule
\multicolumn{1}{l|}{\multirow{8}{*}{ETTh2}} & \multicolumn{1}{c|}{\multirow{8}{*}{96}} & 24 & 1.079 & 0.675 & 0.333* & 0.397* & \multicolumn{1}{c}{-} & \multicolumn{1}{c|}{-} & \multicolumn{1}{c}{-} & \multicolumn{1}{c|}{-} & \multicolumn{1}{c}{-} & \multicolumn{1}{c|}{-} & \multicolumn{1}{c}{-} & - & \multicolumn{1}{c}{-} & \multicolumn{1}{c}{-} \\
\multicolumn{1}{l|}{} & \multicolumn{1}{l|}{} & 48 & 1.358 & 0.669 & 0.380* & 0.419* & \multicolumn{1}{c}{-} & \multicolumn{1}{c|}{-} & \multicolumn{1}{c}{-} & \multicolumn{1}{c|}{-} & \multicolumn{1}{c}{-} & \multicolumn{1}{c|}{-} & \multicolumn{1}{c}{-} & - & \multicolumn{1}{c}{-} & \multicolumn{1}{c}{-} \\
\multicolumn{1}{l|}{} & \multicolumn{1}{l|}{} & 96 & 1.721 & 0.721 & 0.416* & 0.447* & \multicolumn{1}{c}{-} & \multicolumn{1}{c|}{-} & \multicolumn{1}{c}{-} & \multicolumn{1}{c|}{-} & \multicolumn{1}{c}{-} & \multicolumn{1}{c|}{-} & \multicolumn{1}{c}{-} & - & \multicolumn{1}{c}{-} & \multicolumn{1}{c}{-} \\
\multicolumn{1}{l|}{} & \multicolumn{1}{l|}{} & 192 & 1.004 & 0.625 & 0.419* & 0.449* & \multicolumn{1}{c}{-} & \multicolumn{1}{c|}{-} & \multicolumn{1}{c}{-} & \multicolumn{1}{c|}{-} & \multicolumn{1}{c}{-} & \multicolumn{1}{c|}{-} & \multicolumn{1}{c}{-} & - & 0.416 & 0.426 \\
\multicolumn{1}{l|}{} & \multicolumn{1}{l|}{} & 384 & 0.827 & 0.598 & \multicolumn{1}{l}{\textcolor{gray}{0.419*}} & \multicolumn{1}{l|}{\textcolor{gray}{0.449*}} & \multicolumn{1}{c}{-} & \multicolumn{1}{c|}{-} & \multicolumn{1}{c}{-} & \multicolumn{1}{c|}{-} & \multicolumn{1}{c}{-} & \multicolumn{1}{c|}{-} & \multicolumn{1}{c}{-} & - & 0.385 & 0.409 \\
\multicolumn{1}{l|}{} & \multicolumn{1}{l|}{} & 768 & 0.925 & 0.64 & \multicolumn{1}{l}{\textcolor{gray}{0.419*}} & \multicolumn{1}{l|}{\textcolor{gray}{0.449*}} & 0.402 & 0.408 & 0.575 & 0.499 & 0.389 & 0.420 & 0.357 & \multicolumn{1}{l|}{0.400} & 0.346 & 0.385 \\
\multicolumn{1}{l|}{} & \multicolumn{1}{l|}{} & 1536 & \multicolumn{1}{l}{\textcolor{gray}{0.925}} & \multicolumn{1}{l|}{\textcolor{gray}{0.64}} & \multicolumn{1}{l}{\textcolor{gray}{0.419*}} & \multicolumn{1}{l|}{\textcolor{gray}{0.449*}} & 0.311 & 0.353 & 0.376 & 0.402 & 0.300 & 0.360 & 0.291 & \multicolumn{1}{l|}{0.349} & 0.315 & 0.363 \\
\multicolumn{1}{l|}{} & \multicolumn{1}{l|}{} & 3072 & \multicolumn{1}{l}{\textcolor{gray}{0.925}} & \multicolumn{1}{l|}{\textcolor{gray}{0.64}} & \multicolumn{1}{l}{\textcolor{gray}{0.419*}} & \multicolumn{1}{l|}{\textcolor{gray}{0.449*}} & 0.298 & 0.351 & 0.429 & 0.438 & 0.288 & 0.357 & 0.292 & \multicolumn{1}{l|}{0.353} & 0.321 & 0.368 \\ \midrule
\multicolumn{1}{l|}{\multirow{8}{*}{ILI}} & \multicolumn{1}{c|}{\multirow{8}{*}{24}} & 52 & 22.828 & 3.641 & 4.055 & 1.502 & \multicolumn{1}{c}{-} & \multicolumn{1}{c|}{-} & \multicolumn{1}{c}{-} & \multicolumn{1}{c|}{-} & \multicolumn{1}{c}{-} & \multicolumn{1}{c|}{-} & \multicolumn{1}{c}{-} & - & \multicolumn{1}{c}{-} & \multicolumn{1}{c}{-} \\
\multicolumn{1}{l|}{} & \multicolumn{1}{l|}{} & 104 & 3.495 & 1.288 & 4.098 & 1.502 & \multicolumn{1}{c}{-} & \multicolumn{1}{c|}{-} & \multicolumn{1}{c}{-} & \multicolumn{1}{c|}{-} & \multicolumn{1}{c}{-} & \multicolumn{1}{c|}{-} & \multicolumn{1}{c}{-} & - & \multicolumn{1}{c}{-} & \multicolumn{1}{c}{-} \\
\multicolumn{1}{l|}{} & \multicolumn{1}{l|}{} & 156 & 2.67 & 1.092 & \multicolumn{1}{l}{\textcolor{gray}{4.098}} & \multicolumn{1}{l|}{\textcolor{gray}{1.502}} & 2.427 & 1.079 & 2.348 & 1.092 & 2.675 & 1.213 & \multicolumn{1}{c}{-} & - & 4.553 & 1.556 \\
\multicolumn{1}{l|}{} & \multicolumn{1}{l|}{} & 208 & 2.521 & 1.067 & \multicolumn{1}{l}{\textcolor{gray}{4.098}} & \multicolumn{1}{l|}{\textcolor{gray}{1.502}} & 2.230 & 0.998 & 2.356 & 1.096 & 2.006 & 0.949 & 2.069 & \multicolumn{1}{l|}{0.980} & 4.080 & 1.462 \\
\multicolumn{1}{l|}{} & \multicolumn{1}{l|}{} & 260 & 2.536 & 1.091 & \multicolumn{1}{l}{\textcolor{gray}{4.098}} & \multicolumn{1}{l|}{\textcolor{gray}{1.502}} & 2.134 & 0.971 & 2.753 & 1.144 & 2.128 & 0.973 & 2.032 & \multicolumn{1}{l|}{0.956} & 3.466 & 1.333 \\
\multicolumn{1}{l|}{} & \multicolumn{1}{l|}{} & 312 & 2.846 & 1.167 & \multicolumn{1}{l}{\textcolor{gray}{4.098}} & \multicolumn{1}{l|}{\textcolor{gray}{1.502}} & 2.282 & 1.049 & 2.506 & 1.110 & 1.972 & 0.941 & 2.048 & \multicolumn{1}{l|}{0.961} & 3.097 & 1.244 \\
\multicolumn{1}{l|}{} & \multicolumn{1}{l|}{} & 364 & 2.859 & 1.177 & \multicolumn{1}{l}{\textcolor{gray}{4.098}} & \multicolumn{1}{l|}{\textcolor{gray}{1.502}} & 2.101 & 0.932 & 2.224 & 0.968 & 2.245 & 1.005 & 1.998 & \multicolumn{1}{l|}{0.962} & 3.084 & 1.243 \\
\multicolumn{1}{l|}{} & \multicolumn{1}{l|}{} & 416 & 2.833 & 1.174 & \multicolumn{1}{l}{\textcolor{gray}{4.098}} & \multicolumn{1}{l|}{\textcolor{gray}{1.502}} & 2.093 & 0.975 & 2.223 & 1.043 & 2.359 & 1.109 & 2.070 & \multicolumn{1}{l|}{0.939} & 2.900 & 1.194 \\ \bottomrule
\end{tabular}%
}
\caption{Few-shot multivariate forecasting numerical results. Missing values, denoted by "--", correspond to the cases where there are too few observations for supervised tuning or prediction horizon overflow for zero-shot approaches. Gray values denote instances of prediction horizon overflow in zero-shot approaches. The input is truncated in such scenarios to adhere to the maximum model input size. Values, denoted by "*", correspond to the cases where there are missing values in models' predictions (even though they are few in number) and metrics are derived without taking missing values into account.}
\label{table:fewshot}
\end{table*}

\paragraph{Few-Shot Learning}

Predicting a time series within a few-shot setting represents a crucial task. In order to grasp the impact of the training sample size on outcomes produced by supervised methods, we intentionally constrained the number of data points within the training sample. We conduct the experiments on GPT2, DLinear, PatchTST, and ITransformer models, utilizing the ETTh1, ETTh2, and ILI datasets. This deliberate limitation allows us to systematically evaluate and comprehend the influence of varying training sample sizes on the predictive capabilities of these models across diverse temporal datasets.

We partly reproduced the methodology of the few-shot learning task from \cite{zhou2023one}. However, our strategy for a few-shot task involves retaining the most recent values of the training segment immediately adjacent to the validation part. Also, we do not fix the size of the training segment in percents; we adjust its length, proportionally aligning with the period of the shortest seasonal pattern inherent to the corresponding frequency (e.g., 24 for hourly data). Here, the term "seasonality period" denotes the duration of the time interval within which specific repetitive patterns occur in the time series data. This approach allows us to eliminate length dependency for series extremely varying across datasets. As an illustration, 5\% for ETTh1 training subset is 432 points, whereas the corresponding value for ILI is 34 points. Such a significant difference in the number of points in train part can seriously affect forecasting performance and make it difficult to compare (see details in Appendix~\ref{app:few_shot_details}).

Additionally, we consider the same datasets and the same procedure of validation as in the preceding Long-Term Multivariate forecasting (and subsequent Zero-shot Learning) tasks. This extensive exploration aims to clarify the impact of quantity (many or a few) and quality (real or synthetic) data on the performance of models. We seek to distinguish the quality achievable with abundant or scarce data across different model types, including those that necessitate no training and represent zero-shot models trained on synthetic data. This investigation raises the question of whether having an almost infinite amount of synthetic data provides advantages or if a modest quantity of real data remains sufficient.

We train the models to predict the minimum horizon using in the Long-Term Multivariate forecasting task specific to each dataset (24 for ILI, 96 for the others). The look-back window lengths for models set at 104 on ILI and 336 on ETTh for GPT2(6), GPT2(0), DLinear, 336 on ILI and 148 on ETTh for PatchTST, and 96 on both for ITransformer, aligning with established conventions from prior research papers.
For GPT models, we employed the hyperparameters provided in \cite{zhou2023one}, while for all other models, the hyperparameters were derived from TimesNet \cite{wu2023timesnet}.

To maintain a balance between conducting a manageable number of experiments and ensuring visibility in the dynamic of results while the training sample size is enhancing, we consider 8 options for the parameter "train budget" ($B$) for each dataset. B indicates the number of available training points out of all possible ones (which are feasible in long-term forecasting tasks). Except for the shortest ILI dataset, we selected powers of two multiplied by the estimated period length to encompass a sufficiently broad range of interest for the B. Consequently, for the ETTh datasets where period $P = 24$, $B \in \{24, 48, 96, 192, 384, 768, 1536, 3072\}$, while for ILI with $P=52$, $B \in \{52, 104, 156, 208, 260, 312, 364, 416\}$. 

The results of few-shot learning experiment across various training sample sizes are presented in Table~\ref{table:fewshot}. Notably, the DLinear, PatchTST, and ITransformer models are supervised, necessitating an adequate number of samples in the training segment. To utilize these models, the length of the series must be at least $L + H$, where $L$ denotes the look-back window of the model and $H$ signifies the prediction horizon. Consequently, it becomes evident that with small training sample sizes, there are insufficient samples available for training the model and getting any results.

At the same time, our zero-shot model and ForecastPFN, though not requiring pre-training, operate with a constrained context length. Consequently, upon reaching a specific threshold in the size of the training segment of the time series, the training and application outcomes of the model cease to exhibit significant changes, as aptly illustrated in the table. It is important that reliable reproduction of results for ForecastPFN was challenging due to the model predicting missing values when faced with periods featuring unchanging values in the time series.

In cases where the training segment is restricted to a notably small size, our zero-shot model exhibits inferior performance compared to ForecastPFN. However, as the number of observations within the training  part increases, the quality of predictions generated by our zero-shot model improves, and the performance of ForecastPFN either remains stagnant or even diminishes. Surprisingly, once the number of samples reaches a sufficient threshold to train and infer supervised models, their performance almost immediately demonstrates competence.

Table~\ref{table:zero_few_sup} plays a significant role in deriving conclusions from our analysis. It illustrates a substantial improvement in model performance even with a minimal number of data points compared to virtually any zero-shot approach. Notably, the relationship is most unequivocal for transformer-based architectures and longer series, underscoring the remarkable capacity of the attention mechanism to generalize across a vast number of points. However, the MLP-based DLinear demonstrates commendable performance, particularly evident in the short ILI dataset, albeit lacking comparable effectiveness in few-shot scenarios compared to other approaches.

Thus, we assume that our zero-shot model performs well when afforded a reasonable number of data points to detect patterns in the series' behavior, yet remains within a size small enough to train a supervised model. Nevertheless, in cases where we have very limited train data but sufficient only to provide at least a few samples for the supervised model's training, then the final performance is likely to surpass that of zero-shot models, at least those trained on synthetic data. Additionally, the quality is expected to converge rather quickly towards the levels observed in the long-term forecasting task.

\paragraph{Zero-shot Forecasting}

\begin{table*}[!htbt]
\resizebox{\textwidth}{!}{%
\begin{tabular}
{c|c|rrrr|rrrr|rrrr|rrrr} \toprule
\multirow{2}{*}{\textbf{Models}} &  & \multicolumn{4}{c|}{\textbf{Weather}} & \multicolumn{4}{c|}{\textbf{ETTh1}} & \multicolumn{4}{c|}{\textbf{ETTh2}} & \multicolumn{4}{c}{\textbf{ETTm1}} 
\\ 
 & H & \multicolumn{1}{c}{\textbf{96}} & \multicolumn{1}{c}{\textbf{192}} & \multicolumn{1}{c}{\textbf{336}} & \multicolumn{1}{c|}{\textbf{720}} & \multicolumn{1}{c}{\textbf{96}} & \multicolumn{1}{c}{\textbf{192}} & \multicolumn{1}{c}{\textbf{336}} & \multicolumn{1}{c|}{\textbf{720}} & \multicolumn{1}{c}{\textbf{96}} & \multicolumn{1}{c}{\textbf{192}} & \multicolumn{1}{c}{\textbf{336}} & \multicolumn{1}{c|}{\textbf{720}} & \multicolumn{1}{c}{\textbf{96}} & \multicolumn{1}{c}{\textbf{192}} & \multicolumn{1}{c}{\textbf{336}} & \multicolumn{1}{c}{\textbf{720}} 
 \\ \midrule
\multirow{2}{*}{\textbf{Ours}} & MSE & 0.379 & 0.772 & 1.643 & 6.451 & \textbf{0.828} & \textbf{1.280} & 2.299 & 4.999 & 0.827 & 1.828 & 3.453 & 11.443 & \textbf{0.763} & 1.401 & 2.706 & 7.549 
\\ 
 & MAE & 0.353 & 0.491 & 0.688 & 1.131 & \textbf{0.639} & \textbf{0.794} & 1.036 & 1.471 & 0.583 & 0.816 & 1.078 & 1.575 & \textbf{0.571} & 0.752 & 1.019 & 1.547 
 \\ \midrule
\multirow{2}{*}{\textbf{FPFN}} & MSE & \textbf{0.251} & \textbf{0.291} & \textbf{0.347} & \textbf{0.422} & 1.856 & 1.847 & \textbf{1.877} & \textbf{2.025} & \textbf{0.361} & \textbf{0.402} & \textbf{0.410} & \textbf{0.451} & 1.266 & \textbf{1.304} & \textbf{1.385} & \textbf{1.400} 
\\ 
 & MAE & \textbf{0.281} & \textbf{0.308} & \textbf{0.343} & \textbf{0.390} & 0.915 & 0.926 & \textbf{0.941} & \textbf{0.993} & \textbf{0.410} & \textbf{0.438} & \textbf{0.444} & \textbf{0.471} & 0.697 & \textbf{0.714} & \textbf{0.744} & \textbf{0.762} 
 \\
 \midrule
\multirow{2}{*}{\textbf{Models}} &  & 
\multicolumn{4}{c|}{\textbf{ETTm2}} & \multicolumn{4}{c|}{\textbf{ECL}} & \multicolumn{4}{c|}{\textbf{ILI}} & \multicolumn{4}{c}{\textbf{Traffic}} \\ 
 & H & 
 \multicolumn{1}{c}{\textbf{96}} & \multicolumn{1}{c}{\textbf{192}} & \multicolumn{1}{c}{\textbf{336}} & \multicolumn{1}{c|}{\textbf{720}} & \multicolumn{1}{c}{\textbf{96}} & \multicolumn{1}{c}{\textbf{192}} & \multicolumn{1}{c}{\textbf{336}} & \multicolumn{1}{c|}{\textbf{720}} & \multicolumn{1}{c}{\textbf{24}} & \multicolumn{1}{c}{\textbf{36}} & \multicolumn{1}{c}{\textbf{48}} & \multicolumn{1}{c|}{\textbf{60}} & \multicolumn{1}{c}{\textbf{96}} & \multicolumn{1}{c}{\textbf{192}} & \multicolumn{1}{c}{\textbf{336}} & \multicolumn{1}{c}{\textbf{720}} \\ \midrule
\multirow{2}{*}{\textbf{Ours}} & MSE & 
0.500 & 1.093 & 2.465 & 10.480 & \textbf{0.575} & \textbf{0.886} & \textbf{1.600} & 4.014 & \textbf{2.714} & \textbf{2.531} & \textbf{2.314} & \textbf{2.135} & \textbf{1.223} & \textbf{1.454} & \textbf{2.011} & 3.349 \\ 
 & MAE & 
 0.411 & 0.566 & 0.796 & 1.280 & \textbf{0.534} & \textbf{0.622} & \textbf{0.776} & 1.098 & \textbf{1.082} & \textbf{1.028} & \textbf{0.981} & \textbf{0.968} & \textbf{0.700} & \textbf{0.766} & \textbf{0.897} & \textbf{1.146} \\ \midrule
\multirow{2}{*}{\textbf{FPFN}} & MSE & 
\textbf{0.188} & \textbf{0.213} & \textbf{0.246} & \textbf{0.301} & 1.629 & 1.600 & 1.613 & \textbf{1.651} & 5.230 & 4.748 & 4.318 & 4.316 & 3.059 & 2.980 & 2.994 & \textbf{2.951} \\ 
 & MAE & 
 \textbf{0.289} & \textbf{0.307} & \textbf{0.328} & \textbf{0.361} & 0.976 & 0.969 & 0.974 & \textbf{0.988} & 1.743 & 1.617 & 1.521 & 1.513 & 1.230 & 1.214 & 1.218 & 1.211 \\
 
\bottomrule
\end{tabular}%
}
\caption{Zero-shot long-term multivariate forecasting: Comparison between ForecastPFN (FPFN) and our approach for different horizons. Our approach demonstrates superior suitability for short forecasting horizons and, more broadly, for short time series datasets (ILI, ECL, Traffic).}
\label{table:long_zero_shot}
\end{table*}

\begin{table*}[h!]
\centering
\resizebox{0.75\textwidth}{!}{%
\begin{tabular}{lc|ccc|ccc|ccc}
\toprule
\textbf{} & \multicolumn{1}{l|}{\textbf{Models}} & \multicolumn{3}{c|}{\textbf{GPT2(6)}} & \multicolumn{3}{c|}{\textbf{DLinear}} & \multicolumn{3}{c}{\textbf{PatchTST}} \\
\textbf{Dataset} & \multicolumn{1}{l|}{\textbf{Budget}} & \multicolumn{1}{l}{\textbf{MSE}} & \multicolumn{1}{l}{\textbf{MAE}} & \multicolumn{1}{l|}{\textbf{SMAPE}} & \multicolumn{1}{l}{\textbf{MSE}} & \multicolumn{1}{l}{\textbf{MAE}} & \multicolumn{1}{l|}{\textbf{SMAPE}} & \multicolumn{1}{l}{\textbf{MSE}} & \multicolumn{1}{l}{\textbf{MAE}} & \multicolumn{1}{l}{\textbf{SMAPE}} \\
\midrule
\multirow{10}{*}{\textbf{Weather}} & \multicolumn{1}{c|}{ZS} & 0.0663 & 0.08 & 25.235 & 0.072 & 0.074 & 24.152 & 0.142 & 0.205 & 54.151 \\
 & 384 & 0.084 & 0.132 & 39.356 & 0.204 & 0.31 & 81.875 & 0.085 & 0.127 & 38.29 \\
 & 768 & 0.069 & 0.101 & 31.244 & 0.124 & 0.211 & 61.175 & 0.066 & 0.092 & 30.184 \\
 & 1536 & 0.066 & 0.089 & 28.257 & 0.088 & 0.153 & 46.503 & 0.06 & 0.078 & 26.559 \\
 & 3072 & 0.066 & 0.091 & 28.934 & 0.076 & 0.132 & 40.953 & 0.06 & 0.08 & 26.986 \\
 & 6144 & 0.064 & 0.086 & 27.286 & 0.066 & 0.109 & 36.105 & 0.059 & 0.076 & 25.957 \\
 & \multicolumn{1}{c|}{FS} & 0.061 & 0.083 & 26.696 & 0.063 & 0.063 & 29.321 & 0.073 & 0.106 & 32.681 \\
\midrule

\multirow{10}{*}{\textbf{ETTh1}} & \multicolumn{1}{c|}{ZS} & 0.6727 & 0.525 & 85.258 & 0.838 & 0.527 & 84.32 & 0.681 & 0.533 & 96.370 \\
 & 192 & 0.36 & 0.395 & 73.667 & 0.385 & 0.416 & 77.865 & \multicolumn{1}{c}{-} & \multicolumn{1}{c}{-} & \multicolumn{1}{c}{-} \\
 & 384 & 0.273 & 0.341 & 66.56 & 0.285 & 0.351 & 68.431 & 0.406 & 0.428 & 79.118 \\
 & 768 & 0.24 & 0.319 & 64.636 & 0.248 & 0.319 & 65.262 & 0.294 & 0.358 & 70.151 \\
 & 1536 & 0.238 & 0.317 & 63.986 & 0.239 & 0.312 & 64.14 & 0.261 & 0.334 & 67.227 \\
 & 3072 & 0.222 & 0.303 & 62.94 & 0.226 & 0.3 & 62.316 & 0.235 & 0.311 & 63.986 \\
  & \multicolumn{1}{c|}{FS} & 0.221 & 0.301 & 62.289 & 0.239 & 0.239 & 63.406 & 0.256 & 0.325 & 65.482 \\
 \midrule

 \multirow{10}{*}{\textbf{ETTh2}} & \multicolumn{1}{c|}{ZS} & 0.1473 & 0.254 & 45.334 & 0.167 & 0.254 & 43.241 & 0.241 & 0.325 & 54.106 \\
 & 192 & 0.156 & 0.259 & 45.305 & 0.189 & 0.299 & 48.892 & \multicolumn{1}{c}{-} & \multicolumn{1}{c}{-} & \multicolumn{1}{c}{-} \\
 & 384 & 0.114 & 0.220 & 39.860 & 0.112 & 0.217 & 39.366 & 0.196 & 0.292 & 49.296 \\
 & 768 & 0.113 & 0.214 & 38.498 & 0.248 & 0.319 & 65.262 & 0.121 & 0.226 & 41.147 \\
 & 1536 & 0.109 & 0.214 & 39.271 & 0.239 & 0.312 & 64.140 & 0.109 & 0.212 & 39.406 \\
 & 3072 & 0.116 & 0.216 & 39.301 & 0.226 & 0.300 & 62.316 & 0.111 & 0.213 & 39.065 \\
 & \multicolumn{1}{c|}{FS} & 0.108 & 0.211 & 38.260 & 0.108 & 0.212 & 38.031 & 0.111 & 0.217 & 40.128 \\ 
\midrule
 
 \multirow{10}{*}{\textbf{ILI}} & \multicolumn{1}{c|}{ZS} & 1.5458 & 0.761 & 64.522 & 1.414 & 0.667 & 58.172 & 3.699 & 1.415 & 109.215 \\
 & 156 & 2.427 & 1.079 & 84.110 & 2.132 & 0.948 & 76.983 & \multicolumn{1}{c}{-} & \multicolumn{1}{c}{-} & \multicolumn{1}{c}{-} \\
 & 208 & 2.230 & 0.998 & 73.720 & 1.808 & 0.858 & 70.274 & 2.298 & 0.990 & 74.915 \\
 & 260 & 2.134 & 0.971 & 72.881 & 1.642 & 0.838 & 67.927 & 2.288 & 0.993 & 75.516 \\
 & 312 & 2.282 & 1.049 & 79.839 & 1.186 & 0.712 & 62.827 & 2.056 & 0.956 & 73.864 \\
 & 364 & 2.101 & 0.932 & 70.381 & 1.048 & 0.673 & 61.506 & 1.348 & 0.745 & 62.716 \\
 & 416 & 2.093 & 0.975 & 75.978 & 1.039 & 0.695 & 63.273 & 1.274 & 0.772 & 70.764 \\
 & \multicolumn{1}{c|}{FS} & 1.034 & 0.639 & 60.097 & 0.895 & 0.586 & 47.126 & 1.034 & 0.640 & 61.243 \\
 \bottomrule
 
\end{tabular}%
}
\caption{Model performance dependence on the model budget - from zero-shot (ZS) to full train size provided to model (FS). In this experiment, the horizon was set to 6, and the look-back window length for GPT2(6) is 104, 104 for DLinear, and 148 for PatchTST. Due to the large look-back window length for PatchTST, some results for a small train budget are unavailable. Zero-shot results were provided for pretraining on the M4 Weekly dataset because it was one of the best source datasets (see Figure~\ref{app:win_rate} in Appendix). As observed, for a transformer-based architecture, particularly evident in the case of PatchTST, a minimal number of data points proves sufficient, and the model's performance notably enhanced with a more significant number of points in the training budget compared to the zero-shot setup. Conversely, the dependence could be more complex for an MLP-based architecture such as DLinear.}
\label{table:zero_few_sup}
\end{table*}

There is a lack of literature addressing zero-shot setup in time series, where the objective is to perform prediction tasks on new series without any prior training on them. One approach to cope with this challenge involves training the model on a source dataset and subsequently inference it to a target dataset. A similar approach is demonstrated in \cite{zhou2023one}, where the authors showcase results on "paired" datasets characterized by similar nature and granularity. We aim to assess the efficiency of zero-shot learning without relying on pairwise datasets and to compare the obtained results with those achieved by models specifically designed for zero-shot scenarios. Our objective is twofold: firstly, to show the extent to which the choice of the "optimal" source dataset influences the outcome, and secondly, to conduct a comparative analysis between the results of supervised models and zero-shot models trained on synthetic data. This investigation aims to address the fundamental question of whether data generation is essential or if superior outcomes can be achieved by leveraging real-world data. 

To conduct the experiment, we trained GPT2, PatchTST, and DLinear on  M3 and M4 datasets with varying granularities. We evaluate their performance on Weather, Electricity, Traffic, ILI, and the four ETT datasets. For models trained on the M3 dataset, we acquired their performance results on the M4 dataset. Additionally, we evaluate the zero-shot models ForecastPFN and our zero-shot model. GPT2(0) was not included in this task as it consistently underperformed GPT2(6) in all prior scenarios. ITransformer was also excluded from consideration, as our methodology aligns with \cite{zhou2023one}, where time series are treated as independent entities and examined separately. In contrast, the ITransformer is designed for a multivariate context, assuming the consideration of time series in a multivariate setting.

\begin{wrapfigure}{l}{0.3\textwidth}
  \centering
    \resizebox{0.3\columnwidth}{!}{%
    \begin{tabular}{l|rrr}
    \toprule
    \textbf{Source} & \multicolumn{1}{l}{\textbf{MSE}} & \multicolumn{1}{l}{\textbf{MAE}} & \multicolumn{1}{l}{\textbf{SMAPE}} \\
    \midrule
    M3 Monthly & 4.271 & 4.343 & 4.243 \\
    M3 Other & 4.729 & 4.729 & 4.900 \\
    M3 Quarterly & 5.171 & 5.257 & 5.171 \\
    M3 Yearly & 5.671 & 5.657 & 5.643 \\
    M4 Daily & 4.638 & 3.950 & \textbf{3.825} \\
    M4 Hourly & 3.975 & 4.050 & 4.325 \\
    M4 Monthly & \textbf{3.525} & \textbf{3.850} & 4.325 \\
    M4 Quarterly & 4.750 & 5.000 & 4.650 \\
    M4 Weekly & 4.650 & 4.550 & 4.650 \\
    M4 Yearly & 6.238 & 6.125 & 5.800 \\
    \bottomrule
    \end{tabular}}%
    \caption{The source dataset rank averaged within target dataset and model. The best source dataset for a given metric is highlighted in \textbf{bold}.}
    \label{table:zeroshot_rank}
\end{wrapfigure}

The models have been trained on source datasets within the same time horizons as described in the \cite{zhou2023one} framework. However, there are several distinctions in our vision of experiments. Firstly, predictions for target datasets were limited to 6, a minimum between all typical horizons for datasets. That facilitates a comparative analysis to determine the impact of source dataset selection on the quality of predictions obtained for the target datasets. Secondly, we use all available points in the training part of the source rows instead of $10$, such as in the original paper, and refrain from discarding the last incomplete batch, both in the testing phase and during training. Another distinction lies in our utilizing of Mean Squared Error (MSE) instead of Symmetric Mean Absolute Percentage Error (SMAPE) for the loss calculation. Finally, we have another procedure of validation while testing on the M4 dataset. In contrast to the approach taken by the original paper's authors, we employ the validation methodology adopted in the M4 competition. Specifically, we differ from the standard benchmark validation, where validation occurs through a sliding window on a test segment of data. In our case, we predict only once for the horizon points, essentially addressing the time series prediction problem by solving it once for the designated horizon points ahead. For other datasets, the validation procedure remained the same.

Due to the small sizes of time series in the source datasets, we decided to constrain the look-back window to $L=104$ for GPT2(6) and DLinear and to $L=148$ for PatchTST in all scenarios. We employed the implementations and hyperparameters provided in \cite{zhou2023one} for the GPT(6) model, and \cite{Yuqietal-2023-PatchTST} for PatchTST and DLinear. In instances where models had not been previously executed for specific datasets, we adopted hyperparameters from similar datasets for which such parameters have already been tested. Additionally, for some models, parameters were adjusted due to their excessively long training time. For example, in the case of PatchTST trained on M4 Daily, we took the opportunity to take the patch setting to reduce training time with minimal compromise in quality.

\begin{wrapfigure}{r}{0.4\textwidth}
    \centering
    \resizebox{0.4\columnwidth}{!}{%
    \begin{tabular}{l|rrrrr}
    \toprule
    \textbf{Source} & \multicolumn{1}{l}{\textbf{PatchTST}} & \multicolumn{1}{l}{\textbf{GPT2(6)}} & \multicolumn{1}{l}{\textbf{DLinear}} & \multicolumn{1}{l}{\textbf{FPFN}} & \multicolumn{1}{l}{\textbf{Ours}} \\
    \midrule
    M3 Monthly & 0 & \textbf{0.57} & 0 & 0.07 & 0.36 \\
    M3 Other & 0 & 0 & 0 & 0.29 & \textbf{0.71} \\
    M3 Quarterly & 0 & 0.07 & 0 & 0.29 & \textbf{0.64} \\
    M3 Yearly & 0 & 0 & 0 & 0.29 & \textbf{0.71} \\
    M4 Daily & 0 & 0 & \textbf{0.75} & 0 & 0.25 \\
    M4 Hourly & 0 & 0.12 & \textbf{0.75} & 0 & 0.12 \\
    M4 Monthly & 0 & \textbf{0.5} & 0.25 & 0 & 0.25 \\
    M4 Quarterly & 0 & 0 & \textbf{0.75} & 0 & 0.25 \\
    M4 Weekly & 0.12 & \textbf{0.38} & 0.25 & 0 & 0.25 \\
    M4 Yearly & 0 & 0.25 & \textbf{0.38} & 0 & \textbf{0.38}\\
    \bottomrule
    \end{tabular}}%
    \caption{The model wining rate inside the source dataset according to the MAE.
    The best model for a given source dataset is highlighted in \textbf{bold}. Our approach outperforms ForecastPFN (FPFN) for all granularities of M3 and M4 datasets.
    Results for MSE and SMAPE can be seen in Table \ref{app:win_rate} of Appendix. 
    }
    \label{table:zeroshot_wins_mae}
\end{wrapfigure}

In the process, we identified that the quality of the models is enhanced when the target data is transformed to align with the scale of the source dataset. We tested several various scaling methods on the validation set and described this experiment in \ref{sseq:zero_shot_scaling}. In the conclusive version, to obtain predictions for the datasets from prior tasks (that is, excluding M4) for all supervised models, with the exception of DLinear trained on the M3 dataset, the target dataset were scaled using StandardScaler method to align the target dataset's data with the source one.

Results of zero-shot learning are presented in \ref{table:zeroshot_average_gain}, \ref{table:zeroshot_rank} and \ref{table:zeroshot_wins_mae}. Raw detailed results are available in Tables~\ref{table:zeroshot_m3_bench}, \ref{table:zeroshot_m3_m4}, \ref{table:zeroshot_m4_bench} in Appendix. It is noticeable that the effectiveness of the model is highly dependent on the choice of the source dataset. For instance, a poorly chosen source dataset can result in a considerable weakening of model performance on the target dataset, as evidenced in the cases of M3 Yearly and M3 Quarterly. And in such cases, the superiority of our zero-shot method is clear. Conversely, certain datasets, such as specific granularities within M4, demonstrate the capability to achieve satisfactory performance even without the need for additional training of supervised models.

A complete horizon comparison for the zero-shot models is presented in Table~\ref{table:long_zero_shot}. As can be seen, for short horizons and short series like ILI, our approach is superior to FPFN.

Table~\ref{table:zeroshot_average_gain} further validates the source dataset's significant influence on models' performance in a zero-shot learning context. This data underscores the potential utility of synthetic data for zero-shot setups, particularly in scenarios where selecting a source dataset poses challenges. Notably, our zero-shot approach exhibits comparatively lower variability in performance when the source dataset is altered, thereby conferring an advantage in situations requiring consistent forecasting quality across different source datasets. Notably, our approach demonstrates superior performance compared to ForecastPFN (FPFN) across all granularities of the M3 source dataset, while Dlinear exhibits near-optimal performance for the M4 source dataset.

\begin{table*}[!ht]
\resizebox{\textwidth}{!}{%
\begin{tabular}{l|rrrrr|rrrrr|rrrrr}
\toprule
 & \multicolumn{5}{c|}{\textbf{MSE}} & \multicolumn{5}{c|}{\textbf{MAE}} & \multicolumn{5}{c}{\textbf{SMAPE}} \\ 
\textbf{Source} & \multicolumn{1}{l}{\textbf{PatchTST}} & \multicolumn{1}{l}{\textbf{GPT2(6)}} & \multicolumn{1}{l}{\textbf{DLinear}} & \multicolumn{1}{l}{\textbf{FPFN}} & \multicolumn{1}{l|}{\textbf{Ours}} & \multicolumn{1}{l}{\textbf{PatchTST}} & \multicolumn{1}{l}{\textbf{GPT2(6)}} & \multicolumn{1}{l}{\textbf{DLinear}} & \multicolumn{1}{l}{\textbf{FPFN}} & \multicolumn{1}{l|}{\textbf{Ours}} & \multicolumn{1}{l}{\textbf{PatchTST}} & \multicolumn{1}{l}{\textbf{GPT2(6)}} & \multicolumn{1}{l}{\textbf{DLinear}} & \multicolumn{1}{l}{\textbf{FPFN}} & \multicolumn{1}{l}{\textbf{Ours}} \\
\midrule
M3 Monthly & 4.304 & 1.971 & 3.134 & 2.954 & \textbf{1.204} & 1.599 & 0.649 & 1.514 & 1.069 & \textbf{0.570} & 1.400 & 0.518 & 1.015 & 0.800 & \textbf{0.370} \\
M3 Other & 4.747 & 5.761 & 3.380 & 2.954 & \textbf{1.204} & 1.856 & 2.081 & 1.505 & 1.069 & \textbf{0.570} & 1.461 & 1.621 & 1.386 & 0.800 & \textbf{0.370} \\
M3 Quarterly & 11.915 & 3.433 & 20.086 & 2.954 & \textbf{1.204} & 3.538 & 1.436 & 3.518 & 1.069 & \textbf{0.570} & 1.696 & 0.858 & 3.219 & 0.800 & \textbf{0.370} \\
M3 Yearly & 31.476 & 17.060 & 12.215 & 2.954 & \textbf{1.204} & 6.283 & 3.870 & 3.059 & 1.069 & \textbf{0.570} & 2.411 & 1.606 & 2.838 & 0.800 & \textbf{0.370} \\
M4 Daily & 3.541 & 1.198 & \textbf{0.968} & 3.291 & 1.400 & 1.479 & 0.421 & \textbf{0.298} & 1.275 & 0.801 & 0.911 & 0.206 & \textbf{0.150} & 0.868 & 0.529 \\
M4 Hourly & 1.966 & 0.889 & \textbf{0.559} & 3.291 & 1.400 & 1.066 & 0.524 & \textbf{0.341} & 1.275 & 0.801 & 0.820 & 0.325 & \textbf{0.219} & 0.868 & 0.529 \\
M4 Monthly & 2.356 & \textbf{0.780} & 0.879 & 3.291 & 1.400 & 1.165 & \textbf{0.356} & 0.433 & 1.275 & 0.801 & 0.843 & \textbf{0.238} & 0.345 & 0.868 & 0.529 \\
M4 Quarterly & 4.420 & 1.201 & \textbf{0.970} & 3.291 & 1.400 & 1.868 & 0.564 & \textbf{0.398} & 1.275 & 0.801 & 0.958 & 0.405 & \textbf{0.288} & 0.868 & 0.529 \\
M4 Weekly & 1.938 & 1.543 & 1.505 & 3.291 & \textbf{1.400} & 1.033 & 0.633 & \textbf{0.591} & 1.275 & 0.801 & 0.839 & 0.310 & \textbf{0.309} & 0.868 & 0.529 \\
M4 Yearly & 32.061 & \textbf{1.334} & 1.943 & 3.291 & 1.400 & 6.790 & \textbf{0.586} & 0.669 & 1.275 & 0.801 & 1.850 & 0.393 & \textbf{0.325} & 0.868 & 0.529 \\ \bottomrule
\end{tabular}%
}
\caption{Average relative metric from the best score among models and source datasets inside \textbf{target} dataset wins for SMAPE inside source dataset. The average relative metric quantifies the average percentage deviation in error from the best score, with smaller values indicating a lesser deviation from the optimal performance. This metric provides valuable insights into the robustness of various approaches across diverse source datasets. The best-performing model for each source dataset is denoted in \textbf{bold}.}
\label{table:zeroshot_average_gain}
\end{table*}

Our zero-shot forecasting experiment reveals the limitations of the widely used transferability approach discussed in the referenced papers. It underscores the critical role of selecting an appropriate source dataset and suggests the potential application of synthetic data in instances where source dataset selection is challenging.

\section{Conclusion}
\label{seq:conclusion}

Developing a supervised model devoted to short time series is only sometimes feasible, encouraging a foundation time series model capable of operating in a zero-shot setting to get on the stage. It is straightforward to train such a model on synthetic data since its generation can be controlled, and we can also set an unlimited horizon and history to allow for all the time-dependent characteristics, the extent of the diversification, or the quantity of data. However, our experiments demonstrate that even a limited number of real samples with possible misrepresentations is better than versatile and flexible artificial ones for model performance in zero-shot. In that case, the model trained in a supervised regime on data at our disposal tends to outperform zero-shot settings on synthetic data in the prevailing number of cases. To sum up, one should trust and leverage real-life time series instead of compelling synthetic ones in case of zero-shot and switch to a supervised setting if one has little real data.

Nevertheless, the controlled generation of synthetics can be helpful in some occasions. Firstly, in our experiments, we showed that in transferability setup, metrics on the target dataset substantially vary depending on the selected source dataset. This fact undermines the feasibility of consistently selecting the source dataset to maintain satisfactory model performance, a common strategy employed in numerous time series studies utilizing a zero-shot setup. Another valuable case for synthetic data is the possibility of getting reliable datasets for zero-shot multi-task learning, in which each sample will own a definite property connected with an anomaly or trend. Finally, in the instance of synthetic data, we are insured against leakage between the source dataset and the target dataset. Therefore, models trained on controlled synthetics can still be helpful as a baseline for honest comparisons.
\section{Limitations}
\label{seq:limitations}

One limitation of our work is that all findings are given for regular time series. Additionally, we focus only on the controllable, data-free generation of artificial examples, leaving data-dependent strategies behind. To reach a verdict on synthetic data, we must explore examples generated in more distinguished ways. 


\bibliographystyle{unsrtnat}

\begin{thebibliography}{50}
\providecommand{\natexlab}[1]{#1}
\providecommand{\url}[1]{\texttt{#1}}
\expandafter\ifx\csname urlstyle\endcsname\relax
  \providecommand{\doi}[1]{doi: #1}\else
  \providecommand{\doi}{doi: \begingroup \urlstyle{rm}\Url}\fi

\bibitem[Zhou et~al.(2021)Zhou, Zhang, Peng, Zhang, Li, Xiong, and Zhang]{zhou2021informer}
Haoyi Zhou, Shanghang Zhang, Jieqi Peng, Shuai Zhang, Jianxin Li, Hui Xiong, and Wancai Zhang.
\newblock Informer: Beyond efficient transformer for long sequence time-series forecasting.
\newblock In \emph{Proceedings of the AAAI conference on artificial intelligence}, volume~35, pages 11106--11115, 2021.

\bibitem[Wu et~al.(2021)Wu, Xu, Wang, and Long]{wu2021autoformer}
Haixu Wu, Jiehui Xu, Jianmin Wang, and Mingsheng Long.
\newblock Autoformer: Decomposition transformers with auto-correlation for long-term series forecasting.
\newblock \emph{Advances in Neural Information Processing Systems}, 34:\penalty0 22419--22430, 2021.

\bibitem[Zhou et~al.(2022)Zhou, Ma, Wen, Wang, Sun, and Jin]{zhou2022fedformer}
Tian Zhou, Ziqing Ma, Qingsong Wen, Xue Wang, Liang Sun, and Rong Jin.
\newblock Fedformer: Frequency enhanced decomposed transformer for long-term series forecasting.
\newblock In \emph{International Conference on Machine Learning}, pages 27268--27286. PMLR, 2022.

\bibitem[Smyl and Kuber(2016)]{smyl2016data}
Slawek Smyl and Karthik Kuber.
\newblock Data preprocessing and augmentation for multiple short time series forecasting with recurrent neural networks.
\newblock In \emph{36th international symposium on forecasting}, 2016.

\bibitem[Makridakis and Hibon(2000)]{makridakis2000m3}
Spyros Makridakis and Michele Hibon.
\newblock The m3-competition: results, conclusions and implications.
\newblock \emph{International journal of forecasting}, 16\penalty0 (4):\penalty0 451--476, 2000.

\bibitem[Box and Jenkins(1968)]{box1968some}
George~EP Box and Gwilym~M Jenkins.
\newblock Some recent advances in forecasting and control.
\newblock \emph{Journal of the Royal Statistical Society. Series C (Applied Statistics)}, 17\penalty0 (2):\penalty0 91--109, 1968.

\bibitem[Taylor and Letham(2018)]{taylor2018forecasting}
Sean~J Taylor and Benjamin Letham.
\newblock Forecasting at scale.
\newblock \emph{The American Statistician}, 72\penalty0 (1):\penalty0 37--45, 2018.

\bibitem[Shi et~al.(2020)Shi, Yin, Cai, Cichocki, Yokota, Chen, Yuan, and Zeng]{shi2020block}
Qiquan Shi, Jiaming Yin, Jiajun Cai, Andrzej Cichocki, Tatsuya Yokota, Lei Chen, Mingxuan Yuan, and Jia Zeng.
\newblock Block hankel tensor arima for multiple short time series forecasting.
\newblock In \emph{Proceedings of the AAAI Conference on Artificial Intelligence}, volume~34, pages 5758--5766, 2020.

\bibitem[Gasthaus et~al.(2019)Gasthaus, Benidis, Wang, Rangapuram, Salinas, Flunkert, and Januschowski]{gasthaus2019probabilistic}
Jan Gasthaus, Konstantinos Benidis, Yuyang Wang, Syama~Sundar Rangapuram, David Salinas, Valentin Flunkert, and Tim Januschowski.
\newblock Probabilistic forecasting with spline quantile function rnns.
\newblock In \emph{The 22nd international conference on artificial intelligence and statistics}, pages 1901--1910. PMLR, 2019.

\bibitem[Oreshkin et~al.(2019)Oreshkin, Carpov, Chapados, and Bengio]{oreshkin2019n}
Boris~N Oreshkin, Dmitri Carpov, Nicolas Chapados, and Yoshua Bengio.
\newblock N-beats: Neural basis expansion analysis for interpretable time series forecasting.
\newblock In \emph{International Conference on Learning Representations}, 2019.

\bibitem[Pham et~al.(2022)Pham, Liu, Sahoo, and Hoi]{pham2022learning}
Quang Pham, Chenghao Liu, Doyen Sahoo, and Steven Hoi.
\newblock Learning fast and slow for online time series forecasting.
\newblock In \emph{The Eleventh International Conference on Learning Representations}, 2022.

\bibitem[George et~al.(2023)George, Dey, Banerjee, Mukherjee, and Suri]{george2023online}
Arun~M George, Sounak Dey, Dighanchal Banerjee, Arijit Mukherjee, and Manan Suri.
\newblock Online time-series forecasting using spiking reservoir.
\newblock \emph{Neurocomputing}, 518:\penalty0 82--94, 2023.

\bibitem[Michael et~al.(2023)Michael, Cucuringu, and Howison]{michael2023ofter}
Nikolas Michael, Mihai Cucuringu, and Sam Howison.
\newblock Ofter: An online pipeline for time series forecasting.
\newblock \emph{arXiv preprint arXiv:2304.03877}, 2023.

\bibitem[Zhang et~al.(2023{\natexlab{a}})Zhang, Wen, Wang, Chen, Sun, Zhang, Wang, Jin, and Tan]{zhang2023onenet}
YiFan Zhang, Qingsong Wen, Xue Wang, Weiqi Chen, Liang Sun, Zhang Zhang, Liang Wang, Rong Jin, and Tieniu Tan.
\newblock Onenet: Enhancing time series forecasting models under concept drift by online ensembling.
\newblock In \emph{Thirty-seventh Conference on Neural Information Processing Systems}, 2023{\natexlab{a}}.

\bibitem[Dooley et~al.(2023)Dooley, Khurana, Mohapatra, Naidu, and White]{dooley2023forecastpfn}
Samuel Dooley, Gurnoor~Singh Khurana, Chirag Mohapatra, Siddartha Naidu, and Colin White.
\newblock Forecastpfn: Synthetically-trained zero-shot forecasting.
\newblock \emph{arXiv preprint arXiv:2311.01933}, 2023.

\bibitem[Oreshkin et~al.(2021)Oreshkin, Carpov, Chapados, and Bengio]{oreshkin2021meta}
Boris~N Oreshkin, Dmitri Carpov, Nicolas Chapados, and Yoshua Bengio.
\newblock Meta-learning framework with applications to zero-shot time-series forecasting.
\newblock In \emph{Proceedings of the AAAI Conference on Artificial Intelligence}, volume~35, pages 9242--9250, 2021.

\bibitem[Zhou et~al.(2023)Zhou, Niu, Wang, Sun, and Jin]{zhou2023one}
Tian Zhou, Peisong Niu, Xue Wang, Liang Sun, and Rong Jin.
\newblock One fits all: Power general time series analysis by pretrained lm.
\newblock \emph{arXiv preprint arXiv:2302.11939}, 2023.

\bibitem[Radford et~al.()Radford, Wu, Child, Luan, Amodei, Sutskever, et~al.]{radford2019language}
Alec Radford, Jeffrey Wu, Rewon Child, David Luan, Dario Amodei, Ilya Sutskever, et~al.
\newblock Language models are unsupervised multitask learners.

\bibitem[Yeh et~al.(2023)Yeh, Dai, Chen, Zheng, Fan, Der, Lai, Zhuang, Wang, Wang, et~al.]{yeh2023toward}
Chin-Chia~Michael Yeh, Xin Dai, Huiyuan Chen, Yan Zheng, Yujie Fan, Audrey Der, Vivian Lai, Zhongfang Zhuang, Junpeng Wang, Liang Wang, et~al.
\newblock Toward a foundation model for time series data.
\newblock In \emph{Proceedings of the 32nd ACM International Conference on Information and Knowledge Management}, pages 4400--4404, 2023.

\bibitem[Jin et~al.(2023)Jin, Wang, Ma, Chu, Zhang, Shi, Chen, Liang, Li, Pan, et~al.]{jin2023time}
Ming Jin, Shiyu Wang, Lintao Ma, Zhixuan Chu, James~Y Zhang, Xiaoming Shi, Pin-Yu Chen, Yuxuan Liang, Yuan-Fang Li, Shirui Pan, et~al.
\newblock Time-llm: Time series forecasting by reprogramming large language models.
\newblock \emph{arXiv preprint arXiv:2310.01728}, 2023.

\bibitem[Garza and Mergenthaler-Canseco(2023)]{garza2023timegpt}
Azul Garza and Max Mergenthaler-Canseco.
\newblock Timegpt-1.
\newblock \emph{arXiv preprint arXiv:2310.03589}, 2023.

\bibitem[Box and Pierce(1970)]{box1970distribution}
George~EP Box and David~A Pierce.
\newblock Distribution of residual autocorrelations in autoregressive-integrated moving average time series models.
\newblock \emph{Journal of the American statistical Association}, 65\penalty0 (332):\penalty0 1509--1526, 1970.

\bibitem[Hochreiter and Schmidhuber(1997)]{hochreiter1997long}
Sepp Hochreiter and J{\"u}rgen Schmidhuber.
\newblock Long short-term memory.
\newblock \emph{Neural computation}, 9\penalty0 (8):\penalty0 1735--1780, 1997.

\bibitem[Chung et~al.(2014)Chung, Gulcehre, Cho, and Bengio]{chung2014empirical}
Junyoung Chung, Caglar Gulcehre, KyungHyun Cho, and Yoshua Bengio.
\newblock Empirical evaluation of gated recurrent neural networks on sequence modeling.
\newblock \emph{arXiv preprint arXiv:1412.3555}, 2014.

\bibitem[Bai et~al.(2018)Bai, Kolter, and Koltun]{bai2018empirical}
Shaojie Bai, J~Zico Kolter, and Vladlen Koltun.
\newblock An empirical evaluation of generic convolutional and recurrent networks for sequence modeling.
\newblock \emph{arXiv preprint arXiv:1803.01271}, 2018.

\bibitem[Wang et~al.(2022{\natexlab{a}})Wang, Peng, Huang, Wang, Chen, and Xiao]{wang2022micn}
Huiqiang Wang, Jian Peng, Feihu Huang, Jince Wang, Junhui Chen, and Yifei Xiao.
\newblock Micn: Multi-scale local and global context modeling for long-term series forecasting.
\newblock In \emph{The Eleventh International Conference on Learning Representations}, 2022{\natexlab{a}}.

\bibitem[Vaswani et~al.(2017)Vaswani, Shazeer, Parmar, Uszkoreit, Jones, Gomez, Kaiser, and Polosukhin]{vaswani2017attention}
Ashish Vaswani, Noam Shazeer, Niki Parmar, Jakob Uszkoreit, Llion Jones, Aidan~N Gomez, {\L}ukasz Kaiser, and Illia Polosukhin.
\newblock Attention is all you need.
\newblock \emph{Advances in neural information processing systems}, 30, 2017.

\bibitem[Nie et~al.(2023)Nie, H.~Nguyen, Sinthong, and Kalagnanam]{Yuqietal-2023-PatchTST}
Yuqi Nie, Nam H.~Nguyen, Phanwadee Sinthong, and Jayant Kalagnanam.
\newblock A time series is worth 64 words: Long-term forecasting with transformers.
\newblock In \emph{International Conference on Learning Representations}, 2023.

\bibitem[Liu et~al.(2022)Liu, Wu, Wang, and Long]{liu2022non}
Yong Liu, Haixu Wu, Jianmin Wang, and Mingsheng Long.
\newblock Non-stationary transformers: Exploring the stationarity in time series forecasting.
\newblock \emph{Advances in Neural Information Processing Systems}, 35:\penalty0 9881--9893, 2022.

\bibitem[Challu et~al.(2023)Challu, Olivares, Oreshkin, Ramirez, Canseco, and Dubrawski]{challu2023nhits}
Cristian Challu, Kin~G Olivares, Boris~N Oreshkin, Federico~Garza Ramirez, Max~Mergenthaler Canseco, and Artur Dubrawski.
\newblock Nhits: Neural hierarchical interpolation for time series forecasting.
\newblock In \emph{Proceedings of the AAAI Conference on Artificial Intelligence}, volume~37, pages 6989--6997, 2023.

\bibitem[Zeng et~al.(2023)Zeng, Chen, Zhang, and Xu]{zeng2023transformers}
Ailing Zeng, Muxi Chen, Lei Zhang, and Qiang Xu.
\newblock Are transformers effective for time series forecasting?
\newblock In \emph{Proceedings of the AAAI conference on artificial intelligence}, volume~37, pages 11121--11128, 2023.

\bibitem[Chen et~al.(2023)Chen, Li, Yoder, Arik, and Pfister]{chen2023tsmixer}
Si-An Chen, Chun-Liang Li, Nate Yoder, Sercan~O Arik, and Tomas Pfister.
\newblock Tsmixer: An all-mlp architecture for time series forecasting.
\newblock \emph{arXiv preprint arXiv:2303.06053}, 2023.

\bibitem[Li et~al.(2023)Li, Qi, Li, and Xu]{li2023revisiting}
Zhe Li, Shiyi Qi, Yiduo Li, and Zenglin Xu.
\newblock Revisiting long-term time series forecasting: An investigation on linear mapping.
\newblock \emph{arXiv preprint arXiv:2305.10721}, 2023.

\bibitem[Ma et~al.(2023)Ma, Liu, Zheng, Huang, Zhu, Yu, and Kwok]{ma2023survey}
Qianli Ma, Zhen Liu, Zhenjing Zheng, Ziyang Huang, Siying Zhu, Zhongzhong Yu, and James~T Kwok.
\newblock A survey on time-series pre-trained models.
\newblock \emph{arXiv preprint arXiv:2305.10716}, 2023.

\bibitem[Zhang et~al.(2023{\natexlab{b}})Zhang, Wen, Zhang, Cai, Jin, Liu, Zhang, Liang, Pang, Song, et~al.]{zhang2023self}
Kexin Zhang, Qingsong Wen, Chaoli Zhang, Rongyao Cai, Ming Jin, Yong Liu, James Zhang, Yuxuan Liang, Guansong Pang, Dongjin Song, et~al.
\newblock Self-supervised learning for time series analysis: Taxonomy, progress, and prospects.
\newblock \emph{arXiv preprint arXiv:2306.10125}, 2023{\natexlab{b}}.

\bibitem[Yue et~al.(2022)Yue, Wang, Duan, Yang, Huang, Tong, and Xu]{yue2022ts2vec}
Zhihan Yue, Yujing Wang, Juanyong Duan, Tianmeng Yang, Congrui Huang, Yunhai Tong, and Bixiong Xu.
\newblock Ts2vec: Towards universal representation of time series.
\newblock In \emph{Proceedings of the AAAI Conference on Artificial Intelligence}, volume~36, pages 8980--8987, 2022.

\bibitem[Wang et~al.(2022{\natexlab{b}})Wang, Xu, Zhang, Trajcevski, Zhong, and Zhou]{wang2022learning}
Zhiyuan Wang, Xovee Xu, Weifeng Zhang, Goce Trajcevski, Ting Zhong, and Fan Zhou.
\newblock Learning latent seasonal-trend representations for time series forecasting.
\newblock \emph{Advances in Neural Information Processing Systems}, 35:\penalty0 38775--38787, 2022{\natexlab{b}}.

\bibitem[Gruver et~al.(2023)Gruver, Finzi, Qiu, and Wilson]{gruver2023large}
Nate Gruver, Marc Finzi, Shikai Qiu, and Andrew~Gordon Wilson.
\newblock Large language models are zero-shot time series forecasters.
\newblock \emph{arXiv preprint arXiv:2310.07820}, 2023.

\bibitem[Rasul et~al.(2023)Rasul, Ashok, Williams, Khorasani, Adamopoulos, Bhagwatkar, Bilo{\v{s}}, Ghonia, Hassen, Schneider, et~al.]{rasul2023lag}
Kashif Rasul, Arjun Ashok, Andrew~Robert Williams, Arian Khorasani, George Adamopoulos, Rishika Bhagwatkar, Marin Bilo{\v{s}}, Hena Ghonia, Nadhir~Vincent Hassen, Anderson Schneider, et~al.
\newblock Lag-llama: Towards foundation models for time series forecasting.
\newblock \emph{arXiv preprint arXiv:2310.08278}, 2023.

\bibitem[Das et~al.(2023)Das, Kong, Sen, and Zhou]{das2023decoder}
Abhimanyu Das, Weihao Kong, Rajat Sen, and Yichen Zhou.
\newblock A decoder-only foundation model for time-series forecasting.
\newblock \emph{arXiv preprint arXiv:2310.10688}, 2023.

\bibitem[Yoon et~al.(2019)Yoon, Jarrett, and Van~der Schaar]{yoon2019time}
Jinsung Yoon, Daniel Jarrett, and Mihaela Van~der Schaar.
\newblock Time-series generative adversarial networks.
\newblock \emph{Advances in neural information processing systems}, 32, 2019.

\bibitem[Pei et~al.(2021)Pei, Ren, Yang, Liu, Qin, and Li]{pei2021towards}
Hengzhi Pei, Kan Ren, Yuqing Yang, Chang Liu, Tao Qin, and Dongsheng Li.
\newblock Towards generating real-world time series data.
\newblock In \emph{2021 IEEE International Conference on Data Mining (ICDM)}, pages 469--478. IEEE Computer Society, 2021.

\bibitem[M{\"u}ller et~al.(2021)M{\"u}ller, Hollmann, Arango, Grabocka, and Hutter]{muller2021transformers}
Samuel M{\"u}ller, Noah Hollmann, Sebastian~Pineda Arango, Josif Grabocka, and Frank Hutter.
\newblock Transformers can do bayesian inference.
\newblock In \emph{International Conference on Learning Representations}, 2021.

\bibitem[Hollmann et~al.(2022)Hollmann, M{\"u}ller, Eggensperger, and Hutter]{hollmann2022tabpfn}
Noah Hollmann, Samuel M{\"u}ller, Katharina Eggensperger, and Frank Hutter.
\newblock Tabpfn: A transformer that solves small tabular classification problems in a second.
\newblock In \emph{The Eleventh International Conference on Learning Representations}, 2022.

\bibitem[Devlin et~al.(2018)Devlin, Chang, Lee, and Toutanova]{devlin2018bert}
Jacob Devlin, Ming-Wei Chang, Kenton Lee, and Kristina Toutanova.
\newblock Bert: Pre-training of deep bidirectional transformers for language understanding.
\newblock \emph{arXiv preprint arXiv:1810.04805}, 2018.

\bibitem[Godahewa et~al.(2021)Godahewa, Webb, and Montero-Manso]{godahewa2021monash}
Bergmeir Godahewa, Hyndman Webb, and Montero-Manso.
\newblock Monash time series forecasting archive.
\newblock \emph{arXiv preprint}, 2021.

\bibitem[Wu et~al.(2022)Wu, Hu, Liu, Zhou, Wang, and Long]{wu2022timesnet}
Haixu Wu, Tengge Hu, Yong Liu, Hang Zhou, Jianmin Wang, and Mingsheng Long.
\newblock Timesnet: Temporal 2d-variation modeling for general time series analysis.
\newblock \emph{arXiv preprint arXiv:2210.02186}, 2022.

\bibitem[Dosovitskiy et~al.(2020)Dosovitskiy, Beyer, Kolesnikov, Weissenborn, Zhai, Unterthiner, Dehghani, Minderer, Heigold, Gelly, et~al.]{dosovitskiy2020image}
Alexey Dosovitskiy, Lucas Beyer, Alexander Kolesnikov, Dirk Weissenborn, Xiaohua Zhai, Thomas Unterthiner, Mostafa Dehghani, Matthias Minderer, Georg Heigold, Sylvain Gelly, et~al.
\newblock An image is worth 16x16 words: Transformers for image recognition at scale.
\newblock \emph{arXiv preprint arXiv:2010.11929}, 2020.

\bibitem[Liu et~al.(2023)Liu, Hu, Zhang, Wu, Wang, Ma, and Long]{liu2023itransformer}
Yong Liu, Tengge Hu, Haoran Zhang, Haixu Wu, Shiyu Wang, Lintao Ma, and Mingsheng Long.
\newblock itransformer: Inverted transformers are effective for time series forecasting.
\newblock \emph{arXiv preprint arXiv:2310.06625}, 2023.

\bibitem[Wu et~al.(2023)Wu, Hu, Liu, Zhou, Wang, and Long]{wu2023timesnet}
Haixu Wu, Tengge Hu, Yong Liu, Hang Zhou, Jianmin Wang, and Mingsheng Long.
\newblock Timesnet: Temporal 2d-variation modeling for general time series analysis.
\newblock In \emph{International Conference on Learning Representations}, 2023.

\end{thebibliography}

\newpage
\onecolumn
\appendix
\section{Synthetic prior details}
\label{app:generation_details}

\begin{enumerate}

    \item Fourier coefficients $c_i$ generation: we randomly select the number of coefficients $N_c$ to be sampled
    from 3 to 7, then $c_i \sim U[-2, 2], \forall i \in \{1, \ldots, N_c\}$
    \item Periodicity range: we use discrete uniform distribution $U[8, 199]$ to sample the cardinality of one period to be generated.

    \item Trend: we use linear, $\ln{(x)}$ with probability $~18\%$, $\ln{(1 + x)}$, exponential and quadratic trends equiprobably with probability $~9\%$ along with leaving pure seasonal data (without trend). For each trend type, the sharpness is a multiplicative coefficient in front of the trend is selected as follows: 
    \begin{itemize}
        \item linear: $U[-0.01, -0.0001] \cup U[0.0001, 0.01]$.
        \item $\ln(x)$: $U[-1, -0.01] \cup U[0.01, 1]$.
        \item $\ln(x + 1)$: $U[-1, -0.01] \cup U[0.01, 1]$.
        \item quadratic: $U[-0.001, -0.01] \cup U[0.001, 0.01]$.
        \item exponential: $U[-0.005, -0.0005] \cup U[0.0005, 0.005].$
    \end{itemize}
    \item To ensure the diversity of the historical data beginning, we randomly select the starting point from a discrete uniform distribution from 0 to the given periodicity range from step 2. The number of periods in the train is randomly sampled from the interval from 2 to 7.
    \item According to the previous step, we calculate the end point of the train part as the sum of the start position of the train, the number of periods in the train multiplied by the period length, and a random number sampled from a discrete uniform distribution from 0 to the period length.
    \item Having obtained the number of coefficients, the length of the period, and the number of periods, we generate pairs of coefficients from the interval from -2 to 2, apply the inverse Fourier transform to obtain a time representation from the frequency one, take the real part and duplicate its specified number of periods. Thus, we get a pure seasonality. Based on the previously selected trend, we generate it and add it to the net seasonality, leaving a 2\% probability for a net trend.
    \item We divide the series into train and target parts using separator positions mentioned in step 6 and apply min-max scaling using train part parameters. We also cut the values bigger than two or smaller than -1 to ensure scale consistency of data. To feed data to the transformer, we pad every batch by zeros to the same length and create masks.
\end{enumerate}

\section{Datasets information}
\label{sseq:real_data_details}
The proposed datasets can be described as follows (also see Table~\ref{table:real_data}):
\begin{itemize}
    \item The Weather dataset presents sensor data collected by the weather station located at Beutenberg, Germany. Daily seasonality, as well as characteristics such as heteroscedasticity and local trends, are observed in these series.
    \item The Traffic dataset captures road occupancy rates sourced from sensors in San Francisco. Both monthly and weekly seasonality are observed.
    \item Electricity shows daily electricity consumption patterns.
    \item The ILI (National Illness) dataset provides features related to the spread of influenza-like illness. The dataset displays annual seasonality with monthly periodicity. Additionally, two samples within the dataset exhibit trend and heteroscedasticity patterns.
    \item The ETT (Electricity Transformer Temperature) datasets are sourced from two distinct electric transformers. Daily and semi-daily seasonality, local trends, and regime transitions distinguish the ETTh dataset.
    \item The M4 dataset is recognized for its diverse collection of time series data, spanning various domains such as finance, climate, and healthcare.
    \item The M3 dataset covers economic indicators, demographic trends, and other significant domains.
\end{itemize}

\begin{table*}[!htbt]
\centering
\centering
\begin{tabular}{@{}lccc@{}}
\toprule
Dataset & Number of ts & Length of ts & Frequency \\ \midrule
Weather      & 21                    & 52969                 & 10 min    \\
Electricity  & 321                   & 26304                 & 1 hour    \\
Traffic      & 862                   & 17544                 & 1 hour    \\
ILI          & 7                     & 966                   & 7 days    \\
ETTh1, ETTh2 & 7                     & 17420                 & 1 hour    \\
ETTm1, ETTm2 & 7                     & 69680                 & 15 min    \\
M3 Yearly    & 645                   & 20-47                 & 1 year    \\
M3 Quarterly & 756                   & 24-72                 & 3 months  \\
M3 Monthly   & 1428                  & 66-144                & 1 month   \\
M3 Other     & 174                   & 71-104                & -         \\
M4 Yearly    & 23000                 & 19-841                & 1 year    \\
M4 Quarterly & 24000                 & 24-874                & 3 months  \\
M4 Monthly   & 48000                 & 60-2812               & 1 month   \\
M4 Weekly    & 359                   & 93-2610               & 7 days    \\
M4 Daily     & 4227                  & 107-9933              & 1 day     \\
M4 Hourly    & 414                   & 748-1008              & 1 hour    \\ \bottomrule
\end{tabular}%
\caption{Description of the main characteristics of time series datasets.}
\label{table:real_data}
\end{table*}

\section{Experimental Details}

\paragraph{Long-Term Multivariate forecasting.} The comparison of synthetic data-based approaches with supervised baselines for different horizons is presented in Table \ref{table:sup}. Notably, our approach achieves comparable performance to DLinear and GPT-2(0) for the ILI dataset characterized by short series, surpassing AutoFormer and FEDFormer.

\paragraph{Few-shot forecasting details.}
\label{app:few_shot_details}

\begin{figure}
    \begin{minipage}[b]{0.5\linewidth}
        \includegraphics[width=1\linewidth]{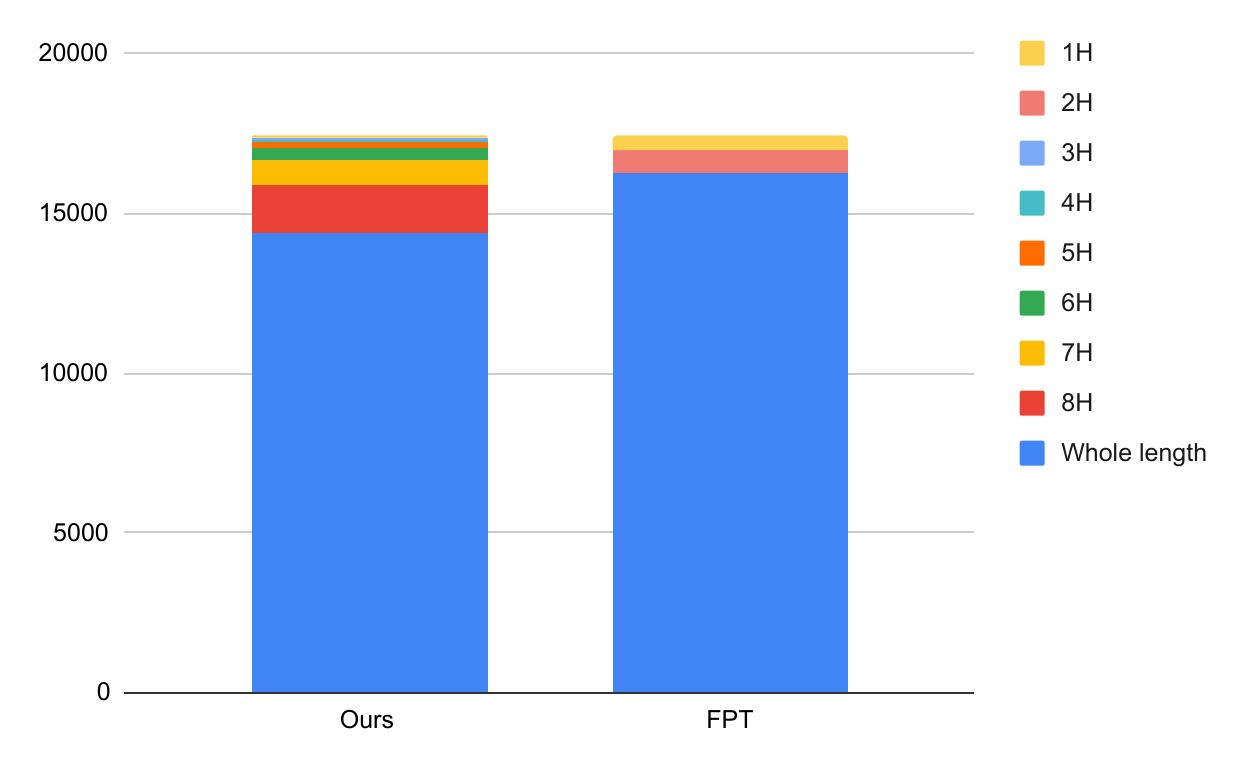}
        \caption{Few-shot samples for ETTh datasets.}
    \end{minipage}
    \begin{minipage}[b]{0.5\linewidth}
        \includegraphics[width=1\linewidth]{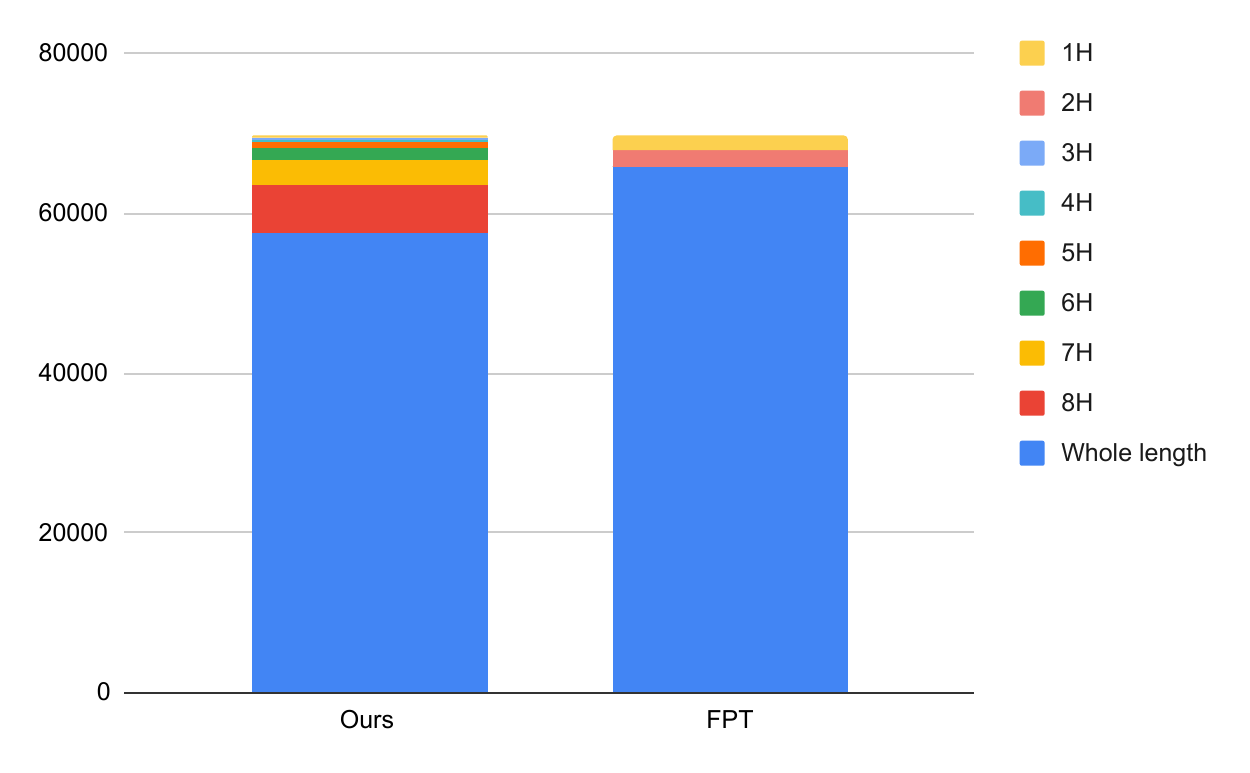}
        \caption{Few-shot samples for ETTm datasets.}
    \end{minipage}
    \begin{minipage}[b]{0.5\linewidth}
        \includegraphics[width=1\linewidth]{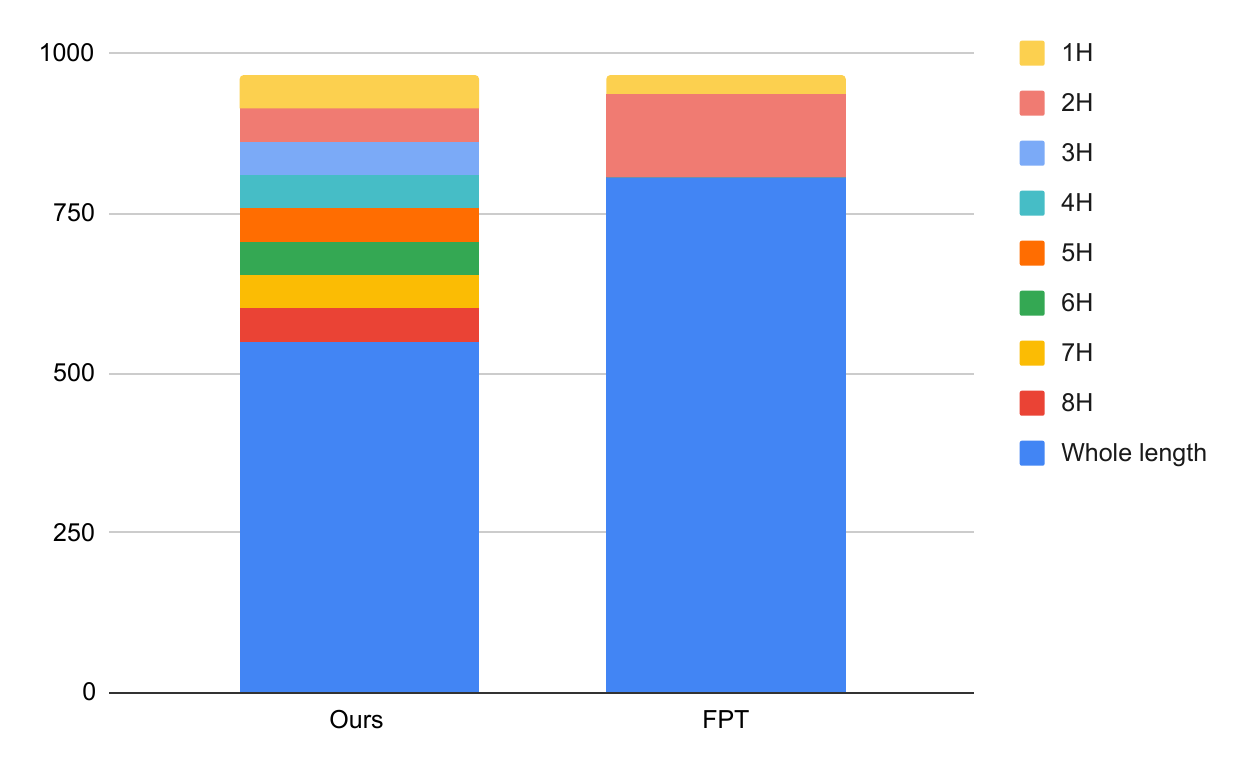}
        \caption{Few-shot samples for ILI datasets.}
    \end{minipage}
    \begin{minipage}[b]{0.5\linewidth}
        \includegraphics[width=1\linewidth]{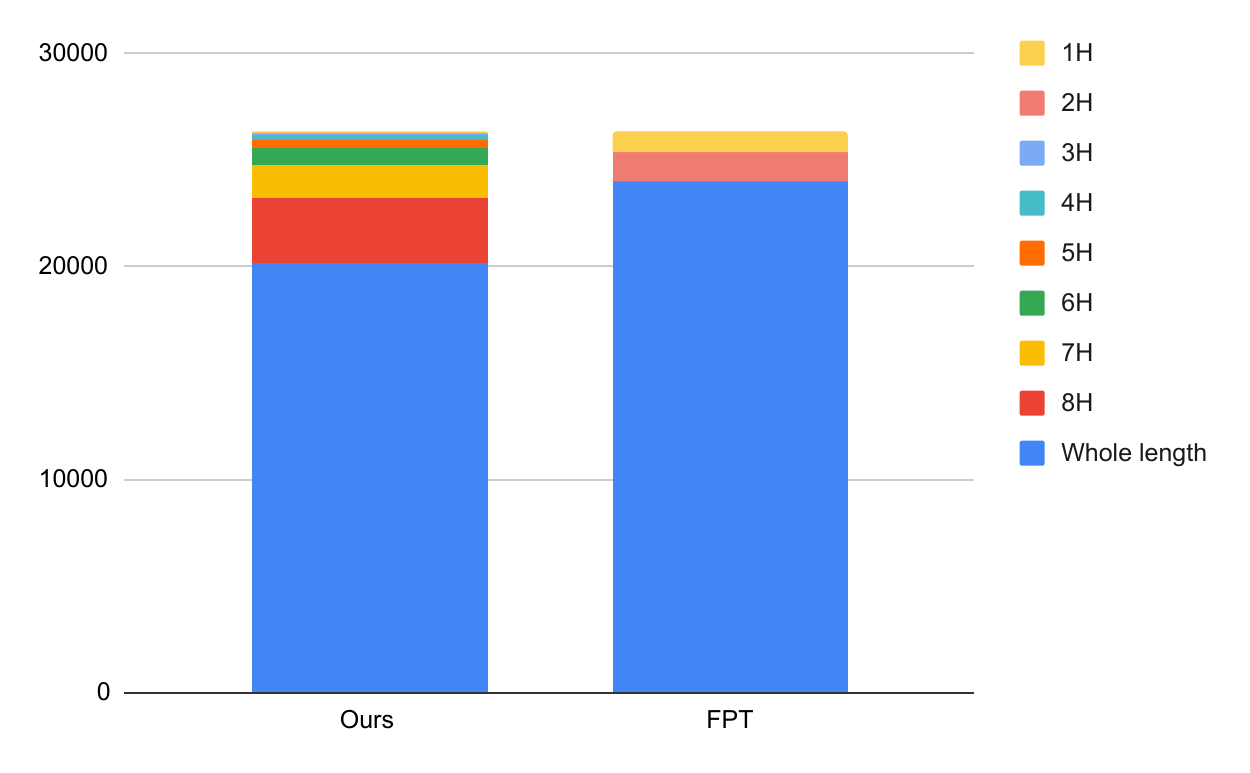}
        \caption{Few-shot samples for ECL datasets.}
    \end{minipage}
    \captionof{figure}{The comparison of training size relative to the entire series length is presented for both our few-shot experiments (\textbf{Ours}) and those conducted by \cite{zhou2023one} (\textbf{FPT}). Each divided bar segment corresponds to the step of the seasonality period, as detailed in Table~\ref{table:data_train_few_shot} and \cite{zhou2023one}.}
    \label{fig:ofa_ours_whole}
\end{figure}

Motivated by recognizing seasonality as a fundamental pattern within time series data, we deviated from the methodology employed in GPT2 for time series \cite{zhou2023one}. We incorporated the seasonality cycle as a pivotal element in sample formation. Consequently, our experimental grid for the few-shot setup encompasses varying numbers of data points corresponding to the range of seasonal periods, from 2 to 8. We include one period for the ILI dataset due to its abbreviated row length. Including 2 periods is based on the notion that this quantity is necessary for capturing the essence of seasonality. You can see details in \ref{table:data_train_few_shot}. Our approach enables the integration of a wider range of data points for few-shot experiments, including referenced samples from \cite{zhou2023one}. We present a comparative visualization of the few-shot sample ratio between our methodology and GPT2 for time series for reference in Figure~\ref{fig:ofa_ours_whole}.

\paragraph{Zero-shot forecasting details.}

\textit{ForecastPFN implementation details.} We faced several problems while using ForecastPFN \cite{dooley2023forecastpfn} source code.
\begin{enumerate}
    \item The inference implementation within the ForecastPFN source code exhibits significant computational inefficiency, rendering it suitable only for single-sample series inference. Furthermore, only the OT target\footnote{Oil Temperature for ETT and ECL datasets, Total number of patients for ILI, CO2 concentration in ppm (parts per million) for Weather and road occupancy rates for 862nd sensor for Traffic.} was considered in the initial paper experiments, diverging from standard time series benchmarking practices. This approach undermines the primary objective of zero-shot learning for time series, which aims to simultaneously enable rapid predictions for numerous short time series. To address this limitation, we undertook efforts to rewrite the inference code, optimizing it for full GPU utilization and implementing sample batching. These considerations are included within our codebase on GitHub.
    \item We encountered the issue of NaNs arising in predictions, a phenomenon absent when utilizing their benchmarking exclusively on the OT target. While we have not definitively determined the root cause of this behavior, we hypothesize that it may stem from local scale estimation. In this process, the calculation of local scale parameters: \texttt{history\_mean} (mean of the last points of history) and \texttt{history\_std} (standard deviation of the last points of history) involves considering the last 6 values from the history, which might be identical for non-OT series. This behaviour is observed for every dataset except for ILI, M4 Weekly and M4 Hourly. However, further investigation is warranted to elucidate this issue thoroughly. Notably, this issue is not anticipated to significantly impact predictions, given that the proportion of such values in predictions is consistently less than 1\% across all datasets.
\end{enumerate}

\begin{table}[ht!]
\centering
\begin{tabular}{ccccc}
\toprule
Dataset & Horizon & Period & Train's length & Train Series \\
\midrule
{\multirow{8}{*}{Weather}} & {\multirow{8}{*}{96}} & {\multirow{8}{*}{24}} & H * 2\textasciicircum1 & 48 \\
\multirow{8}{*} & \multirow{8}{*} & \multirow{8}{*} & H * 2\textasciicircum2 & 96 \\
\multirow{8}{*} & \multirow{8}{*} & \multirow{8}{*} & H * 2\textasciicircum3 & 192 \\
\multirow{8}{*} & \multirow{8}{*} & \multirow{8}{*} & H * 2\textasciicircum4 & 384 \\
\multirow{8}{*} & \multirow{8}{*} & \multirow{8}{*} & H * 2\textasciicircum5 & 768 \\
\multirow{8}{*} & \multirow{8}{*} & \multirow{8}{*} & H * 2\textasciicircum6 & 1536 \\
\multirow{8}{*} & \multirow{8}{*} & \multirow{8}{*} & H * 2\textasciicircum7 & 3072 \\
\multirow{8}{*} & \multirow{8}{*} & \multirow{8}{*} & H * 2\textasciicircum8 & 6144 \\
\midrule
{\multirow{8}{*}{ETTh1, ETTh2}} & {\multirow{8}{*}{96}} & {\multirow{8}{*}{12}} & H * 2\textasciicircum1 & 24 \\
\multirow{8}{*} & \multirow{8}{*} & \multirow{8}{*} & H * 2\textasciicircum2 & 48 \\
\multirow{8}{*} & \multirow{8}{*} & \multirow{8}{*} & H * 2\textasciicircum3 & 96 \\
\multirow{8}{*} & \multirow{8}{*} & \multirow{8}{*} & H * 2\textasciicircum4 & 192 \\
\multirow{8}{*} & \multirow{8}{*} & \multirow{8}{*} & H * 2\textasciicircum5 & 384 \\
\multirow{8}{*} & \multirow{8}{*} & \multirow{8}{*} & H * 2\textasciicircum6 & 768 \\
\multirow{8}{*} & \multirow{8}{*} & \multirow{8}{*} & H * 2\textasciicircum7 & 1536 \\
\multirow{8}{*} & \multirow{8}{*} & \multirow{8}{*} & H * 2\textasciicircum8 & 3072 \\
\midrule
{\multirow{8}{*}{ETTm1, ETTm2}} & {\multirow{8}{*}{96}} & {\multirow{8}{*}{48}} & H * 2\textasciicircum1 & 96 \\
\multirow{8}{*} & \multirow{8}{*} & \multirow{8}{*} & H * 2\textasciicircum2 & 192 \\
\multirow{8}{*} & \multirow{8}{*} & \multirow{8}{*} & H * 2\textasciicircum3 & 384 \\
\multirow{8}{*} & \multirow{8}{*} & \multirow{8}{*} & H * 2\textasciicircum4 & 768 \\
\multirow{8}{*} & \multirow{8}{*} & \multirow{8}{*} & H * 2\textasciicircum5 & 1536 \\
\multirow{8}{*} & \multirow{8}{*} & \multirow{8}{*} & H * 2\textasciicircum6 & 3072 \\
\multirow{8}{*} & \multirow{8}{*} & \multirow{8}{*} & H * 2\textasciicircum7 & 6144 \\
\multirow{8}{*} & \multirow{8}{*} & \multirow{8}{*} & H * 2\textasciicircum8 & 12288 \\
\midrule
{\multirow{8}{*}{ECL}} & {\multirow{8}{*}{96}} & {\multirow{8}{*}{24}} & H * 2\textasciicircum1 & 48 \\
\multirow{8}{*} & \multirow{8}{*} & \multirow{8}{*} & H * 2\textasciicircum2 & 96 \\
\multirow{8}{*} & \multirow{8}{*} & \multirow{8}{*} & H * 2\textasciicircum3 & 192 \\
\multirow{8}{*} & \multirow{8}{*} & \multirow{8}{*} & H * 2\textasciicircum4 & 384 \\
\multirow{8}{*} & \multirow{8}{*} & \multirow{8}{*} & H * 2\textasciicircum5 & 768 \\
\multirow{8}{*} & \multirow{8}{*} & \multirow{8}{*} & H * 2\textasciicircum6 & 1536 \\
\multirow{8}{*} & \multirow{8}{*} & \multirow{8}{*} & H * 2\textasciicircum7 & 3072 \\
\multirow{8}{*} & \multirow{8}{*} & \multirow{8}{*} & H * 2\textasciicircum8 & 6144 \\
\midrule
{\multirow{8}{*}{ILI}} & {\multirow{8}{*}{24}} & {\multirow{8}{*}{52}} & H * 1 & 52 \\
\multirow{8}{*} & \multirow{8}{*} & \multirow{8}{*} & H * 2 & 104 \\
\multirow{8}{*} & \multirow{8}{*} & \multirow{8}{*} & H * 3 & 156 \\
\multirow{8}{*} & \multirow{8}{*} & \multirow{8}{*} & H * 4 & 208 \\
\multirow{8}{*} & \multirow{8}{*} & \multirow{8}{*} & H * 5 & 260 \\
\multirow{8}{*} & \multirow{8}{*} & \multirow{8}{*} & H * 6 & 312 \\
\multirow{8}{*} & \multirow{8}{*} & \multirow{8}{*} & H * 7 & 364 \\
\multirow{8}{*} & \multirow{8}{*} & \multirow{8}{*} & H * 8 & 416 \\
\bottomrule
\end{tabular}
\caption{Train budgets of few-shot forecasting in our experiments.}
\label{table:data_train_few_shot}
\end{table}

\paragraph{Zero-shot forecasting with different scalers.}
\label{sseq:zero_shot_scaling}
The empirical evidence suggests that scaling a target dataset to align with a source dataset can enhance model predictions. Subsequently, we conducted a concise comparative study, evaluating several commonly employed scalers: StandardScaler, MinMaxScaler, and QuantileScaler. StandardScaler transforms data by centering it around zero and scaling it to have a standard deviation of one. MinMaxScaler transforms features to a predefined range, typically between 0 and 1. Finally, QuantileScale scales data using the interquartile range.

The steps entailed in this investigation were the following: Initially, we standardized the target dataset to conform to a ‘default’ distribution by deriving scaler parameters and applying them to this target dataset. Subsequently, utilizing the resultant distribution, we scaled the data to align with the distribution of the source dataset. At the post-inference stage, we applied inverse transformations to the obtained predictions. The outcomes of this process are detailed in Table \ref{table:zero_shot_scalers}, presenting results obtained from the validation subset of the series. The StandardScaler was universally applied to all models and source datasets in the ultimate iteration, except DLinear, trained on M3.

\begin{figure}
    \begin{minipage}[b]{.46\linewidth}
        \centering
        \resizebox{1\textwidth}{!}{
        \begin{tabular}{l|rrrrr}
        \toprule
        \textbf{Source} & \multicolumn{1}{l}{\textbf{PatchTST}} & \multicolumn{1}{l}{\textbf{GPT2(6)}} & \multicolumn{1}{l}{\textbf{DLinear}} & \multicolumn{1}{l}{\textbf{ForecastPFN}} & \multicolumn{1}{l}{\textbf{OURS}} \\
        \midrule
        M3 Monthly & 0 & \textbf{0.5} & 0.14 & 0 & 0.36 \\
        M3 Other & 0 & 0 & 0 & 0.29 & \textbf{0.71} \\
        M3 Quarterly & 0 & 0.07 & 0 & 0.29 & \textbf{0.64} \\
        M3 Yearly & 0 & 0 & 0 & 0.29 & \textbf{0.71} \\
        M4 Daily & 0 & 0 & \textbf{0.62} & 0 & 0.38 \\
        M4 Hourly & 0 & 0.12 & \textbf{0.75} & 0 & 0.12 \\
        M4 Monthly & 0 & \textbf{0.38} & 0.25 & 0 & \textbf{0.38} \\
        M4 Quarterly & 0 & 0 & \textbf{0.62} & 0 & 0.38 \\
        M4 Weekly & 0 & 0.25 & \textbf{0.38} & 0 & \textbf{0.38} \\
        M4 Yearly & 0 & \textbf{0.38} & 0.25 & 0 & \textbf{0.38} \\ 
        \bottomrule
        \end{tabular}}
  \end{minipage}\hfill
  \begin{minipage}[b]{.45\linewidth}
    \centering
    \resizebox{1\textwidth}{!}{
    \begin{tabular}{l|rrrrr}
        \toprule
        \textbf{Source} & \multicolumn{1}{l}{\textbf{PatchTST}} & \multicolumn{1}{l}{\textbf{GPT2(6)}} & \multicolumn{1}{l}{\textbf{DLinear}} & \multicolumn{1}{l}{\textbf{ForecastPFN}} & \multicolumn{1}{l}{\textbf{Ours}} \\
        \midrule
        M3 Monthly & 0.07 & \textbf{0.57} & 0.07 & 0 & 0.29 \\
        M3 Other & 0 & 0 & 0.07 & 0.14 & \textbf{0.79} \\
        M3 Quarterly & 0 & 0 & 0 & 0.14 & \textbf{0.86} \\
        M3 Yearly & 0 & 0 & 0 & 0.14 & \textbf{0.86} \\
        M4 Daily & 0 & 0 & \textbf{0.75} & 0 & 0.25 \\
        M4 Hourly & 0 & 0.12 & \textbf{0.75} & 0 & 0.12 \\
        M4 Monthly & 0 & \textbf{0.75} & 0 & 0 & 0.25 \\
        M4 Quarterly & 0 & 0 & \textbf{0.75} & 0 & 0.25 \\
        M4 Weekly & 0 & \textbf{0.38} & \textbf{0.38} & 0 & 0.25 \\
        M4 Yearly & 0 & 0.25 & \textbf{0.5} & 0 & 0.25 \\
        \bottomrule
    \end{tabular}}
  \end{minipage}
    \begin{minipage}[b]{.5\linewidth}
        \centering
        \includegraphics[width=1\linewidth]{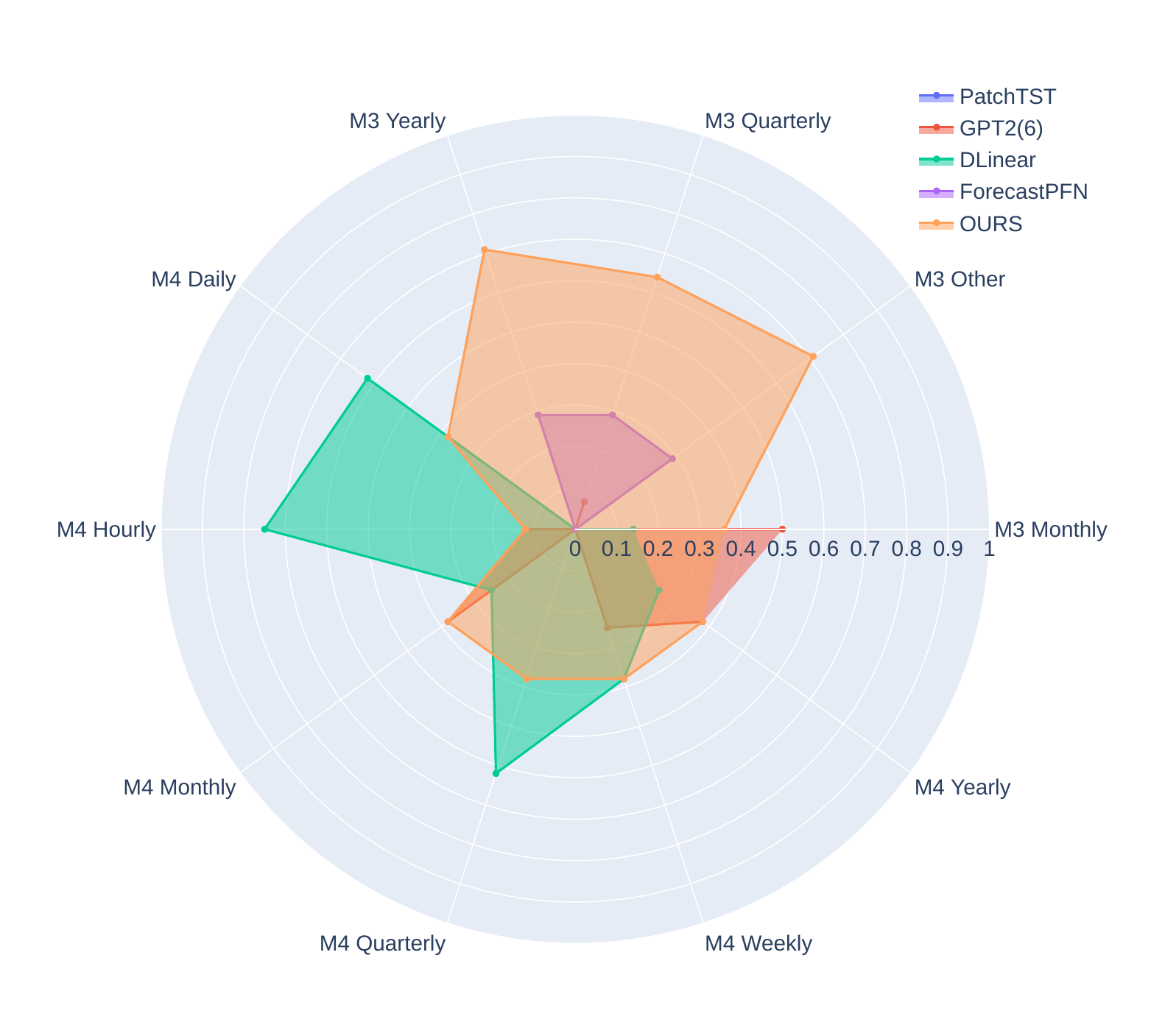}
        \captionof{figure}{Share of wins for MSE.}
  \end{minipage}\hfill
  \begin{minipage}[b]{.5\linewidth}
        \centering
        \includegraphics[width=1\linewidth]{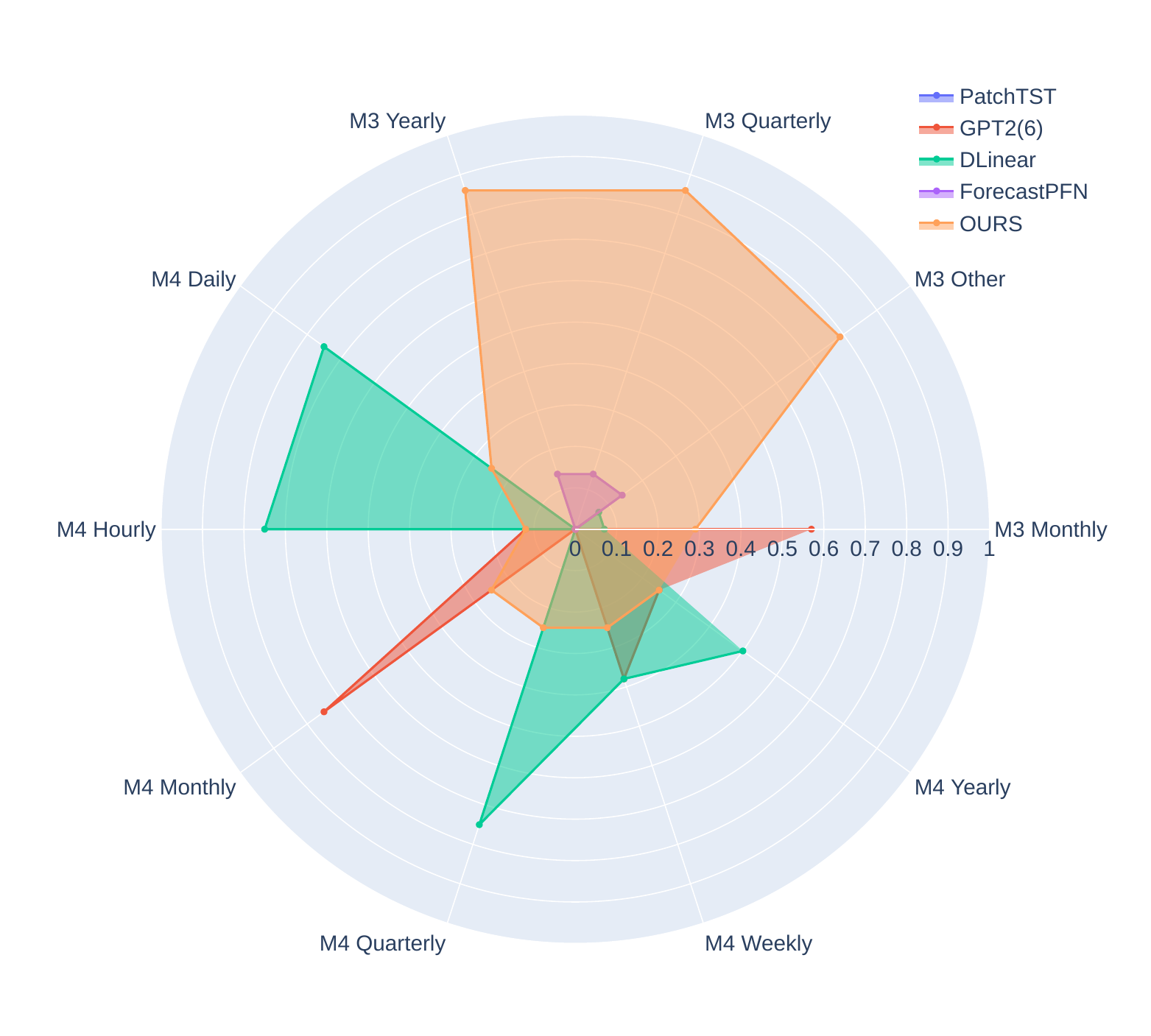}
        \captionof{figure}{Share of wins for SMAPE.}
  \end{minipage}
  \captionof{figure}{The model winning rate -- the ratio of superiority cases over other models -- inside the source dataset according to the MSE and SMAPE. Our approach outperforms ForecastPFN (FPFN) for all granularities of M3 and M4 datasets. In the tables, the best model for a given source dataset is highlighted in \textbf{bold}.}
  \label{app:win_rate}
\end{figure}

The comparison of synthetic data-based approaches with supervised baselines transferability between datasets is presented in Figure~\ref{app:win_rate} and Tables~\ref{table:zeroshot_m3_bench}, \ref{table:zeroshot_m4_bench}, \ref{table:zeroshot_m3_m4}, \ref{table:zero_few_sup}.

\begin{table*}[!htbt]
\begingroup
\setlength{\tabcolsep}{3pt} 
\resizebox{\textwidth}{!}{%

\begin{tabular}{@{}cc|c|cccc|cccc|cccc@{}}
\toprule
\multicolumn{1}{c}{\textbf{Source}} & \multicolumn{1}{|c|}{\textbf{Models}} & \multicolumn{1}{c|}{\multirow{1}{*}{\textbf{Target}}} & \footnotesize \textbf{DLinear SS} & \footnotesize \textbf{DLinear MMS} & \footnotesize \textbf{DLinear QT} & \footnotesize \textbf{DLinear} & \footnotesize \textbf{GPT4TS SS} & \footnotesize \textbf{GPT4TS MMS} & \footnotesize \textbf{GPT4TS QT} & \footnotesize \textbf{GPT4TS} & \footnotesize \textbf{PatchTST SS} & \footnotesize \textbf{PatchTST MMS} & \footnotesize \textbf{PatchTST QT} & \footnotesize \textbf{PatchTST} \\
\textbf{Dataset} & \multicolumn{1}{|c|}{\textbf{Horizon}} & \multicolumn{1}{c|}{\textbf{Dataset}} & \textbf{MAE} & \textbf{MAE} & \textbf{MAE} & \textbf{MAE} & \textbf{MAE} & \textbf{MAE} & \textbf{MAE} & \textbf{MAE} & \textbf{MAE} & \textbf{MAE} & \textbf{MAE} & \textbf{MAE} \\ \midrule
\multicolumn{1}{c|}{\multirow{2}{*}{\textbf{M3 Yearly}}} & \multirow{2}{*}{6} & Weather & 0.821 & 78.261 & 3.126 & \underline{\textbf{0.361}} & \textbf{1.019} & 1.019 & 1.067 & 1.020 & \textbf{1.445} & \textbf{1.445} & 1.522 & 1.446 \\
\multicolumn{1}{c|}{} &  & ILI & 1.382 & 1.338 & 1.186 & \underline{\textbf{0.925}} & \textbf{1.964} & 1.964 & 2.679 & 1.964 & \textbf{2.550} & \textbf{2.550} & 3.801 & 2.550 \\
\multicolumn{1}{c|}{\multirow{2}{*}{\textbf{M3 Quarterly}}} & \multirow{2}{*}{8} & Weather & 1.226 & 82.791 & 1.273 & \underline{\textbf{0.379}} & 0.568 & 0.568 & \textbf{0.557} & 0.568 & \textbf{0.907} & 0.907 & 0.948 & 0.908 \\
\multicolumn{1}{c|}{} &  & ILI & 1.941 & 1.455 & 1.767 & \underline{\textbf{0.897}} & 0.993 & 0.993 & \textbf{0.949} & 0.993 & \textbf{1.612} & \textbf{1.612} & 1.816 & 1.612 \\
\multicolumn{1}{c|}{\multirow{2}{*}{\textbf{M3 Monthly}}} & \multirow{2}{*}{18} & Weather & \underline{\textbf{0.270}} & 4.315 & 0.296 & 0.862 & \textbf{0.271} & 0.271 & 0.289 & 0.271 & 0.500 & 0.500 & \textbf{0.497} & 0.500 \\
\multicolumn{1}{c|}{} &  & ILI & 0.749 & 0.749 & \underline{\textbf{0.723}} & 0.945 & 0.736 & 0.736 & \textbf{0.735} & 0.736 & 0.773 & 0.773 & \textbf{0.760} & 0.773 \\
\multicolumn{1}{c|}{\multirow{2}{*}{\textbf{M3 Other}}} & \multirow{2}{*}{8} & Weather & 0.428 & 19.880 & 0.524 & \underline{\textbf{0.400}} & 0.670 & \textbf{0.670} & 1.101 & 0.670 & \textbf{0.473} & \textbf{0.473} & 1.097 & 0.473 \\
\multicolumn{1}{c|}{} &  & ILI & 0.751 & 0.774 & 0.789 & \underline{\textbf{0.737}} & 0.953 & \textbf{0.953} & 0.953 & 0.953 & \textbf{1.381} & \textbf{1.381} & 1.414 & \textbf{1.381} \\
\multicolumn{1}{c|}{\multirow{2}{*}{\textbf{M4 Yearly}}} & \multirow{2}{*}{6} & Weather & \underline{\textbf{0.144}} & 1.751 & 0.201 & 3.346 & \textbf{0.167} & 0.167 & 0.180 & 0.167 & 1.211 & 1.211 & \textbf{1.091} & 1.212 \\
\multicolumn{1}{c|}{} &  & ILI & \underline{\textbf{0.438}} & 0.438 & 0.477 & 3.373 & \textbf{0.581} & 0.581 & 0.584 & 0.581 & \textbf{2.638} & \textbf{2.638} & 3.427 & 2.638 \\
\multicolumn{1}{c|}{\multirow{2}{*}{\textbf{M4 Quarterly}}} & \multirow{2}{*}{8} & Weather & \underline{\textbf{0.173}} & 1.843 & 0.199 & 22.224 & \textbf{0.181} & 0.181 & 0.196 & 0.181 & 0.529 & 0.529 & \textbf{0.517} & 0.529 \\
\multicolumn{1}{c|}{} &  & ILI & 0.552 & 0.552 & \underline{\textbf{0.544}} & 22.206 & 0.607 & 0.607 & 0.631 & \textbf{0.607} & 0.923 & 0.923 & \textbf{0.869} & 0.923 \\
\multicolumn{1}{c|}{\multirow{2}{*}{\textbf{M4 Monthly}}} & \multirow{2}{*}{18} & Weather & \textbf{0.220} & 1.754 & 0.224 & 87.047 & 0.191 & \underline{\textbf{0.191}} & 0.236 & 0.191 & 0.428 & 0.428 & \textbf{0.424} & 0.428 \\
\multicolumn{1}{c|}{} &  & ILI & 0.803 & 0.806 & \textbf{0.788} & 86.951 & 0.752 & \textbf{0.752} & 0.812 & 0.752 & 0.688 & 0.688 & \underline{\textbf{0.676}} & 0.688 \\
\multicolumn{1}{c|}{\multirow{2}{*}{\textbf{M4 Weekly}}} & \multirow{2}{*}{13} & Weather & \underline{\textbf{0.215}} & 2.372 & 0.613 & 9.439 & \textbf{0.216} & 0.216 & 0.572 & 0.216 & \textbf{0.369} & 0.369 & 0.381 & 0.369 \\
\multicolumn{1}{c|}{} &  & ILI & \textbf{0.485} & 0.486 & 0.508 & 9.420 & 0.440 & \underline{\textbf{0.440}} & 0.454 & 0.440 & 0.631 & 0.631 & \textbf{0.630} & 0.631 \\
\multicolumn{1}{c|}{\multirow{2}{*}{\textbf{M4 Daily}}} & \multirow{2}{*}{14} & Weather & \underline{\textbf{0.172}} & 0.791 & 0.190 & 20.387 & 0.198 & \textbf{0.198} & 0.784 & 0.198 & \textbf{0.386} & \textbf{0.386} & 0.731 & 0.386 \\
\multicolumn{1}{c|}{} &  & ILI & \underline{\textbf{0.651}} & 0.651 & 0.667 & 20.397 & \textbf{0.705} & 0.705 & 0.766 & 0.705 & 0.905 & 0.905 & \textbf{0.867} & 0.905 \\
\multicolumn{1}{c|}{\multirow{2}{*}{\textbf{M4 Hourly}}} & \multirow{2}{*}{48} & Weather & \underline{\textbf{0.330}} & 2.079 & 0.820 & 0.554 & 0.399 & \textbf{0.399} & 0.754 & 0.400 & \textbf{0.427} & \textbf{0.427} & 0.593 & 0.427 \\
\multicolumn{1}{c|}{} &  & ILI & \textbf{1.222} & 1.223 & 1.756 & 1.278 & 1.293 & 1.293 & 1.737 & \textbf{1.293} & \underline{\textbf{0.666}} & \underline{\textbf{0.666}} & 1.107 & \underline{\textbf{0.666}} \\ \midrule
\multicolumn{3}{l|}{\textbf{Win Share}} & \textbf{0.55} & 0 & 0.15 & 0.3 & \textbf{0.4} & 0.35 & 0.15 & 0.1 & \textbf{0.55} & 0.45 & 0.45 & 0.1 \\ \bottomrule
\end{tabular}%
}
\caption{Zero-shot forecasting with the application of different scalers on the validation split of the target dataset. SS means using StandardScaler, MMS - MinMaxScaler, QT - QuantileTransformer, raw model's name - absence of any scaler. \textbf{Bold} indicates the best result within the model and source and target datasets for each scaler. Win Share indicates the fraction of the best results within the model (i.e., the fraction of the best results for the model in the column). \underline{Underlined} values indicate the best results for each source and target. Notably, DLinear consistently exhibits superior performance across various experiments, with minimal dependency on scaling factors. However, for specific datasets sourced from M4, PatchTST demonstrates superior performance. This phenomenon occurs particularly when performance across different scalers remains relatively consistent.}
\label{table:zero_shot_scalers}
\endgroup
\end{table*}

\begin{table*}[!htbt]
\resizebox{\textwidth}{!}{%
\begin{tabular}{cc|rr|ll|rr|rr|rr|rr|rr|rr|rr|rr|rr}
\toprule
\multicolumn{2}{c|}{\textbf{Models}} & \multicolumn{2}{c|}{\textbf{Ours}} & \multicolumn{2}{c|}{\textbf{ForecastPFN}} & \multicolumn{2}{c|}{\textbf{GPT2(6)-A}} & \multicolumn{2}{c|}{\textbf{GPT2(6)-F}} & \multicolumn{2}{c|}{\textbf{GPT2(0)}} & \multicolumn{2}{c|}{\textbf{DLinear}} & \multicolumn{2}{c|}{\textbf{PatchTST}} & \multicolumn{2}{c|}{\textbf{TimesNet}} & \multicolumn{2}{c|}{\textbf{FEDformer}} & \multicolumn{2}{c|}{\textbf{Autoformer}} & \multicolumn{2}{c}{\textbf{ITransformer}} \\ 
\multicolumn{1}{c}{\textbf{Dataset}} & \multicolumn{1}{c|}{\textbf{Horizon}} & \multicolumn{1}{c}{\textbf{MSE}} & \multicolumn{1}{c|}{\textbf{MAE}} & \multicolumn{1}{c}{\textbf{MSE}} & \multicolumn{1}{c|}{\textbf{MAE}} & \multicolumn{1}{c}{\textbf{MSE}} & \multicolumn{1}{r|}{\textbf{MAE}} & \multicolumn{1}{c}{\textbf{MSE}} & \multicolumn{1}{c|}{\textbf{MAE}} & \multicolumn{1}{c}{\textbf{MSE}} & \multicolumn{1}{c|}{\textbf{MAE}} & \multicolumn{1}{c}{\textbf{MSE}} & \multicolumn{1}{c|}{\textbf{MAE}} & \multicolumn{1}{c}{\textbf{MSE}} & \multicolumn{1}{c|}{\textbf{MAE}} & \multicolumn{1}{c}{\textbf{MSE}} & \multicolumn{1}{c|}{\textbf{MAE}} & \multicolumn{1}{c}{\textbf{MSE}} & \multicolumn{1}{c|}{\textbf{MAE}} & \multicolumn{1}{c}{\textbf{MSE}} & \multicolumn{1}{c|}{\textbf{MAE}} & \multicolumn{1}{c}{\textbf{MSE}} & \multicolumn{1}{c}{\textbf{MAE}} \\ \midrule

\multicolumn{1}{c}{\multirow{5}{*}{\textbf{Weather}}} &  \textbf{96} & 0.379 & 0.353 & 0.251 & 0.281 & 0.144 & 0.183 & 0.162 & 0.212 & 0.181 & 0.232 & 0.176 & 0.237 & 0.149 & 0.198 & 0.172 & 0.22 & 0.217 & 0.296 & 0.266 & 0.336 & 0.174 & 0.214 \\ 
\multicolumn{1}{c}{}  & \textbf{192} & 0.772 & 0.491 & 0.291 & 0.308 & 0.188 & 0.228 & 0.204 & 0.248 & 0.222 & 0.266 & 0.22 & 0.282 & 0.194 & 0.241 & 0.219 & 0.261 & 0.276 & 0.336 & 0.307 & 0.367 & 0.221 & 0.254 \\ 
\multicolumn{1}{c}{}  & \textbf{336} & 1.643 & 0.688 & 0.347 & 0.343 & 0.239 & 0.268 & 0.254 & 0.286 & 0.27 & 0.299 & 0.265 & 0.319 & 0.245 & 0.282 & 0.28 & 0.306 & 0.339 & 0.38 & 0.359 & 0.395 & 0.278 & 0.296 \\ 
\multicolumn{1}{c}{}  & \textbf{720} & 6.451 & 1.131 & 0.422 & 0.390 & 0.308 & 0.321 & 0.326 & 0.337 & 0.338 & 0.345 & 0.333 & 0.362 & 0.314 & 0.334 & 0.365 & 0.359 & 0.403 & 0.428 & 0.419 & 0.428 & 0.358 & 0.349 \\ 

\multicolumn{1}{c}{}  & \textbf{Avg.} & 2.311	& 0.666	& 0.327	& 0.331	& 0.220	& 0.250	& 0.237	& 0.271	& 0.253	& 0.286	& 0.249	& 0.300	& 0.226	& 0.264	& 0.259	& 0.287	& 0.309	& 0.360	& 0.338	& 0.382	& 0.258	& 0.278 \\ 

\midrule
\multicolumn{1}{c}{\multirow{5}{*}{\textbf{ETTh1}}} &  \textbf{96} & 0.828 & 0.639 & \multicolumn{1}{c}{1.856} & \multicolumn{1}{c|}{0.915} & 0.366 & 0.394 & 0.376 & 0.397 & 0.422 & 0.428 & 0.375 & 0.399 & 0.37 & 0.399 & 0.384 & 0.402 & 0.376 & 0.419 & 0.449 & 0.459 & 0.386 & 0.405 \\ 
\multicolumn{1}{c}{}  & \textbf{192} & 1.280 & 0.794 & \multicolumn{1}{c}{1.847} & \multicolumn{1}{c|}{0.926} & 0.407 & 0.42 & 0.416 & 0.418 & 0.466 & 0.45 & 0.405 & 0.416 & 0.413 & 0.421 & 0.436 & 0.429 & 0.42 & 0.448 & 0.5 & 0.482 & 0.441 & 0.436 \\ 
\multicolumn{1}{c}{} & \textbf{336} & 2.299 & 1.036 & \multicolumn{1}{c}{1.877} & \multicolumn{1}{c|}{0.941} & 0.42 & 0.439 & 0.442 & 0.433 & 0.488 & 0.464 & 0.439 & 0.443 & 0.422 & 0.436 & 0.491 & 0.469 & 0.459 & 0.465 & 0.521 & 0.496 & 0.487 & 0.458 \\ 
\multicolumn{1}{c}{}  & \textbf{720} & 4.999 & 1.471 & \multicolumn{1}{c}{2.025} & \multicolumn{1}{c|}{0.993} & 0.432 & 0.455 & 0.477 & 0.456 & 0.485 & 0.478 & 0.472 & 0.49 & 0.447 & 0.466 & 0.521 & 0.5 & 0.506 & 0.507 & 0.514 & 0.512 & 0.503 & 0.491 \\

\multicolumn{1}{c}{}  & \textbf{Avg.} & 2.351	& 0.985	& 1.901	& 0.944	& 0.406	& 0.427	& 0.428	& 0.426	& 0.465	& 0.455	& 0.423	& 0.437	& 0.413	& 0.431	& 0.458	& 0.450	& 0.440	& 0.460	& 0.496	& 0.487	& 0.454	& 0.448 \\ 
\midrule
\multicolumn{1}{c}{\multirow{5}{*}{\textbf{ETTh2}}} &  \textbf{96} & 0.827 & 0.583 & \multicolumn{1}{c}{0.361} & \multicolumn{1}{c|}{0.410} & 0.269 & 0.331 & 0.285 & 0.342 & 0.318 & 0.368 & 0.289 & 0.353 & 0.274 & 0.336 & 0.34 & 0.374 & 0.358 & 0.397 & 0.346 & 0.388 & 0.297 & 0.349 \\ 
\multicolumn{1}{c}{}  & \textbf{192} & 1.828 & 0.816 & \multicolumn{1}{r}{0.402} & \multicolumn{1}{r|}{0.438} & 0.334 & 0.379 & 0.354 & 0.389 & 0.383 & 0.407 & 0.383 & 0.418 & 0.339 & 0.379 & 0.402 & 0.414 & 0.429 & 0.439 & 0.456 & 0.452 & 0.38 & 0.4 \\ 
\multicolumn{1}{c}{}  & \textbf{336} & 3.453 & 1.078 & \multicolumn{1}{r}{0.410} & \multicolumn{1}{r|}{0.444} & 0.359 & 0.398 & 0.373 & 0.407 & 0.406 & 0.427 & 0.448 & 0.465 & 0.329 & 0.38 & 0.452 & 0.452 & 0.496 & 0.487 & 0.482 & 0.486 & 0.428 & 0.432 \\ 
\multicolumn{1}{c}{}  & \textbf{720} & 11.443 & 1.575 & \multicolumn{1}{r}{0.451} & \multicolumn{1}{r|}{0.471} & 0.392 & 0.433 & 0.406 & 0.441 & 0.42 & 0.446 & 0.605 & 0.551 & 0.379 & 0.422 & 0.462 & 0.468 & 0.463 & 0.474 & 0.515 & 0.511 & 0.427 & 0.445 \\ 

\multicolumn{1}{c}{}  & \textbf{Avg.} & 4.388	& 1.013	& 0.406	& 0.441	& 0.339	& 0.385	& 0.355	& 0.395	& 0.382	& 0.412	& 0.431	& 0.447	& 0.330	& 0.379	& 0.414	& 0.427	& 0.437	& 0.449	& 0.450	& 0.459	& 0.383	& 0.407 \\

\midrule
\multicolumn{1}{c}{\multirow{5}{*}{\textbf{ETTm1}}} &  \textbf{96} & 0.763 & 0.571 & \multicolumn{1}{r}{1.266} & \multicolumn{1}{r|}{0.697} & 0.292 & 0.339 & 0.292 & 0.346 & 0.33 & 0.372 & 0.299 & 0.343 & 0.29 & 0.342 & 0.338 & 0.375 & 0.379 & 0.419 & 0.505 & 0.475 & 0.334 & 0.368 \\ 
\multicolumn{1}{c}{}  & \textbf{192} & 1.401 & 0.752 & 1.304 & 0.714 & 0.33 & 0.363 & 0.332 & 0.372 & 0.371 & 0.394 & 0.335 & 0.365 & 0.332 & 0.369 & 0.374 & 0.387 & 0.426 & 0.441 & 0.553 & 0.496 & 0.377 & 0.391 \\ 
\multicolumn{1}{c}{} & \textbf{336} & 2.706 & 1.019 & 1.385 & 0.744 & 0.36 & 0.379 & 0.366 & 0.394 & 0.398 & 0.409 & 0.369 & 0.386 & 0.366 & 0.392 & 0.41 & 0.411 & 0.445 & 0.459 & 0.621 & 0.537 & 0.426 & 0.42 \\ 
\multicolumn{1}{c}{}  & \textbf{720} & 7.549 & 1.547 & 1.400 & 0.762 & 0.413 & 0.406 & 0.417 & 0.421 & 0.454 & 0.44 & 0.425 & 0.421 & 0.416 & 0.42 & 0.478 & 0.45 & 0.543 & 0.49 & 0.671 & 0.561 & 0.491 & 0.459 \\

\multicolumn{1}{c}{}  & \textbf{Avg.} & 3.105	& 0.972	& 1.339	& 0.729	& 0.349	& 0.372	& 0.352	& 0.383	& 0.388	& 0.404	& 0.357	& 0.379	& 0.351	& 0.381	& 0.400	& 0.406	& 0.448	& 0.452	& 0.588	& 0.517	& 0.407	& 0.410 \\ 

\midrule
\multicolumn{1}{c}{\multirow{5}{*}{\textbf{ETTm2}}} &  \textbf{96} & 0.500 & 0.411 & \multicolumn{1}{r}{0.188} & \multicolumn{1}{r|}{0.289} & 0.16 & 0.247 & 0.173 & 0.262 & 0.192 & 0.281 & 0.167 & 0.269 & 0.165 & 0.255 & 0.187 & 0.267 & 0.203 & 0.287 & 0.255 & 0.339 & 0.18 & 0.264 \\ 
\multicolumn{1}{c}{}  & \textbf{192} & 1.093 & 0.566 & 0.213 & 0.307 & 0.212 & 0.287 & 0.229 & 0.301 & 0.245 & 0.317 & 0.224 & 0.303 & 0.22 & 0.292 & 0.249 & 0.309 & 0.269 & 0.328 & 0.281 & 0.34 & 0.25 & 0.309 \\ 
\multicolumn{1}{c}{}  & \textbf{336} & 2.465 & 0.796 & 0.246 & 0.328 & 0.264 & 0.319 & 0.286 & 0.341 & 0.302 & 0.352 & 0.281 & 0.342 & 0.274 & 0.329 & 0.321 & 0.351 & 0.325 & 0.366 & 0.339 & 0.372 & 0.311 & 0.348 \\ 
\multicolumn{1}{c}{}  & \textbf{720} & 10.480 & 1.280 & 0.301 & 0.361 & 0.355 & 0.376 & 0.378 & 0.401 & 0.399 & 0.408 & 0.397 & 0.421 & 0.362 & 0.385 & 0.408 & 0.403 & 0.421 & 0.415 & 0.433 & 0.432 & 0.412 & 0.407 \\ 

\multicolumn{1}{c}{}  & \textbf{Avg.} & 3.634	& 0.763	& 0.237	& 0.321	& 0.248	& 0.307	& 0.267	& 0.326	& 0.285	& 0.340	& 0.267	& 0.334	& 0.255	& 0.315	& 0.291	& 0.333	& 0.305	& 0.349	& 0.327	& 0.371	& 0.288	& 0.332 \\ 

\midrule
\multicolumn{1}{c}{\multirow{5}{*}{\textbf{ECL}}} &  \textbf{96} & 0.575 & 0.534 & 1.629 & 0.976 & 0.131 & 0.225 & 0.139 & 0.238 & 0.138 & 0.234 & 0.14 & 0.237 & 0.129 & 0.222 & 0.168 & 0.272 & 0.193 & 0.308 & 0.201 & 0.317 & 0.148 & 0.24 \\ 
\multicolumn{1}{c}{}  & \textbf{192} & 0.886 & 0.622 & 1.600 & 0.969 & 0.151 & 0.245 & 0.153 & 0.251 & 0.152 & 0.247 & 0.153 & 0.249 & 0.157 & 0.24 & 0.184 & 0.289 & 0.201 & 0.315 & 0.222 & 0.334 & 0.162 & 0.253 \\ 
\multicolumn{1}{c}{}  & \textbf{336} & 1.600 & 0.776 & 1.613 & 0.974 & 0.162 & 0.254 & 0.169 & 0.266 & 0.168 & 0.263 & 0.169 & 0.267 & 0.163 & 0.259 & 0.198 & 0.3 & 0.214 & 0.329 & 0.231 & 0.338 & 0.178 & 0.269 \\ 
\multicolumn{1}{c}{}  & \textbf{720} & 4.014 & 1.098 & 1.651 & 0.988 & 0.192 & 0.284 & 0.206 & 0.297 & 0.207 & 0.295 & 0.203 & 0.301 & 0.197 & 0.29 & 0.22 & 0.32 & 0.246 & 0.355 & 0.254 & 0.361 & 0.225 & 0.317 \\ 

\multicolumn{1}{c}{}  & \textbf{Avg.} & 1.769	& 0.757	& 1.623	& 0.977	& 0.159	& 0.252	& 0.167	& 0.263	& 0.166	& 0.260	& 0.166	& 0.264	& 0.162	& 0.253	& 0.193	& 0.295	& 0.214	& 0.327	& 0.227	& 0.338	& 0.178	& 0.270 \\ 

\midrule
\multicolumn{1}{c}{\multirow{5}{*}{\textbf{ILI}}} &  \textbf{24} & 2.714 & 1.082 & \multicolumn{1}{c}{5.230} & \multicolumn{1}{c|}{1.743} & \multicolumn{1}{c}{--}	& \multicolumn{1}{c|}{--}	 & 2.063& 0.881 & 2.723	& 1.099 & 2.215 & 1.081 & 1.319 & 0.75 & 2.317 & 0.934 & 2.624 & 1.095 & 2.906 & 1.182 & \multicolumn{1}{c}{--} & \multicolumn{1}{c}{--} \\ 

\multicolumn{1}{c}{}  & \textbf{36} & 2.531 & 1.028 & \multicolumn{1}{r}{4.748} & \multicolumn{1}{r|}{1.617} & \multicolumn{1}{c}{--}	& \multicolumn{1}{c|}{--}	 & 1.868	& 0.892 & 2.027	& 0.966 & 1.963 & 0.963 & 1.579 & 0.87 & 1.972 & 0.92 & 2.516 & 1.021 & 2.585 & 1.038 & \multicolumn{1}{c}{--} & \multicolumn{1}{c}{--} \\ 

\multicolumn{1}{c}{}  & \textbf{48} & 2.314 & 0.981 & \multicolumn{1}{r}{4.318} & \multicolumn{1}{r|}{1.521} & \multicolumn{1}{c}{--}	& \multicolumn{1}{c|}{--}	 & 1.79	& 0.884 & 2.206	& 1.022 & 2.13 & 1.024 & 1.553 & 0.815 & 2.238 & 0.94 & 2.505 & 1.041 & 3.024 & 1.145 & \multicolumn{1}{c}{--} & \multicolumn{1}{c}{--} \\ 

\multicolumn{1}{c}{}  & \textbf{60} & 2.135 & 0.968 & \multicolumn{1}{r}{4.316} & \multicolumn{1}{r|}{1.513} & \multicolumn{1}{c}{--} & \multicolumn{1}{c|}{--} & 1.979	& 0.957 & 1.976	& 0.983 & 2.368 & 1.096 & 1.47 & 0.788 & 2.027 & 0.928 & 2.742 & 1.122 & 2.761 & 1.114 & \multicolumn{1}{c}{--} & \multicolumn{1}{c}{--} \\ 

\multicolumn{1}{c}{}  & \textbf{Avg.} & 2.423	& 1.015	& 4.653	& 1.599	& \multicolumn{1}{c}{--}	& \multicolumn{1}{c|}{--}	& 1.925	& 0.904	& 2.233	& 1.018	& 2.169	& 1.041	& 1.480	& 0.806	& 2.139	& 0.931	& 2.597	& 1.070	& 2.819	& 1.120	& \multicolumn{1}{c}{--} & \multicolumn{1}{c}{--} \\ 

\midrule
\multicolumn{1}{c}{\multirow{5}{*}{\textbf{Traffic}}} &  \textbf{96} & 1.223 & 0.700 & 3.059 & 1.230 & 0.378 & 0.25 & 0.388 & 0.282 & 0.39 & 0.272 & 0.41 & 0.282 & 0.36 & 0.249 & 0.593 & 0.321 & 0.587 & 0.366 & 0.613 & 0.388 & 0.395 & 0.268 \\ 
\multicolumn{1}{c}{}  & \textbf{192} & 1.454 & 0.766 & 2.980 & 1.214 & 0.384 & 0.248 & 0.407 & 0.29 & 0.403 & 0.276 & 0.423 & 0.287 & 0.379 & 0.256 & 0.617 & 0.336 & 0.604 & 0.373 & 0.616 & 0.382 & 0.417 & 0.276 \\ 
\multicolumn{1}{c}{}  & \textbf{336} & 2.011 & 0.897 & 2.994 & 1.218 & 0.393 & 0.255 & 0.412 & 0.294 & 0.413 & 0.28 & 0.436 & 0.296 & 0.392 & 0.264 & 0.629 & 0.336 & 0.621 & 0.383 & 0.622 & 0.337 & 0.433 & 0.283 \\ 
\multicolumn{1}{c}{}  & \textbf{720} & 3.349 & 1.146 & 2.951 & 1.211 & 0.434 & 0.276 & 0.45 & 0.312 & 0.447 & 0.298 & 0.466 & 0.315 & 0.432 & 0.286 & 0.64 & 0.35 & 0.626 & 0.382 & 0.66 & 0.408 & 0.467 & 0.302 \\ 

\multicolumn{1}{c}{}  & \textbf{Avg.} & 2.009	& 0.877	& 2.996	& 1.218	& 0.397	& 0.257	& 0.414	& 0.295	& 0.413	& 0.282	& 0.434	& 0.295	& 0.391	& 0.264	& 0.620	& 0.336	& 0.610	& 0.376	& 0.628	& 0.379	& 0.428	& 0.282 \\ 

\midrule
\multicolumn{2}{c|}{\textbf{Average}}   & 2.749	& 0.881	& 1.685	& 0.820	& 0.265	& 0.281	& 0.518	& 0.408	& 0.573	& 0.432	& 0.562	& 0.437	& 0.451	& 0.386	& 0.597	& 0.433	& 0.670	& 0.480	& 0.734	& 0.507	& 0.300	& 0.303 \\ 

\bottomrule
\end{tabular}%
}
\caption{Supervised long-term multivariate forecasting. Results for GPT2(6)-adapter (GPT(6)-A), GPT2(6)-frozen (GLT(6)-F), GPT2(0), DLinear, PatchTST, TimesNet, FEDformer, and Autoformer were taken from \cite{zhou2023one}. For ILI dataset, we took the results from \cite{Yuqietal-2023-PatchTST} and \cite{wu2022timesnet}; ITransformer metrics are cited form its paper \cite{liu2023itransformer}. Missing values, denoted by "--", correspond to the cases where there are too few observations for supervised learning. The "Ours" and "ForecastPFN" columns showcase the results for the zero-shot setup alongside the respective synthetic datasets.Our approach performs most effectively with short time sequences, surpassing both the zero-shot method ForecastPFN and supervised methodologies, particularly notable in cases involving short ILI dataset.}
\label{table:sup}
\end{table*}

\begin{table*}[h!]
\resizebox{\textwidth}{!}{%
\begin{tabular}{ll|rrr|rrr|rrr||rrr|rrr||rrr|rrr|rrr}
\toprule
\textbf{} & \textbf{} & \multicolumn{3}{c|}{\textbf{PatchTST}} & \multicolumn{3}{c|}{\textbf{GPT2(6)}} & \multicolumn{3}{c||}{\textbf{DLinear}} & \multicolumn{3}{c|}{\textbf{FPFN}} & \multicolumn{3}{c||}{\textbf{Ours}} & \multicolumn{3}{c|}{\textbf{PatchTST sup.}} & \multicolumn{3}{c|}{\textbf{GPT2(6) sup.}} & \multicolumn{3}{c}{\textbf{DLinear sup.}} \\
\textbf{Source} & \textbf{Target} & \multicolumn{1}{l}{\textbf{MSE}} & \multicolumn{1}{l}{\textbf{MAE}} & \multicolumn{1}{l|}{\textbf{SMAPE}} & \multicolumn{1}{l}{\textbf{MSE}} & \multicolumn{1}{l}{\textbf{MAE}} & \multicolumn{1}{l|}{\textbf{SMAPE}} & \multicolumn{1}{l}{\textbf{MSE}} & \multicolumn{1}{l}{\textbf{MAE}} & \multicolumn{1}{l||}{\textbf{SMAPE}} & \multicolumn{1}{l}{\textbf{MSE}} & \multicolumn{1}{l}{\textbf{MAE}} & \multicolumn{1}{l|}{\textbf{SMAPE}} & \multicolumn{1}{l}{\textbf{MSE}} & \multicolumn{1}{l}{\textbf{MAE}} & \multicolumn{1}{l||}{\textbf{SMAPE}} & \multicolumn{1}{l}{\textbf{MSE}} & \multicolumn{1}{l}{\textbf{MAE}} & \multicolumn{1}{l|}{\textbf{SMAPE}} & \multicolumn{1}{l}{\textbf{MSE}} & \multicolumn{1}{l}{\textbf{MAE}} & \multicolumn{1}{l|}{\textbf{SMAPE}} & \multicolumn{1}{l}{\textbf{MSE}} & \multicolumn{1}{l}{\textbf{MAE}} & \multicolumn{1}{l}{\textbf{SMAPE}} \\
\midrule
\textbf{M3 Y} & Weather & 1.72 & 0.84 & 118.34 & 1.04 & 0.59 & 93.10 & 0.25 & 0.35 & 93.37 & 0.14 & 0.18 & 50.19 & 0.15 & 0.20 & 51.91 & 0.07 & 0.11 & 32.68 & 0.06 & 0.08 & 26.70 & 0.06 & 0.06 & 29.32 \\
 & ETTh1 & 9.71 & 2.50 & 137.42 & 4.81 & 1.70 & 122.31 & 1.06 & 0.71 & 107.38 & 1.54 & 0.77 & 92.92 & 0.61 & 0.54 & 91.48 & 0.26 & 0.32 & 65.48 & 0.22 & 0.30 & 62.29 & 0.24 & 0.24 & 63.41 \\
 & ETTh2 & 3.08 & 1.43 & 142.11 & 1.60 & 0.95 & 113.53 & 0.97 & 0.71 & 90.02 & 0.22 & 0.31 & 70.39 & 0.37 & 0.41 & 58.73 & 0.11 & 0.22 & 40.13 & 0.11 & 0.21 & 38.26 & 0.11 & 0.21 & 38.03 \\
 & ETTm1 & 8.31 & 2.16 & 127.60 & 4.54 & 1.55 & 115.54 & 0.64 & 0.56 & 91.59 & 1.14 & 0.63 & 83.19 & 0.41 & 0.43 & 76.74 & 0.09 & 0.19 & 46.53 & 0.10 & 0.20 & 47.83 & 0.10 & 0.10 & 47.88 \\
 & ETTm2 & 1.81 & 1.07 & 121.57 & 1.01 & 0.75 & 98.09 & 0.88 & 0.67 & 85.44 & 0.14 & 0.25 & 61.14 & 0.18 & 0.27 & 45.90 & 0.06 & 0.14 & 27.43 & 0.06 & 0.14 & 27.66 & 0.06 & 0.14 & 27.57 \\
 & ECL & 11.78 & 3.16 & 160.77 & 5.75 & 2.14 & 149.76 & 1.22 & 0.83 & 115.64 & 1.64 & 0.99 & 134.15 & 0.45 & 0.48 & 83.43 & 0.15 & 0.25 & 51.51 & 0.11 & 0.21 & 43.19 & 0.12 & 0.21 & 44.61 \\
 & ILI & 26.36 & 4.24 & 120.66 & 16.31 & 3.23 & 114.41 & 4.51 & 1.65 & 116.25 & 4.13 & 1.42 & 102.92 & 3.06 & 1.16 & 87.48 & 1.03 & 0.64 & 61.24 & 1.03 & 0.64 & 60.10 & 0.90 & 0.59 & 47.13 \\
 & Traffic & 20.22 & 3.93 & 169.82 & 10.41 & 2.74 & 162.21 & 2.08 & 0.99 & 128.59 & 3.14 & 1.27 & 155.49 & 1.23 & 0.68 & 102.64 & 0.32 & 0.23 & 43.03 & 0.36 & 0.26 & 45.89 & 0.54 & 0.36 & 62.10 \\
\midrule

\textbf{M3 Q} & Weather & 0.80 & 0.54 & 95.57 & 0.34 & 0.33 & 69.88 & 0.28 & 0.36 & 90.99 & 0.14 & 0.18 & 50.19 & 0.15 & 0.20 & 51.91 & 0.07 & 0.11 & 32.68 & 0.06 & 0.08 & 26.70 & 0.06 & 0.06 & 29.32 \\
 & ETTh1 & 4.29 & 1.60 & 119.57 & 1.50 & 0.82 & 94.86 & 1.05 & 0.72 & 114.06 & 1.54 & 0.77 & 92.92 & 0.61 & 0.54 & 91.48 & 0.26 & 0.32 & 65.48 & 0.22 & 0.30 & 62.29 & 0.24 & 0.24 & 63.41 \\
 & ETTh2 & 1.40 & 0.92 & 112.49 & 0.51 & 0.49 & 72.24 & 1.13 & 0.76 & 93.46 & 0.22 & 0.31 & 70.39 & 0.37 & 0.41 & 58.73 & 0.11 & 0.22 & 40.13 & 0.11 & 0.21 & 38.26 & 0.11 & 0.21 & 38.03 \\
 & ETTm1 & 3.77 & 1.39 & 111.45 & 1.19 & 0.72 & 90.48 & 0.80 & 0.62 & 101.11 & 1.14 & 0.63 & 83.19 & 0.41 & 0.43 & 76.74 & 0.09 & 0.19 & 46.53 & 0.10 & 0.20 & 47.83 & 0.10 & 0.10 & 47.88 \\
 & ETTm2 & 0.84 & 0.69 & 94.13 & 0.30 & 0.37 & 59.79 & 1.05 & 0.72 & 88.58 & 0.14 & 0.25 & 61.14 & 0.18 & 0.27 & 45.90 & 0.06 & 0.14 & 27.43 & 0.06 & 0.14 & 27.66 & 0.06 & 0.14 & 27.57 \\
 & ECL & 5.26 & 2.02 & 146.84 & 1.74 & 1.02 & 122.38 & 1.21 & 0.88 & 136.72 & 1.64 & 0.99 & 134.15 & 0.45 & 0.48 & 83.43 & 0.15 & 0.25 & 51.51 & 0.11 & 0.21 & 43.19 & 0.12 & 0.21 & 44.61 \\
 & ILI & 13.25 & 2.88 & 109.51 & 6.69 & 2.01 & 103.32 & 4.49 & 1.62 & 119.52 & 4.13 & 1.42 & 102.92 & 3.06 & 1.16 & 87.48 & 1.03 & 0.64 & 61.24 & 1.03 & 0.64 & 60.10 & 0.90 & 0.59 & 47.13 \\
 & Traffic & 9.10 & 2.58 & 160.32 & 3.42 & 1.47 & 146.68 & 2.00 & 1.00 & 145.44 & 3.14 & 1.27 & 155.49 & 1.23 & 0.68 & 102.64 & 0.32 & 0.23 & 43.03 & 0.36 & 0.26 & 45.89 & 0.54 & 0.36 & 62.10 \\
 \midrule

 \textbf{M3 M} & Weather & 0.23 & 0.29 & 67.77 & 0.08 & 0.12 & 34.71 & 0.40 & 0.53 & 118.04 & 0.14 & 0.18 & 50.19 & 0.15 & 0.20 & 51.91 & 0.07 & 0.11 & 32.68 & 0.06 & 0.08 & 26.70 & 0.06 & 0.06 & 29.32 \\
 & ETTh1 & 1.01 & 0.63 & 88.33 & 1.02 & 0.65 & 101.23 & 1.70 & 0.94 & 111.22 & 1.54 & 0.77 & 92.92 & 0.61 & 0.54 & 91.48 & 0.26 & 0.32 & 65.48 & 0.22 & 0.30 & 62.29 & 0.24 & 0.24 & 63.41 \\
 & ETTh2 & 0.37 & 0.41 & 63.95 & 0.21 & 0.32 & 54.52 & 0.55 & 0.60 & 86.67 & 0.22 & 0.31 & 70.39 & 0.37 & 0.41 & 58.73 & 0.11 & 0.22 & 40.13 & 0.11 & 0.21 & 38.26 & 0.11 & 0.21 & 38.03 \\
 & ETTm1 & 1.03 & 0.61 & 87.10 & 0.44 & 0.40 & 76.27 & 0.60 & 0.61 & 90.26 & 1.14 & 0.63 & 83.19 & 0.41 & 0.43 & 76.74 & 0.09 & 0.19 & 46.53 & 0.10 & 0.20 & 47.83 & 0.10 & 0.10 & 47.88 \\
 & ETTm2 & 0.25 & 0.34 & 56.48 & 0.09 & 0.19 & 36.65 & 0.40 & 0.54 & 82.82 & 0.14 & 0.25 & 61.14 & 0.18 & 0.27 & 45.90 & 0.06 & 0.14 & 27.43 & 0.06 & 0.14 & 27.66 & 0.06 & 0.14 & 27.57 \\
 & ECL & 1.22 & 0.84 & 121.49 & 1.66 & 1.05 & 142.07 & 2.44 & 1.26 & 146.09 & 1.64 & 0.99 & 134.15 & 0.45 & 0.48 & 83.43 & 0.15 & 0.25 & 51.51 & 0.11 & 0.21 & 43.19 & 0.12 & 0.21 & 44.61 \\
 & ILI & 4.68 & 1.70 & 104.39 & 2.96 & 1.11 & 85.72 & 2.99 & 1.20 & 90.86 & 4.13 & 1.42 & 102.92 & 3.06 & 1.16 & 87.48 & 1.03 & 0.64 & 61.24 & 1.03 & 0.64 & 60.10 & 0.90 & 0.59 & 47.13 \\
 & Traffic & 2.10 & 1.11 & 142.18 & 2.56 & 1.21 & 156.81 & 3.52 & 1.44 & 158.59 & 3.14 & 1.27 & 155.49 & 1.23 & 0.68 & 102.64 & 0.32 & 0.23 & 43.03 & 0.36 & 0.26 & 45.89 & 0.54 & 0.36 & 62.10 \\
 \midrule

 \textbf{M3 O} & Weather & 0.25 & 0.27 & 62.66 & 0.43 & 0.37 & 72.38 & 0.16 & 0.23 & 62.12 & 0.14 & 0.18 & 50.19 & 0.15 & 0.20 & 51.91 & 0.07 & 0.11 & 32.68 & 0.06 & 0.08 & 26.70 & 0.06 & 0.06 & 29.32 \\
 & ETTh1 & 1.32 & 0.73 & 95.90 & 1.77 & 0.88 & 102.06 & 0.72 & 0.56 & 94.48 & 1.54 & 0.77 & 92.92 & 0.61 & 0.54 & 91.48 & 0.26 & 0.32 & 65.48 & 0.22 & 0.30 & 62.29 & 0.24 & 0.24 & 63.41 \\
 & ETTh2 & 0.48 & 0.46 & 64.73 & 0.64 & 0.52 & 72.15 & 0.29 & 0.37 & 56.91 & 0.22 & 0.31 & 70.39 & 0.37 & 0.41 & 58.73 & 0.11 & 0.22 & 40.13 & 0.11 & 0.21 & 38.26 & 0.11 & 0.21 & 38.03 \\
 & ETTm1 & 1.56 & 0.78 & 101.44 & 2.07 & 0.93 & 100.58 & 0.68 & 0.54 & 93.09 & 1.14 & 0.63 & 83.19 & 0.41 & 0.43 & 76.74 & 0.09 & 0.19 & 46.53 & 0.10 & 0.20 & 47.83 & 0.10 & 0.10 & 47.88 \\
 & ETTm2 & 0.29 & 0.36 & 54.66 & 0.43 & 0.42 & 63.25 & 0.20 & 0.31 & 52.31 & 0.14 & 0.25 & 61.14 & 0.18 & 0.27 & 45.90 & 0.06 & 0.14 & 27.43 & 0.06 & 0.14 & 27.66 & 0.06 & 0.14 & 27.57 \\
 & ECL & 1.65 & 0.98 & 127.67 & 2.07 & 1.13 & 127.55 & 0.88 & 0.74 & 121.12 & 1.64 & 0.99 & 134.15 & 0.45 & 0.48 & 83.43 & 0.15 & 0.25 & 51.51 & 0.11 & 0.21 & 43.19 & 0.12 & 0.21 & 44.61 \\
 & ILI & 7.79 & 2.07 & 126.33 & 4.38 & 1.60 & 109.70 & 4.32 & 1.52 & 114.62 & 4.13 & 1.42 & 102.92 & 3.06 & 1.16 & 87.48 & 1.03 & 0.64 & 61.24 & 1.03 & 0.64 & 60.10 & 0.90 & 0.59 & 47.13 \\
 & Traffic & 2.82 & 1.25 & 146.12 & 2.27 & 1.12 & 141.80 & 1.61 & 0.87 & 135.83 & 3.14 & 1.27 & 155.49 & 1.23 & 0.68 & 102.64 & 0.32 & 0.23 & 43.03 & 0.36 & 0.26 & 45.89 & 0.54 & 0.36 & 62.10 \\
 \bottomrule
 
\end{tabular}%
}
\caption{
Zero-shot model tranferability results for short-term forecasting from the M3 dataset with horizon $H = 6$. Subsets of the M3 dataset are denoted by the first letter: yearly data -- Y, quarterly -- Q, monthly -- M and O stands for Other; suffix sup denotes supervised forecasting of a given dataset and does not depend on source one; thus, quantitative results are repeated for each source dataset to ease the comparison. The results for ForecastPFN (FPFN) and our model remain identical across various source datasets due to their training on synthetic data without source dataset. A double line separates the columns with a pure zero-shot approach based on synthetics in the middle (Ours and ForecastPFN) and supervised models (denoted with \textit{sup.}). While zero-shot learning with synthetic data underperforms supervised methods, choosing a dataset source requires careful consideration. Otherwise, its application may exacerbate results in comparison to zero-shot approaches. While GPT2(6) demonstrates some of the most favorable results among zero-shot methods when trained on monthly data, its performance diminishes significantly when utilizing quarterly data.}
\label{table:zeroshot_m3_bench}

\end{table*}

\begin{table*}[h!]
\resizebox{\textwidth}{!}{%
\begin{tabular}{ll|rrr|rrr|rrr||rrr|rrr||rrr|rrr|rrr}
\toprule
\textbf{} & \textbf{} & \multicolumn{3}{c|}{\textbf{PatchTST}} & \multicolumn{3}{c|}{\textbf{GPT2(6)}} & \multicolumn{3}{c||}{\textbf{DLinear}} & \multicolumn{3}{c|}{\textbf{FPFN}} & \multicolumn{3}{c||}{\textbf{Ours}} & \multicolumn{3}{c|}{\textbf{PatchTST sup.}} & \multicolumn{3}{c|}{\textbf{GPT2(6) sup.}} & \multicolumn{3}{c}{\textbf{DLinear sup.}} \\
\textbf{Source} & \textbf{Target} & \multicolumn{1}{l}{\textbf{MSE}} & \multicolumn{1}{l}{\textbf{MAE}} & \multicolumn{1}{l|}{\textbf{SMAPE}} & \multicolumn{1}{l}{\textbf{MSE}} & \multicolumn{1}{l}{\textbf{MAE}} & \multicolumn{1}{l|}{\textbf{SMAPE}} & \multicolumn{1}{l}{\textbf{MSE}} & \multicolumn{1}{l}{\textbf{MAE}} & \multicolumn{1}{l||}{\textbf{SMAPE}} & \multicolumn{1}{l}{\textbf{MSE}} & \multicolumn{1}{l}{\textbf{MAE}} & \multicolumn{1}{l|}{\textbf{SMAPE}} & \multicolumn{1}{l}{\textbf{MSE}} & \multicolumn{1}{l}{\textbf{MAE}} & \multicolumn{1}{l||}{\textbf{SMAPE}} & \multicolumn{1}{l}{\textbf{MSE}} & \multicolumn{1}{l}{\textbf{MAE}} & \multicolumn{1}{l|}{\textbf{SMAPE}} & \multicolumn{1}{l}{\textbf{MSE}} & \multicolumn{1}{l}{\textbf{MAE}} & \multicolumn{1}{l|}{\textbf{SMAPE}} & \multicolumn{1}{l}{\textbf{MSE}} & \multicolumn{1}{l}{\textbf{MAE}} & \multicolumn{1}{l}{\textbf{SMAPE}} \\
\midrule
\textbf{M4 Y} & Weather & 1.54 & 0.71 & 105.17 & 0.08 & 0.10 & 31.45 & 0.09 & 0.09 & 29.54 & 0.14 & 0.18 & 50.19 & 0.15 & 0.20 & 51.91 & 0.07 & 0.11 & 32.68 & 0.06 & 0.08 & 26.70 & 0.06 & 0.06 & 29.32 \\
 & ETTh1 & 11.31 & 2.67 & 137.09 & 0.83 & 0.56 & 87.21 & 1.28 & 0.66 & 91.18 & 1.54 & 0.77 & 92.92 & 0.61 & 0.54 & 91.48 & 0.26 & 0.32 & 65.48 & 0.22 & 0.30 & 62.29 & 0.24 & 0.24 & 63.41 \\
 & ETTh2 & 2.41 & 1.28 & 136.03 & 0.17 & 0.27 & 48.02 & 0.24 & 0.32 & 51.52 & 0.22 & 0.31 & 70.39 & 0.37 & 0.41 & 58.73 & 0.11 & 0.22 & 40.13 & 0.11 & 0.21 & 38.26 & 0.11 & 0.21 & 38.03 \\
 & ETTm1 & 5.10 & 1.62 & 116.40 & 0.34 & 0.35 & 65.02 & 0.23 & 0.29 & 59.06 & 1.14 & 0.63 & 83.19 & 0.41 & 0.43 & 76.74 & 0.09 & 0.19 & 46.53 & 0.10 & 0.20 & 47.83 & 0.10 & 0.10 & 47.88 \\
 & ETTm2 & 1.35 & 0.90 & 109.92 & 0.08 & 0.18 & 33.72 & 0.10 & 0.19 & 33.61 & 0.14 & 0.25 & 61.14 & 0.18 & 0.27 & 45.90 & 0.06 & 0.14 & 27.43 & 0.06 & 0.14 & 27.66 & 0.06 & 0.14 & 27.57 \\
 & ECL & 14.23 & 3.47 & 162.70 & 1.04 & 0.77 & 110.04 & 1.57 & 0.90 & 104.82 & 1.64 & 0.99 & 134.15 & 0.45 & 0.48 & 83.43 & 0.15 & 0.25 & 51.51 & 0.11 & 0.21 & 43.19 & 0.12 & 0.21 & 44.61 \\
 & ILI & 29.30 & 4.37 & 119.23 & 2.32 & 0.97 & 85.92 & 1.88 & 0.83 & 69.06 & 4.13 & 1.42 & 102.92 & 3.06 & 1.16 & 87.48 & 1.03 & 0.64 & 61.24 & 1.03 & 0.64 & 60.10 & 0.90 & 0.59 & 47.13 \\
 & Traffic & 25.16 & 4.39 & 171.85 & 2.04 & 0.96 & 128.19 & 2.76 & 1.08 & 119.22 & 3.14 & 1.27 & 155.49 & 1.23 & 0.68 & 102.64 & 0.32 & 0.23 & 43.03 & 0.36 & 0.26 & 45.89 & 0.54 & 0.36 & 62.10 \\
 \midrule
 
\textbf{M4 Q} & Weather & 0.27 & 0.31 & 70.66 & 0.07 & 0.10 & 31.40 & 0.07 & 0.09 & 29.79 & 0.14 & 0.18 & 50.19 & 0.15 & 0.20 & 51.91 & 0.07 & 0.11 & 32.68 & 0.06 & 0.08 & 26.70 & 0.06 & 0.06 & 29.32 \\
 & ETTh1 & 1.36 & 0.77 & 93.50 & 0.87 & 0.57 & 91.59 & 0.81 & 0.53 & 86.47 & 1.54 & 0.77 & 92.92 & 0.61 & 0.54 & 91.48 & 0.26 & 0.32 & 65.48 & 0.22 & 0.30 & 62.29 & 0.24 & 0.24 & 63.41 \\
 & ETTh2 & 0.43 & 0.46 & 70.64 & 0.16 & 0.27 & 48.70 & 0.15 & 0.26 & 45.40 & 0.22 & 0.31 & 70.39 & 0.37 & 0.41 & 58.73 & 0.11 & 0.22 & 40.13 & 0.11 & 0.21 & 38.26 & 0.11 & 0.21 & 38.03 \\
 & ETTm1 & 1.41 & 0.71 & 89.78 & 0.25 & 0.31 & 61.70 & 0.20 & 0.27 & 58.72 & 1.14 & 0.63 & 83.19 & 0.41 & 0.43 & 76.74 & 0.09 & 0.19 & 46.53 & 0.10 & 0.20 & 47.83 & 0.10 & 0.10 & 47.88 \\
 & ETTm2 & 0.30 & 0.38 & 61.30 & 0.08 & 0.18 & 33.50 & 0.07 & 0.16 & 31.96 & 0.14 & 0.25 & 61.14 & 0.18 & 0.27 & 45.90 & 0.06 & 0.14 & 27.43 & 0.06 & 0.14 & 27.66 & 0.06 & 0.14 & 27.57 \\
 & ECL & 1.71 & 1.01 & 123.01 & 1.06 & 0.79 & 116.54 & 0.93 & 0.69 & 101.11 & 1.64 & 0.99 & 134.15 & 0.45 & 0.48 & 83.43 & 0.15 & 0.25 & 51.51 & 0.11 & 0.21 & 43.19 & 0.12 & 0.21 & 44.61 \\
 & ILI & 5.75 & 1.90 & 105.40 & 1.92 & 0.91 & 80.96 & 1.64 & 0.79 & 74.48 & 4.13 & 1.42 & 102.92 & 3.06 & 1.16 & 87.48 & 1.03 & 0.64 & 61.24 & 1.03 & 0.64 & 60.10 & 0.90 & 0.59 & 47.13 \\
 & Traffic & 3.01 & 1.39 & 145.38 & 1.92 & 0.94 & 133.55 & 1.79 & 0.84 & 115.12 & 3.14 & 1.27 & 155.49 & 1.23 & 0.68 & 102.64 & 0.32 & 0.23 & 43.03 & 0.36 & 0.26 & 45.89 & 0.54 & 0.36 & 62.10 \\
 \midrule
 
\textbf{M4 M} & Weather & 0.16 & 0.22 & 57.53 & 0.07 & 0.08 & 25.24 & 0.07 & 0.09 & 31.96 & 0.14 & 0.18 & 50.19 & 0.15 & 0.20 & 51.91 & 0.07 & 0.11 & 32.68 & 0.06 & 0.08 & 26.70 & 0.06 & 0.06 & 29.32 \\
 & ETTh1 & 0.72 & 0.53 & 90.28 & 0.67 & 0.52 & 85.26 & 0.66 & 0.52 & 86.61 & 1.54 & 0.77 & 92.92 & 0.61 & 0.54 & 91.48 & 0.26 & 0.32 & 65.48 & 0.22 & 0.30 & 62.29 & 0.24 & 0.24 & 63.41 \\
 & ETTh2 & 0.27 & 0.35 & 56.20 & 0.15 & 0.25 & 45.33 & 0.14 & 0.25 & 46.91 & 0.22 & 0.31 & 70.39 & 0.37 & 0.41 & 58.73 & 0.11 & 0.22 & 40.13 & 0.11 & 0.21 & 38.26 & 0.11 & 0.21 & 38.03 \\
 & ETTm1 & 1.04 & 0.62 & 95.84 & 0.18 & 0.26 & 55.25 & 0.19 & 0.26 & 56.83 & 1.14 & 0.63 & 83.19 & 0.41 & 0.43 & 76.74 & 0.09 & 0.19 & 46.53 & 0.10 & 0.20 & 47.83 & 0.10 & 0.10 & 47.88 \\
 & ETTm2 & 0.19 & 0.30 & 52.66 & 0.07 & 0.16 & 29.56 & 0.07 & 0.16 & 33.19 & 0.14 & 0.25 & 61.14 & 0.18 & 0.27 & 45.90 & 0.06 & 0.14 & 27.43 & 0.06 & 0.14 & 27.66 & 0.06 & 0.14 & 27.57 \\
 & ECL & 0.87 & 0.75 & 132.01 & 0.82 & 0.68 & 103.35 & 0.86 & 0.71 & 111.45 & 1.64 & 0.99 & 134.15 & 0.45 & 0.48 & 83.43 & 0.15 & 0.25 & 51.51 & 0.11 & 0.21 & 43.19 & 0.12 & 0.21 & 44.61 \\
 & ILI & 4.29 & 1.57 & 109.58 & 1.55 & 0.76 & 64.52 & 2.05 & 0.89 & 74.26 & 4.13 & 1.42 & 102.92 & 3.06 & 1.16 & 87.48 & 1.03 & 0.64 & 61.24 & 1.03 & 0.64 & 60.10 & 0.90 & 0.59 & 47.13 \\
 & Traffic & 1.48 & 0.87 & 144.60 & 1.58 & 0.85 & 122.37 & 1.64 & 0.89 & 128.73 & 3.14 & 1.27 & 155.49 & 1.23 & 0.68 & 102.64 & 0.32 & 0.23 & 43.03 & 0.36 & 0.26 & 45.89 & 0.54 & 0.36 & 62.10 \\
 \midrule
 
\textbf{M4 W} & Weather & 0.14 & 0.21 & 54.15 & 0.08 & 0.10 & 29.74 & 0.08 & 0.10 & 30.69 & 0.14 & 0.18 & 50.19 & 0.15 & 0.20 & 51.91 & 0.07 & 0.11 & 32.68 & 0.06 & 0.08 & 26.70 & 0.06 & 0.06 & 29.32 \\
 & ETTh1 & 0.68 & 0.53 & 96.37 & 1.16 & 0.67 & 91.74 & 1.23 & 0.67 & 90.97 & 1.54 & 0.77 & 92.92 & 0.61 & 0.54 & 91.48 & 0.26 & 0.32 & 65.48 & 0.22 & 0.30 & 62.29 & 0.24 & 0.24 & 63.41 \\
 & ETTh2 & 0.24 & 0.32 & 54.11 & 0.19 & 0.29 & 49.31 & 0.19 & 0.29 & 49.64 & 0.22 & 0.31 & 70.39 & 0.37 & 0.41 & 58.73 & 0.11 & 0.22 & 40.13 & 0.11 & 0.21 & 38.26 & 0.11 & 0.21 & 38.03 \\
 & ETTm1 & 0.81 & 0.57 & 95.63 & 0.34 & 0.36 & 65.91 & 0.24 & 0.31 & 62.81 & 1.14 & 0.63 & 83.19 & 0.41 & 0.43 & 76.74 & 0.09 & 0.19 & 46.53 & 0.10 & 0.20 & 47.83 & 0.10 & 0.10 & 47.88 \\
 & ETTm2 & 0.17 & 0.28 & 50.22 & 0.09 & 0.19 & 34.42 & 0.09 & 0.18 & 34.26 & 0.14 & 0.25 & 61.14 & 0.18 & 0.27 & 45.90 & 0.06 & 0.14 & 27.43 & 0.06 & 0.14 & 27.66 & 0.06 & 0.14 & 27.57 \\
 & ECL & 0.82 & 0.75 & 137.58 & 1.31 & 0.89 & 106.98 & 1.29 & 0.85 & 103.19 & 1.64 & 0.99 & 134.15 & 0.45 & 0.48 & 83.43 & 0.15 & 0.25 & 51.51 & 0.11 & 0.21 & 43.19 & 0.12 & 0.21 & 44.61 \\
 & ILI & 3.70 & 1.42 & 109.22 & 0.93 & 0.60 & 56.27 & 1.03 & 0.63 & 62.81 & 4.13 & 1.42 & 102.92 & 3.06 & 1.16 & 87.48 & 1.03 & 0.64 & 61.24 & 1.03 & 0.64 & 60.10 & 0.90 & 0.59 & 47.13 \\
 & Traffic & 1.38 & 0.80 & 151.55 & 2.34 & 1.02 & 115.81 & 2.43 & 1.01 & 112.02 & 3.14 & 1.27 & 155.49 & 1.23 & 0.68 & 102.64 & 0.32 & 0.23 & 43.03 & 0.36 & 0.26 & 45.89 & 0.54 & 0.36 & 62.10 \\
 \midrule
 
\textbf{M4 D} & Weather & 0.18 & 0.22 & 52.23 & 0.08 & 0.08 & 25.77 & 0.07 & 0.07 & 24.15 & 0.14 & 0.18 & 50.19 & 0.15 & 0.20 & 51.91 & 0.07 & 0.11 & 32.68 & 0.06 & 0.08 & 26.70 & 0.06 & 0.06 & 29.32 \\
 & ETTh1 & 1.16 & 0.76 & 122.52 & 0.89 & 0.56 & 86.08 & 0.84 & 0.53 & 84.32 & 1.54 & 0.77 & 92.92 & 0.61 & 0.54 & 91.48 & 0.26 & 0.32 & 65.48 & 0.22 & 0.30 & 62.29 & 0.24 & 0.24 & 63.41 \\
 & ETTh2 & 0.37 & 0.42 & 59.28 & 0.18 & 0.27 & 44.49 & 0.17 & 0.25 & 43.24 & 0.22 & 0.31 & 70.39 & 0.37 & 0.41 & 58.73 & 0.11 & 0.22 & 40.13 & 0.11 & 0.21 & 38.26 & 0.11 & 0.21 & 38.03 \\
 & ETTm1 & 1.07 & 0.68 & 108.06 & 0.17 & 0.25 & 54.49 & 0.16 & 0.24 & 52.48 & 1.14 & 0.63 & 83.19 & 0.41 & 0.43 & 76.74 & 0.09 & 0.19 & 46.53 & 0.10 & 0.20 & 47.83 & 0.10 & 0.10 & 47.88 \\
 & ETTm2 & 0.19 & 0.29 & 48.53 & 0.08 & 0.16 & 28.35 & 0.07 & 0.15 & 27.66 & 0.14 & 0.25 & 61.14 & 0.18 & 0.27 & 45.90 & 0.06 & 0.14 & 27.43 & 0.06 & 0.14 & 27.66 & 0.06 & 0.14 & 27.57 \\
 & ECL & 1.53 & 1.00 & 134.66 & 1.12 & 0.74 & 98.03 & 0.94 & 0.67 & 94.87 & 1.64 & 0.99 & 134.15 & 0.45 & 0.48 & 83.43 & 0.15 & 0.25 & 51.51 & 0.11 & 0.21 & 43.19 & 0.12 & 0.21 & 44.61 \\
 & ILI & 6.38 & 1.72 & 122.06 & 1.57 & 0.76 & 65.77 & 1.41 & 0.67 & 58.17 & 4.13 & 1.42 & 102.92 & 3.06 & 1.16 & 87.48 & 1.03 & 0.64 & 61.24 & 1.03 & 0.64 & 60.10 & 0.90 & 0.59 & 47.13 \\
 & Traffic & 2.48 & 1.09 & 135.62 & 2.12 & 0.90 & 111.29 & 1.90 & 0.81 & 107.50 & 3.14 & 1.27 & 155.49 & 1.23 & 0.68 & 102.64 & 0.32 & 0.23 & 43.03 & 0.36 & 0.26 & 45.89 & 0.54 & 0.36 & 62.10 \\
 \midrule
 
\textbf{M4 H} & Weather & 0.17 & 0.24 & 59.95 & 0.09 & 0.13 & 36.85 & 0.09 & 0.11 & 32.37 & 0.14 & 0.18 & 50.19 & 0.15 & 0.20 & 51.91 & 0.07 & 0.11 & 32.68 & 0.06 & 0.08 & 26.70 & 0.06 & 0.06 & 29.32 \\
 & ETTh1 & 0.63 & 0.51 & 91.29 & 0.49 & 0.47 & 81.68 & 0.32 & 0.37 & 71.31 & 1.54 & 0.77 & 92.92 & 0.61 & 0.54 & 91.48 & 0.26 & 0.32 & 65.48 & 0.22 & 0.30 & 62.29 & 0.24 & 0.24 & 63.41 \\
 & ETTh2 & 0.28 & 0.34 & 55.75 & 0.13 & 0.24 & 43.27 & 0.13 & 0.23 & 40.99 & 0.22 & 0.31 & 70.39 & 0.37 & 0.41 & 58.73 & 0.11 & 0.22 & 40.13 & 0.11 & 0.21 & 38.26 & 0.11 & 0.21 & 38.03 \\
 & ETTm1 & 0.71 & 0.54 & 94.59 & 0.31 & 0.36 & 68.10 & 0.31 & 0.35 & 68.63 & 1.14 & 0.63 & 83.19 & 0.41 & 0.43 & 76.74 & 0.09 & 0.19 & 46.53 & 0.10 & 0.20 & 47.83 & 0.10 & 0.10 & 47.88 \\
 & ETTm2 & 0.19 & 0.29 & 50.53 & 0.08 & 0.19 & 36.09 & 0.09 & 0.19 & 35.64 & 0.14 & 0.25 & 61.14 & 0.18 & 0.27 & 45.90 & 0.06 & 0.14 & 27.43 & 0.06 & 0.14 & 27.66 & 0.06 & 0.14 & 27.57 \\
 & ECL & 0.71 & 0.68 & 125.25 & 0.32 & 0.42 & 74.14 & 0.19 & 0.30 & 58.09 & 1.64 & 0.99 & 134.15 & 0.45 & 0.48 & 83.43 & 0.15 & 0.25 & 51.51 & 0.11 & 0.21 & 43.19 & 0.12 & 0.21 & 44.61 \\
 & ILI & 4.38 & 1.50 & 110.54 & 4.58 & 1.61 & 101.41 & 3.66 & 1.47 & 101.67 & 4.13 & 1.42 & 102.92 & 3.06 & 1.16 & 87.48 & 1.03 & 0.64 & 61.24 & 1.03 & 0.64 & 60.10 & 0.90 & 0.59 & 47.13 \\
 & Traffic & 1.29 & 0.77 & 138.33 & 1.20 & 0.68 & 96.12 & 0.88 & 0.53 & 81.76 & 3.14 & 1.27 & 155.49 & 1.23 & 0.68 & 102.64 & 0.32 & 0.23 & 43.03 & 0.36 & 0.26 & 45.89 & 0.54 & 0.36 & 62.10 \\
 \bottomrule
 
\end{tabular}%
}
\caption{
Zero-shot model tranferability results for short-term forecasting from the M4 dataset with horizon $H = 6$. Subsets of the M3 dataset are denoted by the first latter: yearly data -- Y, quarterly -- Q, monthly -- M, weekly -- W, daily -- D, hourly -- H; suffix sup denotes supervised forecasting for a given dataset and does not depend on source one, thus quantitative results are repeated for each source dataset to ease the comparison. The results for ForecastPFN (FPFN) and our model remain identical across various source datasets due to their training on synthetic data without source dataset. A double line separates the columns with a pure zero-shot approach based on synthetics in the middle (Ours and ForecastPFN) and supervised models (denoted with \textit{sup.}). A comparable scenario arises when examining the preceding Table~\ref{table:zeroshot_m3_bench}, where the selection of the source dataset prominently impacts the model's performance. This observation is particularly evident for PatchTST. Conversely, for the M4 dataset, GPT2(6) and DLinear demonstrate the ability to maintain consistent quality in the transferability setup, suggesting that architecture also plays a significant role.}
\label{table:zeroshot_m4_bench}

\end{table*}

\begin{table*}[h!]
\resizebox{\textwidth}{!}{%
\begin{tabular}{ll|rrr|rrr|rrr||rrr|rrr}
\toprule

\textbf{} & \textbf{} & \multicolumn{3}{c|}{\textbf{PatchTST}} & \multicolumn{3}{c|}{\textbf{GPT2(6)}} & \multicolumn{3}{c||}{\textbf{DLinear}} & \multicolumn{3}{c|}{\textbf{FPFN}} & \multicolumn{3}{c}{\textbf{Ours}} \\
\textbf{Source} & \textbf{Target} & \multicolumn{1}{c}{\textbf{MSE}} & \multicolumn{1}{c}{\textbf{MAE}} & \multicolumn{1}{c|}{\textbf{SMAPE}} & \multicolumn{1}{c}{\textbf{MSE}} & \multicolumn{1}{l}{\textbf{MAE}} & \multicolumn{1}{c|}{\textbf{SMAPE}} & \multicolumn{1}{c}{\textbf{MSE}} & \multicolumn{1}{l}{\textbf{MAE}} & \multicolumn{1}{c||}{\textbf{SMAPE}} & \multicolumn{1}{c}{\textbf{MSE}} & \multicolumn{1}{l}{\textbf{MAE}} & \multicolumn{1}{c|}{\textbf{SMAPE}} & \multicolumn{1}{c}{\textbf{MSE}} & \multicolumn{1}{c}{\textbf{MAE}} & \multicolumn{1}{c}{\textbf{SMAPE}} \\
 \midrule
 
\textbf{M3 Y} & m4 y & 11513456.00 & 2128.49 & 34.61 & 6237766.00 & 1430.28 & 24.14 & 6009699.50 & 1378.28 & 23.53 & 16465432.218 & 2564.440 & 46.97 & \textbf{4400483.00} & \textbf{1046.66} & \textbf{17.71} \\
 & m4 q & 31463080.00 & 4102.82 & 53.40 & 8581998.00 & 1531.70 & 25.05 & 5597225.00 & 1453.16 & 31.97 & 5903342.781 & 1204.653 & 20.14 & \textbf{2944189.50} & \textbf{847.83} & \textbf{15.06} \\
 & m4 m & 35995652.00 & 3795.86 & 56.46 & 20737152.00 & 2470.83 & 42.24 & 7764455.00 & 1824.64 & 54.33 & 5251701.859 & 1080.925 & 23.19 & \textbf{2220906.75} & \textbf{650.69} & \textbf{14.90} \\
 & m4 w & 13053633.00 & 2421.21 & 41.97 & 4988010.00 & 1338.56 & 28.41 & 11481673.00 & 2234.78 & 62.90 & \textbf{751991.026} & \textbf{485.865} & \textbf{11.80} & 1002220.19 & 526.44 & 13.78 \\
 & m4 d & 7361305.00 & 1137.91 & 17.72 & 3434693.50 & 637.80 & 10.55 & 15111677.00 & 2589.88 & 57.06 & 951519.99 & 257.15 & 4.55 & \textbf{469240.38} & \textbf{223.86} & \textbf{3.70} \\
 & m4 h & 473064400.00 & 4189.93 & 72.56 & 361901100.00 & 3582.05 & 65.83 & 273323600.00 & 2466.15 & 50.43 & 39078221.126 & 1171.31 & 43.77 & \textbf{19900000.00} & \textbf{718.00} & \textbf{24.00}\\
  \midrule
  
\textbf{M3 Q} & m4 y & 13072291.00 & 2466.60 & 48.91 & 7317029.50 & 1674.70 & 30.86 & 11786057.00 & 2307.18 & 45.80 & 16465432.218 & 2564.440 & 46.97 & \textbf{4400483.00} & \textbf{1046.66} & \textbf{17.71} \\
 & m4 q & 9534236.00 & 1922.57 & 31.04 & 3877109.20 & 1031.58 & 18.70 & 5971352.00 & 1539.03 & 31.88 & 5903342.781 & 1204.653 & 20.14 & \textbf{2944189.50} & \textbf{847.83} & \textbf{15.06} \\
 & m4 m & 14431752.00 & 2243.24 & 41.00 & 5169829.00 & 1160.08 & 25.80 & 9657181.00 & 2061.39 & 61.40 & 5251701.859 & 1080.925 & 23.19 & \textbf{2220906.75} & \textbf{650.69} & \textbf{14.90} \\
 & m4 w & 4918492.00 & 1404.43 & 30.52 & 1345859.00 & 632.14 & 15.83 & 13399762.00 & 2245.55 & 61.70 & \textbf{751991.026} & \textbf{485.865} & \textbf{11.80} & 1002220.19 & 526.44 & 13.78 \\
 & m4 d & 2837104.00 & 672.05 & 11.90 & 1160018.10 & 363.23 & 6.76 & 18665982.00 & 2819.59 & 62.15 & 951519.99 & 257.15 & 4.55 & \textbf{469240.38} & \textbf{223.86} & \textbf{3.70} \\
 & m4 h & 90272040.00 & 1955.85 & 46.73 & \textbf{5985285.00} & \textbf{574.17} & 31.77 & 777291700.00 & 3937.91 & 72.37 & 39078221.126 & 1171.31 & 43.77 & 19900000.00 & 718.00 & \textbf{24.00} \\
  \midrule
  
\textbf{M3 M} & m4 y & 15971729.00 & 3180.59 & 128.05 & 4014987.00 & 1322.89 & 41.36 & 3797488.80 & 1325.22 & 40.50 & 16465432.218 & 2564.440 & 46.97 & \textbf{4400483.00} & \textbf{1046.66} & \textbf{17.71} \\
 & m4 q & 10547673.00 & 2175.64 & 49.35 & \textbf{2324206.50} & \textbf{722.85} & 14.14 & 2330858.00 & 751.31 & 14.64 & 5903342.781 & 1204.653 & 20.14 & 2944189.50 & 847.83 & \textbf{15.06} \\
 & m4 m & 3952263.50 & 1109.88 & 25.32 & \textbf{1273698.40} & \textbf{485.19} & 11.85 & 1407775.50 & 518.08 & 12.66 & 5251701.859 & 1080.925 & 23.19 & 2220906.75 & 650.69 & \textbf{14.90} \\
 & m4 w & 1879165.50 & 805.57 & 20.04 & \textbf{416051.34} & \textbf{350.82} & \textbf{10.02} & 571927.70 & 408.89 & 11.80 & 751991.026 & 485.865 & 11.80 & 1002220.19 & 526.44 & 13.78 \\
 & m4 d & 1161992.80 & 437.42 & 7.72 & \textbf{316829.30} & \textbf{171.40} & 3.00 & 267392.97 & 182.67 & \textbf{2.98} & 951519.99 & 257.15 & 4.55 & 469240.38 & 223.86 & 3.70 \\
 & m4 h & 120649720.00 & 1881.40 & 27.60 & 67445850.00 & 1428.50 & 49.18 & 41906224.00 & 1227.78 & 55.19 & 339078221.126 & 1171.31 & 43.77 & \textbf{19900000.00} & \textbf{718.00} & \textbf{24.00} \\
  \midrule
  
\textbf{M3 O} & m4 y & 24711976.00 & 3602.69 & 91.31 & 38270230.00 & 4747.50 & 134.75 & 32571130.00 & 4323.16 & 115.48 & 16465432.218 & 2564.440 & 46.97 & \textbf{4400483.00} & \textbf{1046.66} & \textbf{17.71} \\
 & m4 q & 14687010.00 & 2592.26 & 55.40 & 12203159.00 & 2274.19 & 47.23 & 10212250.00 & 2090.96 & 43.26 & 5903342.781 & 1204.653 & 20.14 & \textbf{2944189.50} & \textbf{847.83} & \textbf{15.06} \\
 & m4 m & 6362342.50 & 1397.35 & 32.22 & 7021934.00 & 1395.72 & 30.69 & 4285071.00 & 1091.02 & 24.50 & 5251701.859 & 1080.925 & 23.19 & \textbf{2220906.75} & \textbf{650.69} & \textbf{14.90} \\
 & m4 w & 2316830.50 & 901.06 & 20.64 & 2587111.20 & 900.57 & 21.44 & 1992984.50 & 815.82 & 17.43 & \textbf{751991.026} & \textbf{485.865} & \textbf{11.80} & 1002220.19 & 526.44 & 13.78 \\
 & m4 d & 1353906.10 & 541.05 & 10.28 & 1951016.00 & 562.14 & 9.71 & 1613977.10 & 718.86 & 11.87 & 951519.99 & 257.15 & 4.55 & \textbf{469240.38} & \textbf{223.86} & \textbf{3.70} \\
 & m4 h & 37831484.00 & 1125.90 & 35.50 & 35096910.00 & 995.15 & 35.42 & 53973530.00 & 1156.90 & 31.53 & 339078221.126 & 1171.31 & 43.77 & \textbf{19900000.00} & \textbf{718.00} & \textbf{24.00} \\
  \bottomrule
\end{tabular}%
}
\caption{
Zero-shot model transferability results for short-term forecasting from M3 to M4 dataset with horizon $H = 6$. Subsets of the M3 dataset are denoted by the first letter: yearly data -- Y, quarterly -- Q, monthly -- M and O stands for Other. The results for ForecastPFN (FPFN) and our model remain identical across various source datasets due to their training on synthetic data without source dataset. The best scores are highlighted in \textbf{bold}. A double line separates the columns with a pure zero-shot approach based on synthetics (Ours and ForecastPFN). Our approach, encompassing data generation methodology and architectural design, consistently outperforms the zero-shot ForecastPFN and well-known architectures for time series analysis trained on real data across various scenarios.}
\label{table:zeroshot_m3_m4}
\end{table*}

\end{document}